\documentclass{article}

\usepackage{PRIMEarxiv}

\usepackage[utf8]{inputenc} 
\usepackage[T1]{fontenc}    
\usepackage{hyperref}       
\usepackage{url}            
\usepackage{booktabs}       
\usepackage{amsfonts}       
\usepackage{nicefrac}       
\usepackage{microtype}      
\usepackage{lipsum}
\usepackage{natbib}
\usepackage{fancyhdr}       
\usepackage{graphicx}       
\graphicspath{{media/}{si_media/}}     
\usepackage{xr}
\usepackage{lineno}
\usepackage{amsmath}
\usepackage{graphicx}
\newcommand{\minus}{\scalebox{0.75}[1.0]{$-$}}

\usepackage{soul}
\usepackage{float}
\usepackage[figuresright]{rotating}
\usepackage[T1]{fontenc}
\newcommand{\silink}[2]{\hyperref[#1]{#2}}
\title{Should All Noises Be Treated Equally: Impact of Input Noise Variability on Neural Network Robustness
\thanks{\textit{\underline{Citation}}: 
\textbf{Alsinan, S., Makarenko, M., Liu, S., Aldawood, A., \& Hoteit, I. (2026). Should all noises be treated equally: Impact of input noise variability on neural network robustness. Journal of Geophysical Research: Machine Learning and Computation, 3, e2025JH000968. https://doi.org/10.1029/2025JH000968}}}

\author{S. Alsinan$^{1,2}$, M. Makarenko$^{2}$, S. Liu$^{1}$, A. Aldawood$^{3}$, and I. Hoteit$^{1}$ \\
$^{1}${King Abdullah University of Science and Technology, Saudi Arabia}\\
$^{2}${Aramco Upstream Research Center at KAUST, Saudi Arabia}\\
$^{3}${EXPEC Advanced Research Center, Aramco, Saudi Arabia}\\
salma.alsinan@kaust.edu.sa}

\begin{document}
\maketitle
\begin{abstract}
Geophysical data collected from active field sites are often contaminated by complex and heterogeneous noise, obscuring weak seismic events, and complicating automated interpretation. Although deep learning offers promising solutions for seismic processing, its performance is highly sensitive to the nature of training noise, especially under out-of-distribution (OOD) conditions. This study investigates the influence of noise parameters, such as type, scale, and complexity on the performance, generalization and robustness of neural networks in two geophysical tasks: first break picking and denoising. We simulate seismic-while-drilling data and apply controlled input source noise augmentation using stochastic generators to vary the noise characteristics. Different neural networks are trained on fixed noise types and scales, then evaluated across both seen and unseen noise scenarios. We incrementally increase the complexity of the noise by introducing compound noise mixtures and assess the performance of the model under increasingly challenging OOD conditions. This yields a robustness matrix that captures the generalizability of each model relative to its training configuration. Results indicate that larger noise scales boost generalization, and that effective alignment between noise type, task complexity, and architecture is key for maximizing generalization gains. In addition, training with compound noises mitigate weaknesses associated with single-noise training, acting as an additional implicit regularizer to improving robustness. These findings highlight key factors influencing model resilience in noisy geophysical environments and offer guidance for developing deep learning models that generalize effectively across diverse and unpredictable noise conditions.
\end{abstract}

\keywords{Noise \and Generalization \and Distribution-shift \and out-of-distribution generalization \and compound noise training \and denoising \and first break picking \and seismic-while-drilling \and dense prediction.}

\section{Introduction}
Noise is an inevitable factor contributing to variability in repeated measurements, often referred to as the component in observed data which cannot be explained by a simple deterministic model \cite{scalesandsnieder}. In seismic data, identifying the source of this variability is particularly challenging due to the inherit non-uniqueness of the seismic response. The ill-condition nature of the inverse problem \cite{Debski1995}, and the band-limited characteristics of seismic data keep the inverse solution space highly uncertain, which is further amplified in the presence of noise \cite{Cambois,houck, Buland&Omre2003, downton2005seismic}. Therefore suppressing noise during acquisition and attenuating it during processing is curial for successful interpretation \cite{chopra2014avo}.

In seismic recordings, noise can manifest in various forms (e.g., coherent, random, etc.) and exhibit distinct characteristics attributed to acquisition setups, source and receiver types, as well as environmental conditions \cite{yilmaz2001seismic,BONNEFOYCLAUDET2006205,bormann2013seismic}. Although certain types of noise offer valuable insight into the near-surface conditions, noise is generally attenuated during the modeling and stacking stages to reveal the underlying information of the subsurface reflectors and enable interpretation. Hence, modeling and managing noise in seismic data have a substantial impact on the quality of seismic images and the feasibility of conducting accurate inversions \cite{yilmaz2001seismic,downton2005seismic, tarantola2005inverse,Heidari}. Achieving reliable results in seismic characterization involves striking a careful balance between suppressing noise and preserving signal amplitude \cite{Cambois,chopra2014avo}. Over the years, various denoising techniques have been developed to isolate targeted noises from the seismic signal \cite[e.g.,][]{Larner,Bednar,canales1984random,jonesandlevy,ulrych1999,chase1992random,abmaandClaerbout,HarrisandWhite,BekaraandVan}. The selection of an appropriate attenuation method and separation domain depends on the nature and characteristics of the noise \cite{yilmaz2001seismic}, and is addressed across the processing workflow. However, noise attenuation becomes significantly more challenging in real-time acquisition and monitoring environments \cite{poletto2004seismic,maxwell2014microseismic,birinie2016,Tsuji}. Unlike quiet acquisition, continuous interference from various noise sources creates complex patterns that can mask weak events, which present a challenge for current noise suppression techniques. To overcome this challenge, a combination of denoising techniques is typically applied, with each method targeting a specific type of noise \cite{dando2016realtime}.

In recent years, nonlinear neural networks have been used to address various seismic processing tasks \cite{yu2019deep,ANIKIEV2023104371,anjom2024machine} under different learning paradigms (supervised, semi-supervised, unsupervised, self-supervised, etc.). The extensive body of literature highlights the inherent complexity of modeling noise for training neural networks and the ongoing challenges in developing robust, generalizable solutions that can adapt to the diverse types of noise encountered in real-world scenarios. Since the effectiveness of AI-based approaches often depends on the assumptions made during noise modeling, their applicability tends to diminish when faced with field data that differ from the training conditions \cite{kim2018geophysical,kouw2019introductiondomainadaptationtransfer,zheng2019applications,geirhos2020generalisationhumansdeepneural,smith2021robust,smith2022robust,alkhalifah2022mlreal}. This topic is particularly important in the field of geophysics, where forward modeling is heavily relied upon to generate synthetic data for training AI solutions. This approach helps address the limited availability of labeled field examples needed for training neural networks and reduces the risk of bias introduced by manual labeling, which can compromise the reliability of the network’s predictions \cite{adler2021deep,alkhalifah2022mlreal}. Therefore, incorporating realistic noise characteristics into synthetic data is essential for accurately simulating field conditions, highlighting the need for continued research into alternative noise modeling approaches that better capture real-world variability \cite{liu2020,BIRNIE202147,SaadandChen,birnie2022leveraging,liu2022coherentnoisesuppressionselfsupervised}. Without such realism, AI models might not be trained effectively to recognize or adapt to real-world scenarios.  

The topic of noise in AI is expansive, spanning research across various domains and applications, which highlights its multifaceted role. Among these research directions is the use of noise to promote generalization in neural networks. Early studies provided theoretical and empirical evidence that noise injection during training can improve generalization by discouraging overfitting \cite{SIETSMA199167, Holmstrom, 155944}. Based on these observations, \cite{bishop1995} demonstrated that training a neural network with small-scale Gaussian noise added to the input is mathematically equivalent to applying Tikhonov regularization ($\ell_2$) to a least-squares optimization problem. These insights have motivated the development of a variety of regularization techniques, including deterministic methods such as early stopping \cite{FINNOFF1993771}, which limit overfitting without introducing noise, as well as stochastic approaches that inject randomness at different stages of the AI pipeline ( i.e., data, network, training dynamics, and post-training) to improve generalization to unseen or variable conditions. 

In the data domain, noise can be introduced into inputs, labels, input–label pairs, or into the feature representations to enhance variability \cite[e.g.,][]{szegedy2015rethinkinginceptionarchitecturecomputer, schluter-2015-exploring, devries2017datasetaugmentationfeaturespace, zhang2018mixupempiricalriskminimization, guo2019augmentingdatamixupsentence, yun2019cutmixregularizationstrategytrain, hendrycks2020augmixsimpledataprocessing, müller2020doeslabelsmoothinghelp, camuto2021a,song2022learningnoisylabelsdeep}. These techniques aim to reduce the gap that may exist between the source (training) and target domains, which can arise due to variations in data collection processes, mislabeling caused by human error, or environmental conditions. In addition, these techniques also serve to mitigate the tendency of networks to become overly confident in their predictions \cite{kukačka2017regularizationdeeplearningtaxonomy}. Noise can also be introduced into various internal components of the network, including activations and weights, as a form of explicit regularization \cite{krogh1991, Srivastava2014, blundell2015weightuncertaintyneuralnetworks, kingma2015variationaldropoutlocalreparameterization, gal2016dropoutbayesianapproximationrepresenting, noh2017regularizingdeepneuralnetworks, pereyra2017regularizingneuralnetworkspenalizing}. Beyond improving generalization, such noise injections introduce output variability that enhances predictive reliability and supports more effective uncertainty estimation \cite{camuto2021a}. Noise can also be introduced into the gradients to promote exploration of the loss landscape \cite{welling2011,neelakantan2015addinggradientnoiseimproves}, or leveraged through the implicit regularization effects of stochastic optimization methods \cite{neyshabur2015searchrealinductivebias}, such as Batch Normalization \cite{ioffe2015batchnormalizationacceleratingdeep} and Stochastic Gradient Descent (SGD) \cite{zhang2017understandingdeeplearningrequires}. Another important application of noise in AI is to strengthen the robustness of pre-trained networks against noise perturbations commonly encountered in real-world settings. The goal is to mitigate the influence of such distortions on the behavior of the network and the accuracy of the prediction. This concept was first introduced by \cite{Biggio_2013} and \cite{szegedy2014intriguingpropertiesneuralnetworks}, who demonstrated that networks are highly fragile and susceptible to small noise perturbations, thereby introducing the concept of adversarial attacks and the existence of adversarial examples. This concept is based on the understanding that machine learning algorithms are trained on an incomplete segment of the theoretical data distribution. As a result, new or unseen data may exhibit properties not represented in the training set \cite{chio2018machine}. Adversarial attacks exploit these gaps by introducing subtle perturbations to inputs or labels, potentially leading to incorrect or manipulated predictions. This discovery has significantly heightened awareness of the need for robustness in AI models and the importance of identifying vulnerabilities through adversarial testing \cite{goodfellow2015explainingharnessingadversarialexamples,akhtar2018threatadversarialattacksdeep,chakraborty2018adversarialattacksdefencessurvey,madry2019deeplearningmodelsresistant,gu2019, gao2020backdoorattackscountermeasuresdeep}. Although such attacks are typically performed during the deployment phase, data poisoning can also be leveraged during model development to proactively assess and improve resilience to backdoor attacks \cite{li2022backdoorlearningsurvey}. Beyond robustness, noise is leveraged in various applications, including generating realistic data and images, as well as in tasks like denoising and image restoration \cite{goodfellow,chen2015learningoptimizedreactiondiffusion,diffusion,yue2020dualadversarialnetworkrealworld}. 

Neural networks demonstrate a remarkable capacity to generalize, even in settings that challenge conventional learning theory. This was notably demonstrated in the context of image classification by \cite{zhang2017understandingdeeplearningrequires}, who provided empirical evidence that convolutional neural networks can achieve strong generalization on i.i.d test datasets, even in the absence of regularization or when trained with noisy labels. Their study revealed that modifications to network architecture or the use of data augmentation resulted in greater gains in generalization compared to tuning implicit or explicit regularization techniques. This suggests that while regularization methods contribute to generalization, they are not the primary drivers. As a result, there has been growing interest in understanding the true causes behind network generalization \cite{neyshabur2015searchrealinductivebias,neyshabur2017exploringgeneralizationdeeplearning,Kawaguchi_2022}. Despite the strong generalization performance of neural networks on i.i.d. test datasets, their learned features fail to transfer effectively to other tasks or datasets without additional fine-tuning, particularly when faced with significant domain or distribution shifts. This suggests that their effectiveness is largely limited to tasks that are closely related \cite{Pan2010,yosinski2014transferablefeaturesdeepneural}.

In recent years, there has been growing interest in \textit{Domain Adaptation} (DA) \cite{DAWang} and \textit{Domain Generalization} (DG) \cite{DGZhuo} techniques as strategies to address distribution shifts between source and target datasets. While both approaches share commonalities in using shallow and deep methods \cite{VenkateswaraandPanchanathan}, the key difference lies in data access; DA has access to the target dataset during training, whereas DG does not. As such, DG aims to learn domain-invariant features that generalize to unseen scenarios \cite{blanchard2011, muandet13}, aligning with the practical realities of deploying AI in environments where distribution shifts are the norm. DG methods can be broadly categorized into data augmentation strategies \cite{TobinFRSZA17, wang2025comprehensivesurveydataaugmentation}, learning-based strategies (e.g., meta-learning \cite{Li2017LearningTG}, self-supervised learning \cite{9086055}, ensemble learning \cite{MOHAMMED2023757}, etc.) and domain alignment strategies \cite{muandet13}. These methodologies seek to mitigate the effects of distributional shifts manifesting in either the input space (covariate shifts) or the label space (semantic shifts). Among these, covariate shifts are predominantly studied and serve as a primary benchmark for assessing model robustness and generalization. To assess robustness, algorithms are evaluated based on their performance on \textit{in-distribution} (ID) versus \textit{out-of-distribution} (OOD) datasets, typically by training on a given dataset and testing on held-out or distributionally shifted data \cite{miller2021accuracylinestrongcorrelation}. Confidence calibration is another key evaluation metric, capturing the alignment between a model’s predicted confidence and its actual accuracy, which does not always correlate with generalization \cite{guo2017calibrationmodernneuralnetworks}. 

The distinction between ID and OOD inputs was first highlighted by \cite{hendrycks2018baselinedetectingmisclassifiedoutofdistribution}, who observed that classifiers tend to assign lower softmax confidence scores to OOD samples compared to ID samples across various computer vision (CV) tasks. This enables a simple yet effective baseline for OOD detection without requiring additional training, using softmax outputs as a proxy for input familiarity. This observation has fueled extensive research on enhancing OOD generalization and detection capabilities \cite{salehi2022unifiedsurveyanomalynovelty,yang2024generalizedoutofdistributiondetectionsurvey}, as well as understanding the correlation between ID and OOD performance in neural networks. In this context, \cite{miller2021accuracylinestrongcorrelation} observed a linear correlation between ID and OOD performance of various deep learning and machine learning models trained on image classification tasks, which is significantly influenced by the covariance of the underlying data distributions. In particular, they observe that alterations in training duration, hyperparameters, and size of the training dataset do not disrupt this linear trend; rather, they cause shifts along the trend. In contrast, certain distribution shifts, such as those introduced by noise perturbations, exhibit a poor correlation between ID and OOD performance. The authors found that when noise is sampled from a distribution with the same covariance as the training data, the previously observed linear trend is preserved, unlike when isotropic Gaussian noise of equivalent scale is applied. \cite{teney2023idoodperformanceinversely} systematically analyzed this relationship under varying non-noise-related distribution shifts, showing that the correlation can weaken or even reverse depending on the severity of the shift. A comparable inverse relationship has been documented in settings with noisy labels, where increasing the severity of label noise leads to a deterioration in the alignment between ID and OOD performance \cite{sanyal2024accuracywronglinepitfalls}. Although architectural choices play a role in shaping the complex ID-OOD relationship, \cite{wenzel2022assayingoutofdistributiongeneralizationtransfer} emphasize that the dataset, the nature of the task, and the specific type of distribution shift are the predominant influences on OOD generalization. These observations underscore the nuanced effects of specific data transformations on performance and the need for caution, as strong ID results do not necessarily imply robust OOD generalization. Consequently, a trade-off  between ID and OOD performance is frequently observed \cite{teney2023idoodperformanceinversely}. To further explore how models handle distributional shifts and distortions, \cite{geirhos2020generalisationhumansdeepneural} investigated the effect of various image distortions, including a range of noise perturbations, on the robustness of convolutional architectures in classification tasks. Their experiments demonstrate that training with specific distortions can enhance network performance, sometimes even exceeding human accuracy on the same types of noise. However, these learned features generalize poorly to unseen distortions, with performance deteriorating sharply as the Signal-to-Noise ratio (SNR) of the test images decreases, revealing a key limitation. This motivates a focused investigation into how input-source noise parameters influence the relationship between ID and OOD performance of neural networks. While prior studies have broadly explored the impact of regularization and noise-based techniques on OOD generalization \cite{chun2020empiricalevaluationrobustnessuncertainty,ferianc2024navigatingnoisestudynoise}, they often do not specifically isolate input-source noise, particularly in the context of segmentation and denoising tasks. The objective of this study is to identify key factors that influence learning under noisy conditions and contribute to enhancing the robustness of neural networks to OOD noise patterns. We present a detailed empirical analysis of how the alignment of noise characteristics, such as type, scale, and complexity, with network architecture and task affects robustness to previously unseen noisy conditions. This work aims to guide researchers and practitioners developing geophysical AI applications in active field environments, as well as those seeking to enhance model generalizability by leveraging domain-specific noise patterns inherent to geophysical settings.

\section{Data Preparation}
A recorded seismogram $x(t)$ is described by two components (Equation~\ref{eq1}): the noise-free traces $s(t)$, produced by convolving the earth's impulse response $r(t)$ with a source wavelet $w(t)$, and the accompanying noise traces $\xi(t)$ \cite{yilmaz2001seismic}: 
\begin{linenomath*}
    \begin{equation} \label{eq1}
    x(t)=\ w(t)\ast\ r(t)+\ \xi(t) \ .
    \end{equation}
\end{linenomath*}
Modeling of the synthetic $s(t)$ and noise $\xi(t)$ traces used for training are described in the subsections below.
\subsection{Synthetic Seismic Data}

In a Seismic While Drilling (SWD) survey, continuous recording of drill-bit vibrations provides the active seismic source while receivers are deployed on the surface \cite{rector1991use,poletto2004seismic}. Because SWD acquisition occurs during drilling operations, the bit signal is typically weak and heavily contaminated by high-amplitude noise, making interpretation challenging without effective noise-suppression and attenuation strategies \cite{poletto2004seismic,silvestrov2021processing}.

\begin{figure}[h]
    \includegraphics[width=\textwidth]{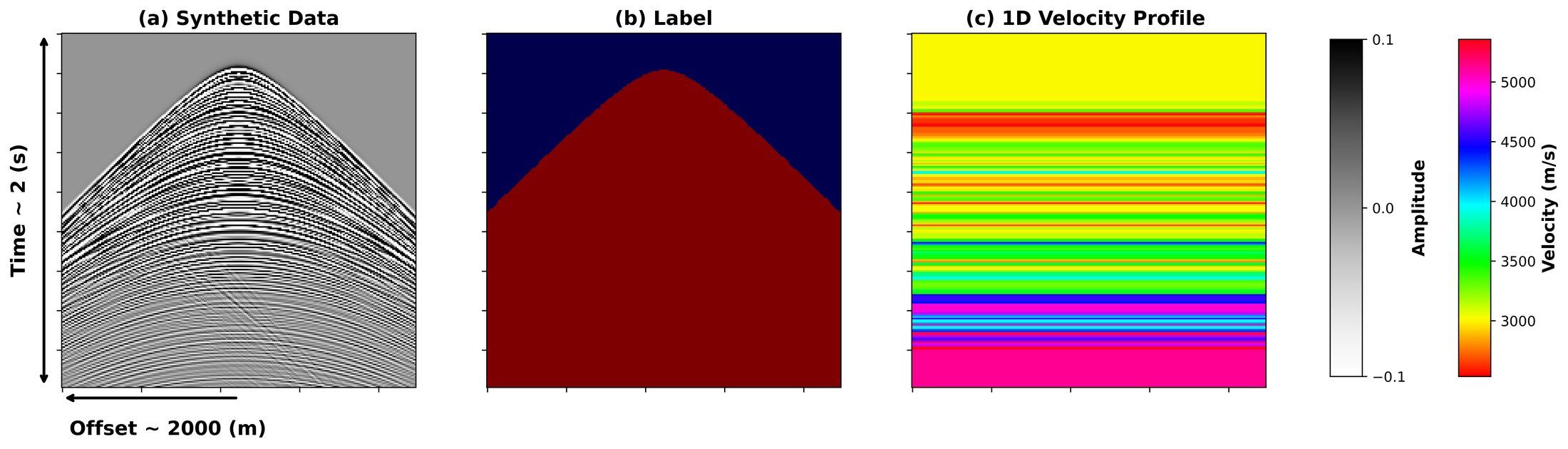} 
    \caption{Random shot gather (a) synthetic shot gather (b) segmentation label (c) 1D velocity profile.}\label{f1}
\end{figure}

The synthetic dataset used in this study is generated for a SWD field survey described by \cite{dawood2021}, using its acquisition geometry and associated velocity model \cite{aldawood2022characterizing}. We replicate a 2D east–west walkaway configuration consisting of 160 receivers symmetrically distributed around the wellbore, with offsets ranging from 25 to 2000 m. Shot positions are spaced at 1 m intervals down to a depth of 2000 m. Using a finite-difference solution to the acoustic wave equation and a Ricker wavelet with a dominant frequency of 35 Hz, we simulate 2,121 synthetic shot gathers $s$(t) within this SWD acquisition geometry (Figure~\ref{f1}).  

\subsection{Noise Forward Models (Stochastic Noise Generator)}\label{NG}

The addition of noise is essential for enhancing the realism of synthetic data and bridging the gap with the observed field data \cite{birnie2022leveraging}. Therefore, to incorporate the missing noise components into the signal, we developed a stochastic noise generator capable of producing various noise profiles $\xi$(t) that may be encountered in the field. These noise profiles are applied directly to the data; some modify the signal $s(t)$, while others are combined with the signal in a linear manner $\xi(t)$. The goal is to systematically diversify and expand the synthetic data through the introduction of input-source noise perturbations, and enable the assessment of how noise type, amplitude, and complexity affect network prediction for first break picking and denoising tasks. 
 
Expanding on Equation~\ref{eq1}, the training data denoted below as $x(t)$ is represented as a combination of two forward models (Equation~\ref{eq2x}), where $F$ is a data model that acts by modifying or removing parts of the signal $s(t)$, and $G$ is the noise model which is added linearly to the output of the data model $F$. The additive noise profiles $\xi(t)$ generated by $G$ are scaled by a parameter $\beta$, which helps control the Signal-to-Noise ratio (SNR) of the output image $x$. 
\begin{linenomath*}
    \begin{equation} \label{eq2x}
        x\left(t\right)=\ F\left(s\left(t\right)\right)+\ \beta\ G\left(n\left(t\right)\right)\ .
    \end{equation}
\end{linenomath*}
The data model $F$ consists of a selection of noises that alter the signal by removing some frequency components (high-pass and low-pass filters) and dropping random traces using a trace-wise mask. This model can also act as an identity function, where no alterations are made to the original signal $s(t)$. The noise model $G$ is responsible for generating additive noise profiles $\xi(t)$ which can be augmented in different domains (i.e, spatial $(x-t)$ and frequency $(f-k)$). $G$ has access to 11 noise functions $n(t)$ to generate both random and coherent noise profiles. In this work, we limit the analysis to 6 types of noise that are frequently used in the development and testing of noise attenuation algorithms. 
\begin{enumerate}
\item \textit{Random noises}: uncorrelated white Gaussian noise applied in $(x-t)$ domain, and Gaussian noise augmented in the Fourier $(f-k)$ domain referred to as Gaussian (f-k). The latter augmentation style could present more localized noise patterns in certain directions or frequencies. 
\item \textit{Spectral noises}: Colored noise applied in the temporal $(t)$ direction, introducing temporal variability while maintaining smooth spatial variation across receiver positions. Bandpassed noise applied in the spatial $(x)$ direction introduce lateral variation across the receiver line. The frequency-based nature of bandpassed and colored noise gives rise to subtle structure, though these remain largely incoherent across space and time, producing irregular, fragmented patterns.
\item \textit{Spatially structured and coherent noises}: Linear and hyperbolic noises augmented in the $(x-t)$ domain and have specific geometrical patterns with spatial and temporal correlations. Additional analysis of the noise’s spatial characteristics is available in \textit{Supporting Information}, Text~\silink{Text2}{S2}, and Figures~\ref{sf0}-~\ref{sfx0}.
\end{enumerate}
To incorporate stochasticity into noise generation process, we employ Monte Carlo sampling to randomly select the parameters ${\ p}_i$ of each noise function $n$(t) from a normal distribution (Equation~\ref{eq3x}). This randomization allows us to generate different realizations of each noise type, which are then weighted by a noise scale $\beta$ and combined with the output of the data model $F$. 
\begin{linenomath*}
    \begin{equation} \label{eq3x}
        G_{single}\ =\left\{\ {n(p}_i\right)\ {:\ p}_i\ \sim \mathcal{N}\left(0,\ \sigma^2I\right)\} \ .
    \end{equation}
\end{linenomath*}

Single noise types can be found in field data, and training neural networks using these augmentations can be beneficial for noise-specific applications. However, success is dictated by the ability to identify, model and isolate these noises, which is challenging in real-time acquisition and monitoring setups \cite{birinie2016}. For example, the noise found in SWD datasets is typically an aggregation of various noise types and sources, requiring the use of different attenuation algorithms for effective mitigation. To address this challenge, we extended the noise generator’s capabilities to produce compounded noise profiles $\xi_c(t)$. In this setup, $G$ linearly combines multiple noise functions $n(t)_{k}$ (Equation~\ref{eq5x}), which can be applied in different domains (spatial and frequency). Furthermore, the generator can combine the data manipulations produced by $F$ with the additive noise profiles $\xi_c(t)$ generated by $G$.  When these noise perturbations are combined, they produce complex noise patterns that distort the original signal. 
\begin{linenomath*}
    \begin{equation} \label{eq5x}
        G_{compound}=\left\{\ \sum_{k}^{K}{n\left(p_1,p_2,..p_i\right)}_k \ \ {:\ p}_i\ \sim \mathcal{N}\left(0,\ \sigma^2I\right)\right\}\ . 
    \end{equation}
\end{linenomath*}
Similar to the single noise generator, the linear combination of noise profiles $\xi_c(t)$ is weighted by a parameter $\beta$ (Equation~\ref{eq2x}) which provides control over the compound noise amplitude. The noise scale $\beta$ is normalized by the square root of the number of noises (Equation~\ref{eq6x}) to ensure consistency across the samples. The $\beta$ parameter can also be randomized during the network training. However, we kept its value constant to evaluate the impact of noise scale on network training, performance and generalizability. Since $\beta$ does not directly restrict the amplitudes generated by $n(t)$ like clipping, the generator can produce a confined range of disturbances to the image. 
\begin{linenomath*}
    \begin{equation} \label{eq6x}
        \beta=\ \frac{Noise\ scale}{\sqrt{K}} \ , \  K= number\ of \ noises, 
    \end{equation}
\end{linenomath*}
these stochastic noise generators enhance the clean synthetic dataset through the addition of realistic noise profiles found in the field. They facilitate the expansion of the training dataset, enable control over SNR, offer the ability to design complex noise perturbations, and provide options for augmentation domain, thereby adding complexity to the noise generation process. Further information on the noise generator and PSNR analysis can be found in the \textit{Supporting Information}, Text~\silink{Text3}{S3}, and Figures~\ref{f2}–~\ref{f7x}.
\section{Methodology} 

\begin{figure}[!htb]
    \noindent
    \includegraphics[width=\textwidth]{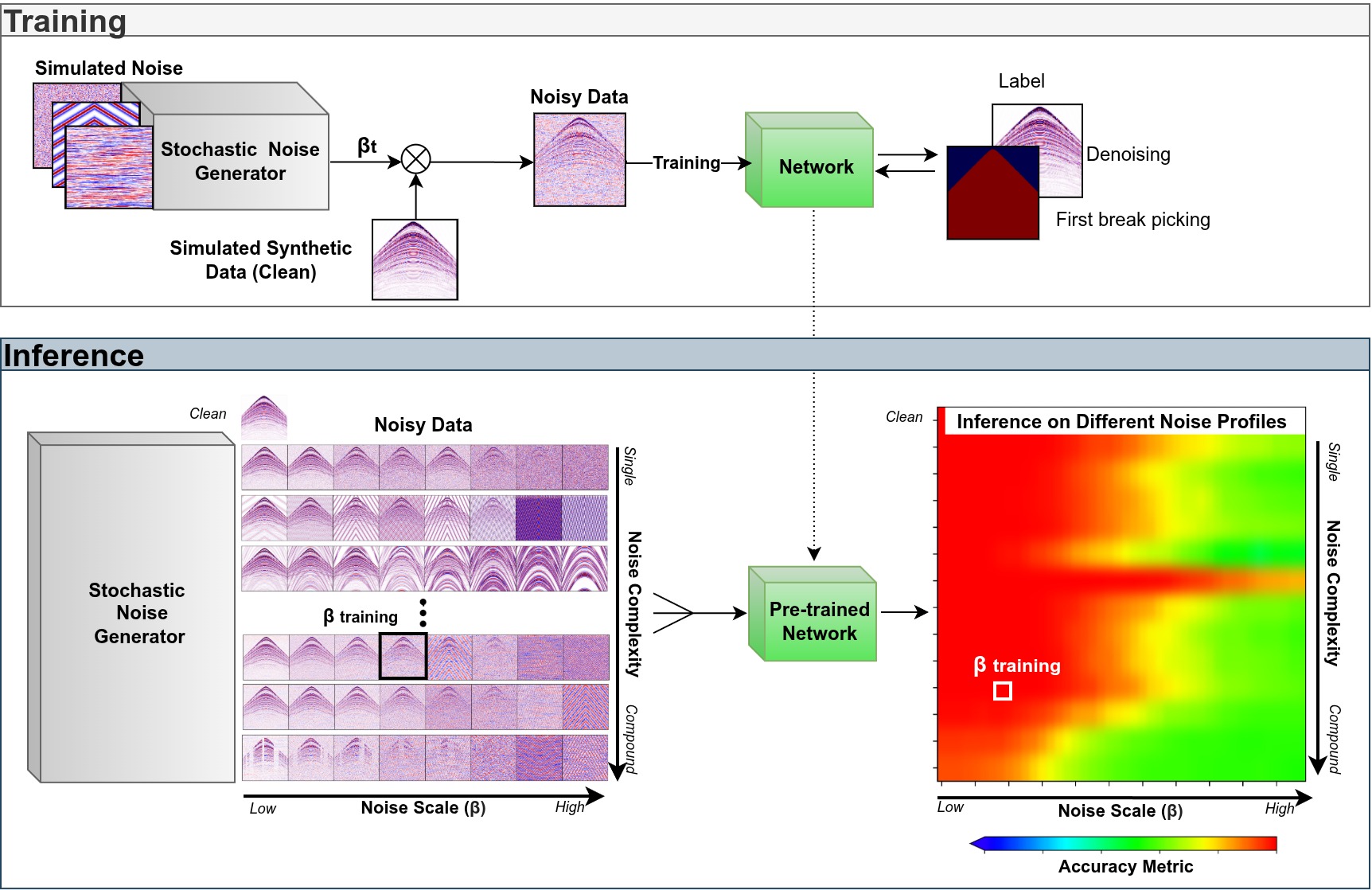} 
    \caption{Training and inference pipelines.}
    \label{f8}
\end{figure}

In this section, we outline the methods employed to investigate the effect of input-source noise augmentations on the training, performance, and generalizability of neural networks. Figure~\ref{f8} illustrates the training and inference pipelines used in this study, where different deep learning architectures are trained using a stochastic noise generator with fixed noise type and $\beta$ parameters. The performance of each network is then evaluated across a range of both seen and unseen noise scenarios. The primary objective is to quantify the network's performance when faced with more complex and higher amplitude noises, as well as simpler and lower amplitude noises produced by the noise generator. This experimental framework resembles an in-distribution (ID) and out-of-distribution (OOD) testing setup, allowing us to explore the effect of certain noise transformations on network performance. Although the synthetic data remains constant, variation in noise characteristic can lead to shifts between the training and testing distributions \cite{miller2021accuracylinestrongcorrelation}. Through this setup, we aim to identify the parameters that influence network performance in the presence of input-source noise.
 
\subsection{Deep Learning Architectures}

Supervised deep learning involves the use of labeled data y to learn a mapping function that connects inputs x to the corresponding outputs $\hat{y}$ (Equation~\ref{eq8}). This parameterized and differentiable function $f_{\theta= \{\boldsymbol{W},b\}}$ is represented by the network architecture, which comprises various linear and nonlinear layers capable of modeling complex relationships. The function $f_\theta$ is optimized by minimizing an objective function $\mathcal{L}$, (Equation~\ref{eq9}), thereby guiding the network to improve its prediction over time.
\begin{linenomath*}
    \begin{equation} \label{eq8}
        \hat{y}=f_\theta\left(x\right)\ ,
    \end{equation}
        \begin{equation} \label{eq9}
        \mathcal{L}\left(\theta\right)=\mathop{\mathbb{E}}_{x,y}\left[\ell\left(f_\theta\left(x\right),y\right)\right] \ , 
    \end{equation}
\end{linenomath*}
where $\mathbb{E}$ denotes the expectation over the data distribution, and $\ell$ is a loss function to measure the network’s prediction error. 

Although there are methods to optimize network architecture design \cite{elsken2019neuralarchitecturesearchsurvey}, the selection of a network architecture is typically based on heuristics that consider factors such as data structure, task, and available computational resources, among others. The underlying guideline is to select the appropriate level of model complexity that aligns with the complexity of the task to avoid overfitting \cite{bishop1995regandcomplexity,bengioanddelalleau,neyshabur2015searchrealinductivebias,Goodfellow-et-al-2016,raghu2017expressivepowerdeepneural}. For this study, we evaluate the performance of three encode-decoder architectures that are typically utilized for image analysis and restoration tasks: 
\begin{enumerate}
\item U-Net \cite{ronneberger2015unetconvolutionalnetworksbiomedical} is a convolutional encoder–decoder architecture designed to capture spatial hierarchies of features through successive down-sampling and up-sampling stages, with ReLU activations \cite{agarap2019deeplearningusingrectified} applied between convolutional layers to introduce nonlinearity. Skip connections link corresponding encoder and decoder levels, allowing the network to retain fine-scale spatial details. U-Net is widely used in medical imaging and geophysical segmentation tasks.

For our experiments, we adopt the lightweight PyTorch implementation of \cite{buda2019association}, which employs four encoder–decoder stages, an initial feature width of 32 channels, and a 512-channel bottleneck. This version differs from the original U-Net by incorporating Batch Normalization \cite{ioffe2015batchnormalizationacceleratingdeep} after each convolution, which improves optimization stability.
\item \textbf{S}hifted \textbf{Win}dow (Swin) Transformer \cite{liu2021swintransformerhierarchicalvision} is a vision transformer that employs shifted window self-attention within a hierarchical structure to capture multi-scale features. Each transformer block employs GELU nonlinearities \cite{hendrycks2023gaussianerrorlinearunits} and applies Layer Normalization \cite{ba2016layernormalization} before its self-attention and MLP sublayers. Its computational efficiency and versatility have led to its widespread adoption across various vision tasks. 
For our experiments, we adopt the Swin-Unet (Swin-U) architecture of \cite{cao2021swinunetunetlikepuretransformer}, which arranges Swin Transformer blocks into a U-Net style encoder–decoder. The encoder consists of four hierarchical stages, each stage contains two Swin Transformer blocks, followed by a symmetric decoder and skip connections. This architecture has been widely used in medical image segmentation due to its ability to combine multiscale feature extraction with efficient local attention. To obtain a lighter version, we reduce the patch-embedding dimension from the original Swin-U to 48, yielding roughly half the width of the standard model while maintaining the same stage structure and depth.
\item \textbf{Restor}ation Transfor\textbf{mer} (Restormer) \cite{zamir2022restormerefficienttransformerhighresolution} is a hierarchical encoder–decoder architecture designed to capture multi-scale and long-range pixel dependencies through a full self-attention mechanism. Its Multi-DConv Head Transposed Self-Attention (MDTA) module is specifically tailored for image restoration, enabling the network to emphasize both local details and broader spatial structures that are essential for recovering signals degraded by noise. The architecture consists of four resolution levels in the encoder, a transformer-based bottleneck, and a symmetric four-level decoder with skip connections, structurally parallel to U-Net and Swin-U. Within each transformer block, Restormer uses Gated-DConv Feed-Forward Networks (GDFN) with GELU nonlinearities inside the gated MLP units, while the surrounding computation remains relatively linear compared to other transformers. similar to Swin-U, the transformer blocks employ Layer Normalization \cite{ba2016layernormalization}. For our experiments, we adopt the public implementation but reduce the base embedding dimension to 24, yielding a lighter model than the original.
\end{enumerate}

Modifications to the architectures were implemented to ensure a fair comparison of model capacities, resulting in comparable training parameters; Restormer (6.67M), Swin-U (6.82M), U-Net (7.76M).  In terms of computational complexity, Restormer is the most demanding, requiring 226.12 billion FLOPs to train a single instance, followed by U-Net at 83.96 billion FLOPs and Swin-U at 11.96 billion FLOPs. All architectures employed an identity function at the output layer, and no extra nonlinear activation was added. Any required nonlinear behavior was handled directly by the task-specific loss function (e.g., the log-softmax operation for segmentation). Further details can be found in the \textit{Supporting Information}, Text~\silink{Text4}{S4}.

\subsection{Training and Inference Setups}

The objective function $\widetilde{\mathcal{L}}(\theta)$ for training a neural network under input source noise $\xi$ is described as 
\begin{linenomath*}
    \begin{equation} \label{eq10}
        \widetilde{\mathcal{L}}(\theta)=\ \mathop{\mathbb{E}}_{x,y,\xi}\left[\ell(f_\theta\left(x+\xi\right),y)\right]\ .
    \end{equation}
\end{linenomath*}

\cite{bishop1995} demonstrated that training with additive low variance input-source Gaussian noise, $\xi \sim \mathcal{N}(\theta, \sigma^2 \text{I})$, where the noise is statistically independent of both the input $x$ and target $y$ distributions, is analogous to adding a regularization term to the objective function. In the case of least squares, this corresponds to Tikhonov regularization (Equation~\ref{eq11}), with a similar smoothing effect observed in cross-entropy classification. Consequently, no additional explicit regularization term is applied to the objective function during training.
\begin{linenomath*}
\begin{equation} \label{eq11}
    \widetilde{\mathcal{L}}(\theta) \approx \mathcal{L}(\theta) + \frac{\sigma^2}{2} \operatorname{Tr} \left[ \nabla_x^2 \mathcal{L}(x, \theta) \right] \ , 
\end{equation}
\end{linenomath*}
where $\mathcal{L}(\theta)$ is the noise-free loss, $\nabla_x^2 \mathcal{L}(x, \theta)$ is the Hessian of the loss with respect to the input $x$, $\operatorname{Tr}$ denotes the trace operator, and $\sigma^2$ is the variance of the input noise.

The scope of our assessment includes two seismic processing tasks that vary in complexity. The first task involves identifying the P-wave arrival, commonly referred to as first break picking. We approach this task as a segmentation problem, utilizing a Binary Cross Entropy (BCE) Loss (Equation~\ref{eq12}) and mean Intersection over Union (mIoU) as our accuracy metric.

\begin{linenomath*}
    \begin{equation} \label{eq12}
        \ell_{\text{BCE}} = -\frac{1}{N} \sum_{i=1}^{N} \left[ y_i \log(\hat{y}_i) + (1 - y_i) \log(1 - \hat{y}_i) \right] \ ,
    \end{equation}
\end{linenomath*}
where $N$ is the number of samples, $y_i \in \{0,1\}$ is the true label, and $(\hat{y}_i) \in \{0,1\}$ is the predicted probability of sample $i$. 
The second task poses a greater challenge, centering on denoising. We formulate this task as a regression problem, utilizing $L_2$ Loss (Equation~\ref{eq13}) and evaluating performance using Root Mean Squared Error (RMSE):

\begin{linenomath*}
    \begin{equation} \label{eq13}
        \ell_{\text{MSE}} = \frac{1}{N} \sum_{i=1}^{N} (\hat{y}_i - y_i)^2 \ ,
    \end{equation}
\end{linenomath*}
where $N$ is the number of samples, $y_i$ is the target value, and $\hat{y_i}$ is the predicted value of sample $i$.

Prior to training, the learning rate and batch size hyperparameters were optimized using a Tree-Structured Parzen Estimator (TPE)~\cite{bergstra} and fixed across all training scenarios, details can be found in the \textit{Supporting Information}, Text\silink{Text5}{S5}, Figures~\ref{f41}–~\ref{f41x} and Table~\ref{T2}.
Each network was trained from scratch for 100 epochs using mini-batch Stochastic Gradient Descent (SGD) and Adam optimizer~\cite{kingma2017adammethodstochasticoptimization}, with a learning rate of $5e^{-5}$ and a batch size of 8. By standardizing implicit regularization across all experiments and applying the aforementioned modifications to network capacity, we aim to ensure a fair and consistent comparison for investigating the effects of noise parameters, architecture, and task on generalization \cite{neyshabur2015searchrealinductivebias}. The input images were resized from $241\times1001$ to $224\times224$ and standardized after applying noise.  To assess the influence of data size on network training, we evaluated different training and validation splits, ranging from $20-80\%$. The results indicate that the training process remains stable when the data split exceeds $40\%$. Consequently, the synthetic images were divided into training, validation, and testing sets using an $80/10/10\%$ split, respectively. During training, the noise generator parameters (noise type and scale $\beta$) are kept fixed. Training was conducted on a single Nvidia A100 GPU. A full description of the data preprocessing pipeline can be found in the \textit{Supporting Information}, Text~\silink{Text1}{S1}, and Table~\ref{T1}. 

During inference, each pre-trained network is tested on different noise profiles (single and compound) generated by the noise generator, with varying types and scales $\beta$. This process yields a robustness matrix $\mathbf{M}$, presented in Figure~\ref{f8}, which illustrates the network's performance as noise complexity and scale $\beta$ increases. The matrix includes results for both ID scenario, which are sampled from the training distribution, as well as ODD scenarios of unseen noise profiles.  The training noise parameters are indicated with a white box (Figure~\ref{f8}). In addition to evaluating the robustness matrix, we also visually assess the performance of the networks on randomly selected samples.  
\section{Results and Discussion}

In this section, we present empirical observations regarding the impact of input-source noise perturbations on three key components: 1. training, 2. ID performance, and 3. OOD generalization and robustness capabilities of neural networks. The results are organized according to the training tasks, where we evaluate the effects of noise parameters and network architectures on the three investigation pillars. 

\subsection{Training and ID Performance Analysis}\label{ID}

\begin{figure}[h] 
\noindent\includegraphics[width=\textwidth]{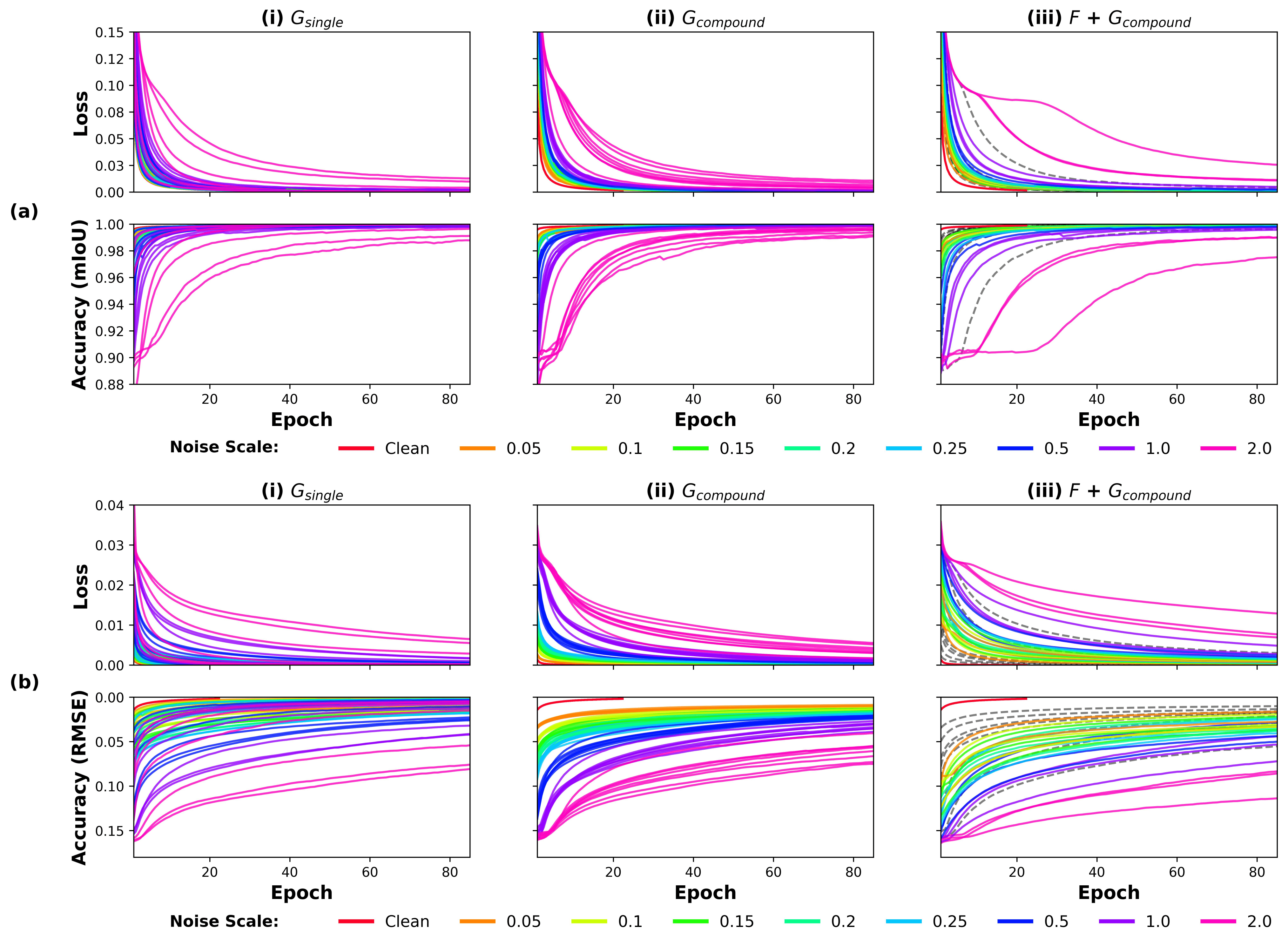} 
\caption{Validation curves of Restormer models under varying noise scales for (a) first break picking, and (b) denoising tasks. Noise was generated by (i) $G_{\text{single}}$, (ii) $G_{\text{compound}}$, and (iii) $G_{\text{compound}}$ with additional data manipulation noise from $F$. Dashed gray line in (iii) indicates performance without $F$.}
\label{f12}
\end{figure}

Analysis of the learning curves reveals consistent patterns across architectures and tasks, highlighting the intertwined influence of noise type, scale, and complexity on neural network training and performance. These parameters exert distinct effects on the learning curves, sometimes amplifying each other's impact, while at other times working in opposition to each other. At high level, noise scale $\beta$ primarily controls the shift in the learning curves (Figure~\ref{f12}), while the noise type affects the steepness. Within each noise scale group, there is a noticeable variation in the learning curves driven by noise type. The gap between the curves progressively widens with the increase in noise scale $\beta$, yet their relative relationship is preserved. In this section, we present the visual results for the Restormer model. Results for the remaining architectures are provided in the \textit{Supporting Information}, Figures~\ref{fs12}–~\ref{fs15}.\par 
 
\begin{figure}[!htb]
\noindent
    \includegraphics[width=\textwidth]{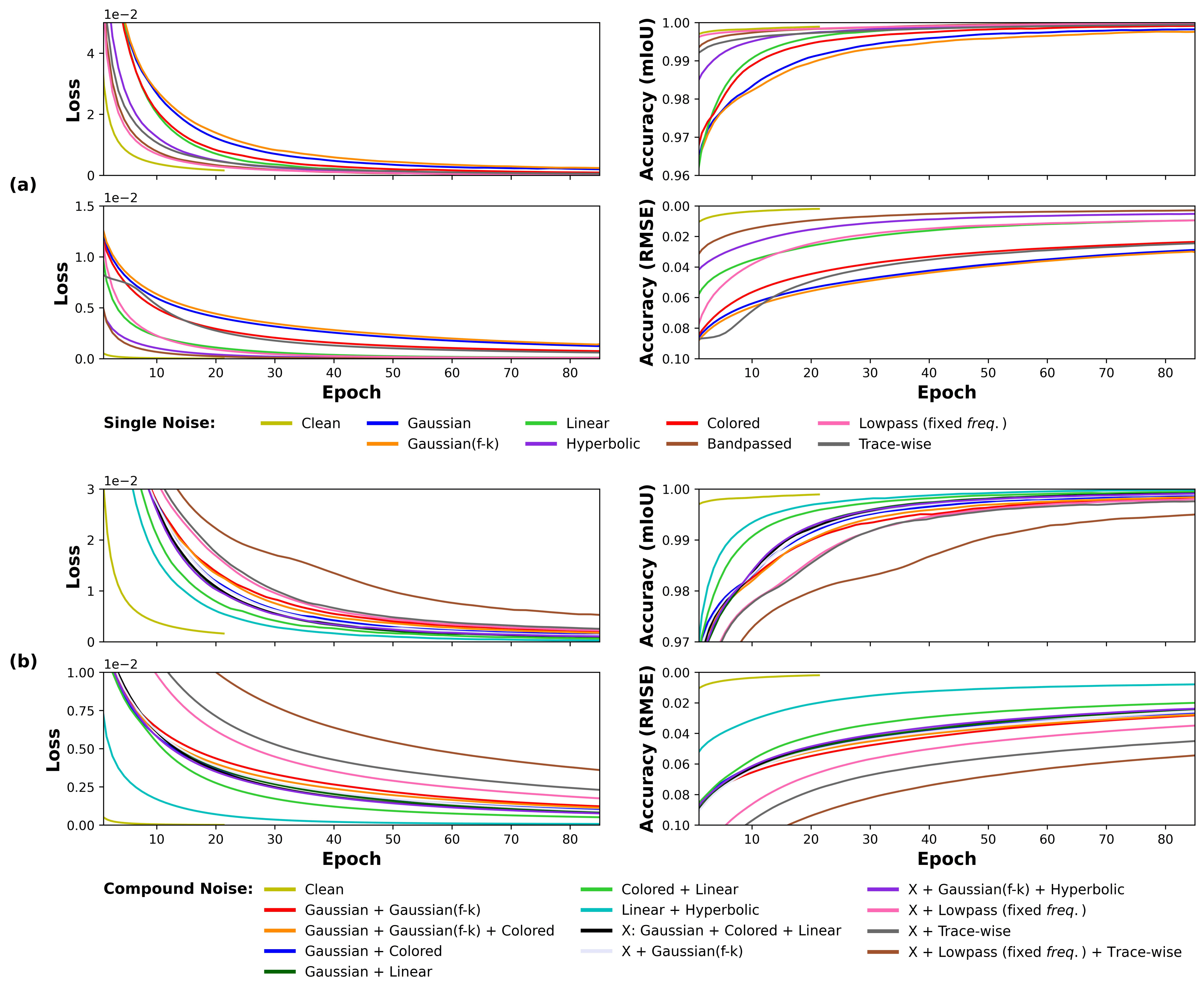} 
    \caption{Mean validation curves of Restormer models trained on (a) single and (b) compound noise types. Top panels show results for first-break picking task, and bottom panels for denoising tasks. Curves are color-coded by noise type. Left column: loss; right column: accuracy.}
    \label{f13}
\end{figure}

Figure~\ref{f13} illustrates the mean behavior of each individual noise type, averaged over noise scales. The loss convergence speed varies slightly by noise type, but they all follow a typical decay pattern. Networks trained without noise exhibit a lower initial loss due to the relative simplicity of the task and tend to plateau more rapidly compared to their noise-trained counterparts. However, it is important to note that training a network for a denoising task without introducing any noise is inherently ineffective, hence the early saturation of the loss curve (bottom panels in Figure~\ref{f13}). However, it is used as a lower-bound baseline, serving to illustrate the performance ceiling when the network is merely reproducing clean input data without the challenge of noise removal. Across all three architectures, additive linear, hyperbolic, and bandpassed noises produced by $G$ consistently promote faster loss convergence (Figure~\ref{f13}a). This suggests that the networks quickly adapt to these structured perturbations, leading to early plateaus in learning. Their limited challenge may warrant the use of additional regularization to prevent premature convergence. An exception is linear noise, which appears to support U-Net's learning process. In contrast, training with random and colored additive noise leads to a more gradual decline in loss, indicating extended optimization trajectories Figure~\ref{f13}a). This behavior suggests that these noise types may introduce task-relevant variability or complexity, which can act as an implicit regularizer, preventing premature convergence and encouraging the model to explore more robust representations across epochs. Finally, data manipulation noises introduced by $F$ (low-pass and trace-wise noise perturbations) seem to alter the training trajectory of transformer-based architectures in the denoising task. However, they do not significantly increase the complexity of the first-break picking task, as all three architectures can complete the task without relying on the amplitude and missing traces. Notably, training with variable low-pass filters introduces persistent difficulty throughout the learning process, resulting in slower convergence and inferior final performance relative to training with a fixed filter. However, this provides a more effective learning strategy to avoid overfitting to the characteristics of the low-pass filter, which can reduce generalizability to unseen filtering conditions \cite{geirhos2020generalisationhumansdeepneural,Sulun_2021}. This underscores the trade-off between accuracy and generalizability, highlighting how noise variability influences the learning dynamics of neural networks. A brief comparison of the effects of fixed and variable low-pass filters on training is provided in the \textit{Supporting Information}, Text~\silink{Text6}{S6}, and Figures~\ref{f11}–~\ref{sf50}.

Figure~\ref{f12} separates the various learning curves based on noise complexity, illustrating its influence on network training and performance. Training with compound noise (column ii) results in learning curves that show less variability across different noise types compared to those trained with single-source noise (column i), highlighting the noise scale $\beta$ as the dominant factor influencing variability in this training setup. Nonetheless, the effect of noise type remains evident, as compound noise typically mirrors the training dynamics of the constituent that most effectively facilitates learning. This is further illustrated in Figure~\ref{f13}b, which presents the mean behavior of different noise combinations averaged across noise scales. The networks trained with a random noise component exhibit similar convergence behavior to Gaussian noise, regardless of the existence of a structural or spectral noise component in the mixture. On the other hand, models trained only using structural noise tend to have a faster convergence rate that is comparable to its individual components. Similarly to single noise, the data manipulation noises imposed by $F$ change the network training trajectory (Figures~\ref{f12}-iii and~\ref{f13}b). When used in tandem, the noises generated by $F$ and $G$ increase the complexity of both tasks, compelling the network to explore more robust representations on the expense of accuracy. This observation aligns with findings in the literature that demonstrate the effectiveness of using low-pass \cite{Sulun_2021} and masking strategies to improve network training and ID performance \cite{Vinvenet, pathak2016contextencodersfeaturelearning,devlin2019, Park_2019,he2021,chen2024}.

The three network architectures in this study achieve high ID performance accuracy on the first break picking task, but show variable performance on the denoising task. The first break picking performance saturates around epochs 50-60, whereas denoising continues to improve beyond 100 epoch, reflecting the task's sensitivity to noise variability. Both Restormer and U-Net maintain strong performance across the noise scales, while Swin-U's performance declines at higher noise levels. We speculate that the high level of noise might disrupt the attention mechanism's ability to capture task-relevant features effectively. The degradation in Swin-U performance at high noise scale can be mitigated with a pre-training strategy discussed in the \textit{Supporting Information}, Text ~\silink{Text7}{S7}, and Figures~\ref{sf2}–~\ref{sf3}. In the context of denoising, Restormer, being purpose-built for denoising, consistently delivers the best performance across noise types and scales, with Swin-U performing second best. In contrast, the U-Net architecture struggles to accurately reconstruct signal amplitudes under noisy training conditions. The complexity and non-stationary nature of the generated noise present a significant challenge for U-Net at this capacity, across noise types and scales. Tabulated results can be found in \textit{Supporting Information}, Table~\ref{T3}.

\subsection{Generalization and Robustness}
\begin{figure}[h]
\noindent
    \includegraphics[width=\textwidth]{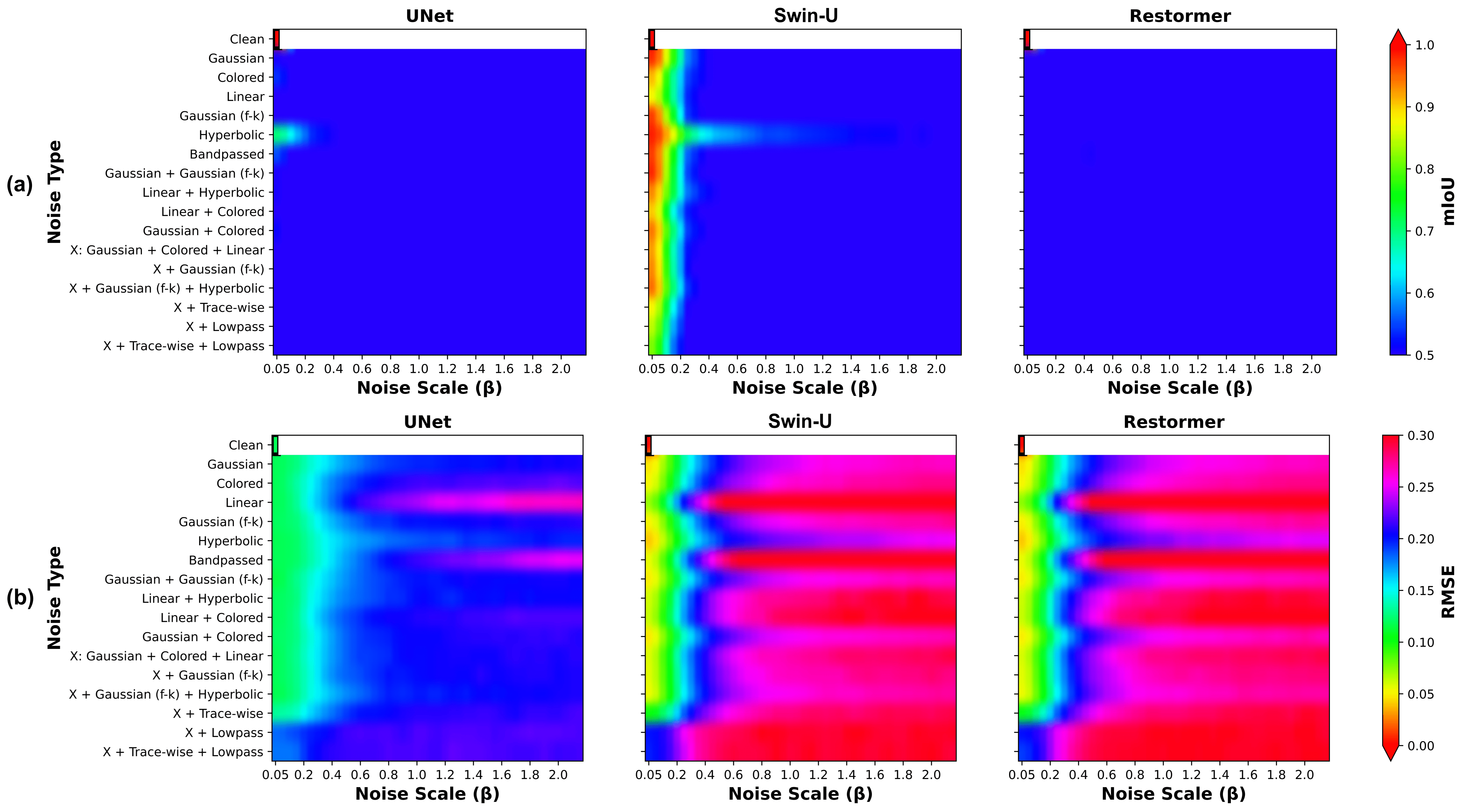} 
    \caption{Initial robustness matrix $\mathbf{M}_{0}$, showing performance of the three architectures trained under clean (noise-free) conditions. (a) First-break picking; (b) denoising.}
    \label{f14}
\end{figure}

In this section we evaluate the ID generalization and OOD robustness capabilities of networks trained under various noise conditions. To achieve this, we train each network on a specific noise type and scale, then evaluate its performance across a range of unseen noise conditions. This results in a robustness matrix $\mathbf{M}$ for each network (Figure~\ref{f14}), which quantifies performance when exposed to both more complex and higher amplitude noise, as well as simpler and lower amplitude noise sampled from the various noise generators described in Section~\ref{NG}. 
A selected subset of results is shown in the main text. Complete robustness matrices, tabulated average generalization scores, and additional visualizations are available in \textit{Supporting Information}, Figures~\ref{f17}–~\ref{f36};~\ref{f20_22x}-~\ref{40s}, and Tables~\ref{T4}-~\ref{T5}.

\begin{figure}[h]
\noindent
    \includegraphics[width=\textwidth]{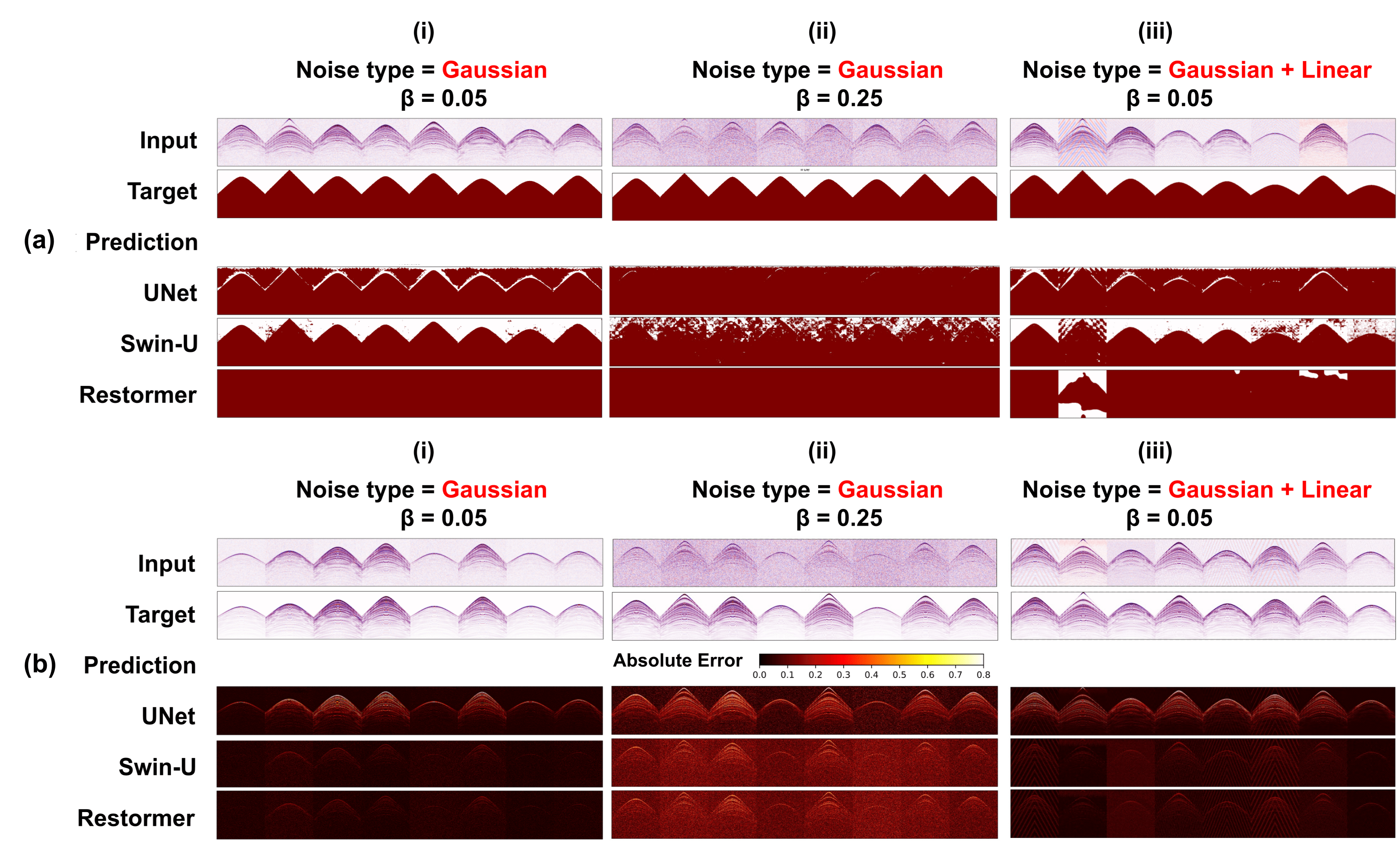} 
    \caption{Predictions from networks trained on clean (noise-free) conditions (black box in Figure~\ref{f14}) when tested under (i) low-scale Gaussian noise, (ii) medium-scale Gaussian noise, and (iii) low-scale compound noise. (a) First break picking; (b) denoising.}
    \label{f15}
\end{figure}

Figures~\ref{f14} and~\ref{f15} illustrate the baseline performance ($\mathbf{M}_{0}$) of the three architectures trained under clean (noise-free) conditions for 30 Epochs. In Figure~\ref{f14}, the noise complexity increases progressively from top to bottom, transitioning from clean inputs to single-source noise and finally to compound noise, while the noise scale $\beta$ increases from left to right. The box in Figure~\ref{f14} denotes the training noise parameters (type and $\beta$); the remainder of the figure is generated through inference. Under clean training conditions, the Swin-U architecture demonstrates inherent robustness to certain types of noise at low intensities and retains performance on the segmentation task (Figure~\ref{f14}a). In contrast, the U-Net and Restormer architectures show low robustness to noise, failing to perform without prior knowledge of noise, even at low intensities. Figure~\ref{f15}a presents visual results of the clean model performance under various noise conditions. At just $5\%$ noise, both Restormer and U-Net fail to complete the first break picking task, whereas Swin-U remains effective at that scale regardless of noise complexity. In the context of denoising (Figure~\ref{f14}b), both Transformer-based architectures, Swin-U and Restormer, show similar robustness profiles. They maintain moderate resilience to certain noise types at low scales but experience a sharp decline in performance as noise intensity increases. In contrast, U-Net exhibits higher error under clean and low-scale noise but degrades more gradually with increasing noise levels. While architectural modifications (e.g., GELU activations \cite{hendrycks2023gaussianerrorlinearunits}) improve U‐Net performance on the denoising task, we retain the original configuration to reveal this task misalignment. At $5\%$ noise, Swin-U and Restormer remain effective against Gaussian noise but struggle with complex or structured distortions (column (iii) in Figure~\ref{f15}a) and with higher-scale Gaussian noise (column (ii) in Figure~\ref{f15}b). Although U-Net has difficulty preserving signal amplitude fidelity, its features are more effective at suppressing high-scale noise components (column (iii) in Figure~\ref{f15}b). Notably, all three architectures demonstrate robustness to low-scale data manipulations in segmentation (i.e. low-pass filtering and missing traces), while Transformer-based models exhibit comparable robustness in denoising. However, this resilience diminishes when the manipulations are combined with additive noise. To assess the impact of input-source noise on model robustness, we compare performance relative to these baseline behaviors.
\subsubsection{Noise Type and Scale}

\begin{figure}[h!]
\centering
    \includegraphics[width=\textwidth,height=0.9\textheight, keepaspectratio]{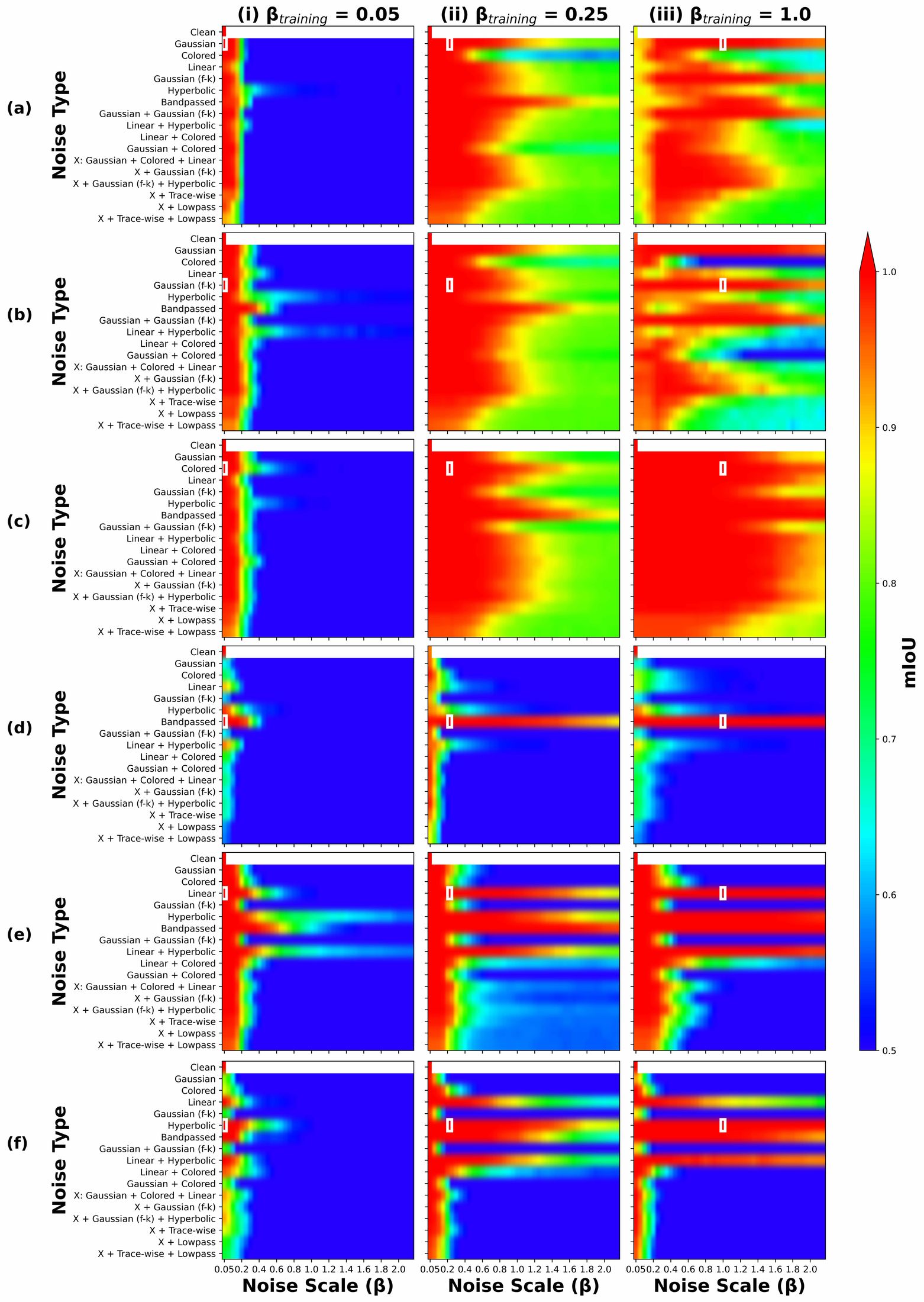} 
    \caption{Robustness matrices for the U-Net architecture trained on the first break picking task with different noise scales, $\beta_{\text{training}} \in \{0.05, 0.25, 1.0\}$, and single noise types: (a) Gaussian, (b) Gaussian (f-k), (c) Colored, (d) Bandpassed, (e) Linear, and (f) Hyperbolic. Results are computed using model weights after 50 epochs of training.}
    \label{f16}
\end{figure}

\begin{figure}[h!]
\centering
    \includegraphics[width=\textwidth,height=0.9\textheight, keepaspectratio]{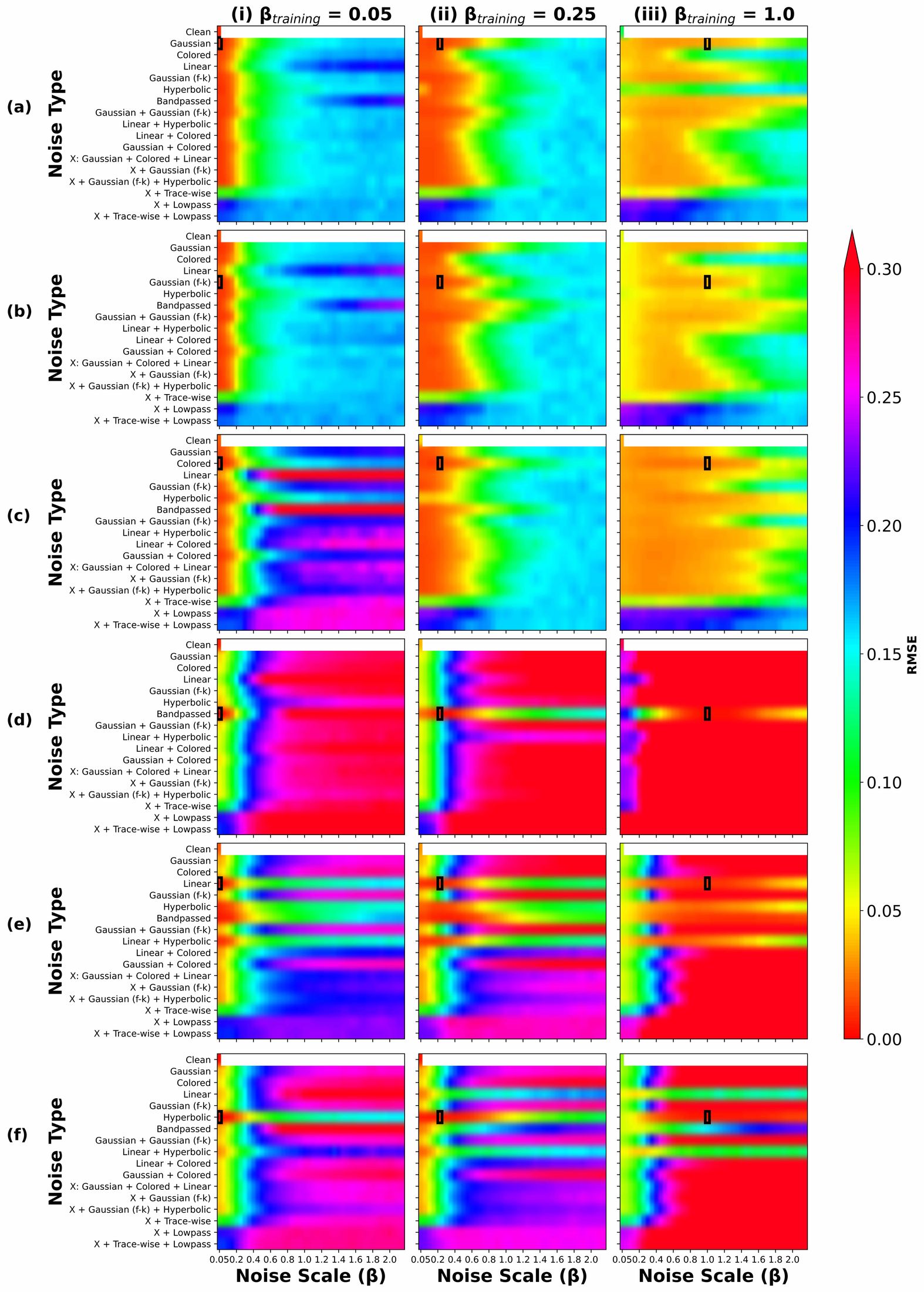} 
    \caption{Robustness matrices for the Restormer architecture trained on the denoising task with different noise scales, $\beta_{\text{training}} \in \{0.05, 0.25, 1.0\}$, and single noise types: (a) Gaussian, (b) Gaussian (f-k), (c) Colored, (d) Bandpassed, (e) Linear, and (f) Hyperbolic. Each model was trained for 100 epochs.}
    \label{f21}
\end{figure}

Training with a small amount of noise leads to immediate improvements in generalization; however, both the type and scale of noise significantly influence performance across different architectures and tasks. Figures~\ref{f16} and~\ref{f21} illustrate the impact of training with noise at varying scales ($\beta_{\text{low}}, \beta_{\text{medium}}, \beta_{\text{high}}$) on the robustness of segmentation and regression tasks. In general, training with noise improves robustness to a variety of single noise types at the same scale level ($\beta_{level}$), compared to their baseline performance ($\mathbf{M}_{0}$ shown in Figure~\ref{f14}). However, different patterns of generalization emerge, which are task- or architecture-specific, showing a preference for training with certain types of single-source noise. These results suggest that the influence of noise is not uniform across networks and tasks.  
\paragraph{Low noise level}

\begin{figure}[h!]
\centering
    \includegraphics[width=\textwidth,height=0.8\textheight, keepaspectratio]{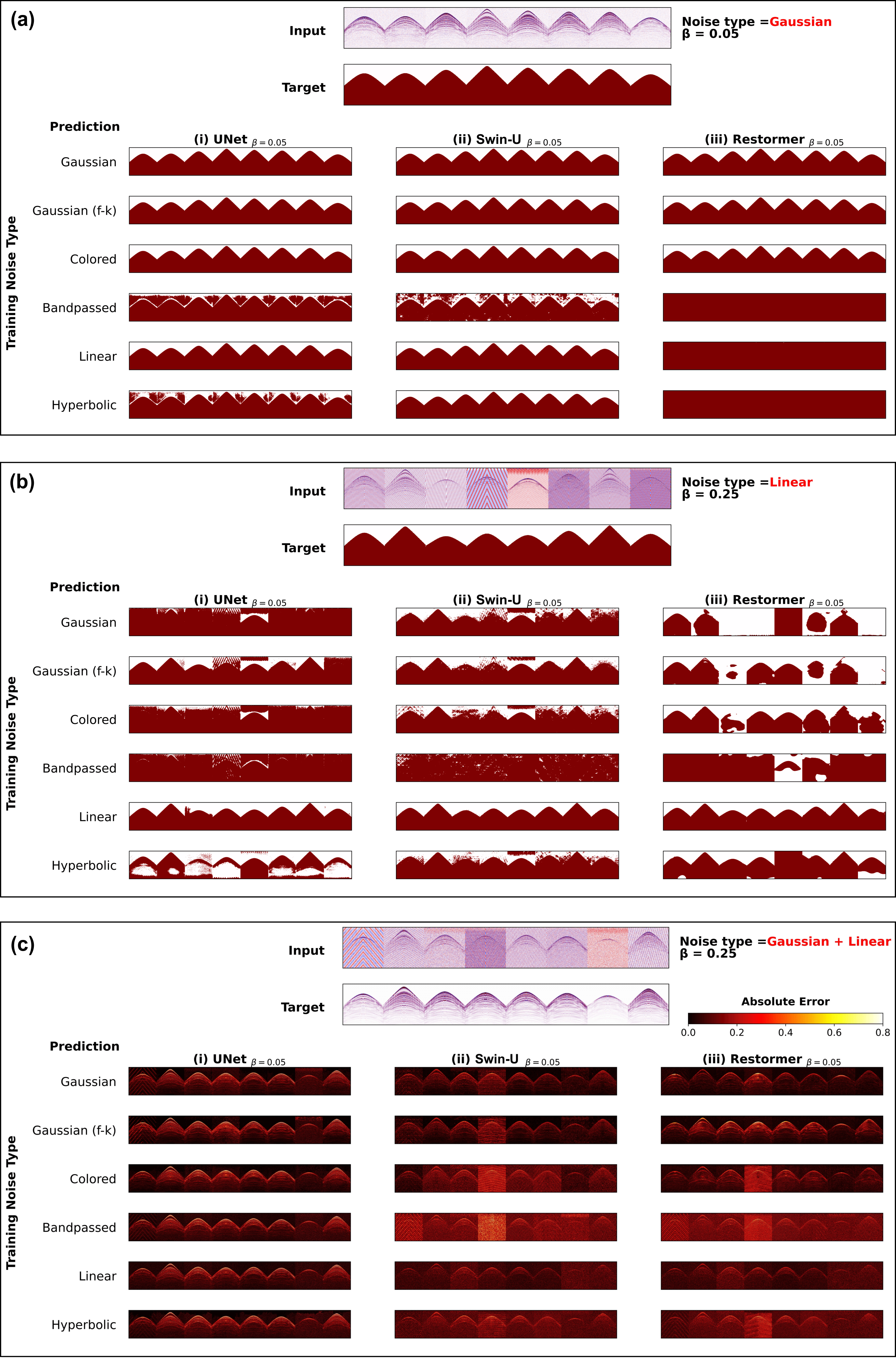} 
    \caption{Impact of low-scale ($\beta = 0.05$) single noise training on model predictions (test images). (a) First break picking task - single Gaussian noise at the same training scale; (b) first break picking task - single linear noise at a higher scale; (c) denoising task - compound noise at a higher scale.}
    \label{f16_19}
\end{figure}

Training with low-scale noise ($\beta_{0.05}$) consistently improves performance across architectures, with all models showing gains over their clean baselines ($\mathbf{M}_{0}$). Improvements are especially pronounced in segmentation, where transformer-based architectures like Swin-U achieve up to $30\%$ OOD generalization gains, reaching a mean IoU of $65\%$. These gains extend across both single and compound noise types, as long as distortions remain within the same low-scale regime. In comparison, the denoising task exhibit limited OOD generalization due to its higher difficulty. While Restormer and Swin-U still improve by $25\%$ and $18\%$ respectively, their performance in this setting is more sensitive to the structure and statistics of the training noise (column (i) in Figure~\ref{f21}).  When relying on activation functions that restrict negative gradient flow (i.e., ReLU), U-Net shows a more modest $8\%$ average gain in performance from its baseline on the denoising task. Although effective in removing noise at scales similar to those seen during training, it tends to attenuate signal amplitude, particularly under high noise levels or compound distortions outside its training distribution (Figure~\ref{f16_19}c). This limitation likely stems from its reliance on rigid, detail-heavy feature maps, which are less adaptable in high-precision, pixel-level tasks. 

Models generalize well within their noise-scale group, showing robustness to different noise types of equal or lower intensity (Figure~\ref{f16_19}a). However, performance degrades when tested on higher-scale distortions, highlighting a limitation in scale-level generalization (Figure~\ref{f16_19}, b and c). In the segmentation task, Restormer, in particular, demonstrates pronounced sensitivity to low-scale noise with repeated spatial patterns, likely due to its local attention mechanism, leading to performance drops even within the same scale group (Figure~\ref{f16_19}a). Beyond architectural differences, variability also exists within noise types themselves. Some noise types support broader generalization across test conditions, while others primarily enhance ID performance but contribute little to OOD robustness.

Random and colored noise consistently yield the most transferable features across both tasks and architectures, promoting strong OOD generalization without compromising clean-data performance. Among random noise types, low scale Gaussian (f–k) provides higher gains in segmentation (Swin-U: $61\%$, U-Net: $31\%$), whereas in regression, Gaussian (x–t) achieves a better trade-off between denoising and signal fidelity, resulting in the highest robustness improvements (Restormer: $46\%$, Swin-U: $31\%$). In contrast, structured noises tend to enhance ID generalization but have limited impact on OOD performance. Notably, bandpassed and hyperbolic noises generalize poorly beyond their training conditions, offering little improvement over clean baselines and failing to support robustness to other distortions (Figure~\ref{f16_19}, b and c).

Linear noise represents a notable exception, yielding strong ID performance alongside measurable gains in OOD generalization, even when evaluated at test scales slightly beyond those seen during training. Visual results (Figure~\ref{f16_19}, b and c) illustrate that models trained on low-scale linear noise, while vulnerable to high-scale random distortions, still perform effectively when structured components are present in compound noise mixtures, indicating that such features can remain informative under more challenging conditions. 

\begin{figure}[hb!]
\centering
    \includegraphics[width=\textwidth,height=0.8\textheight, keepaspectratio]{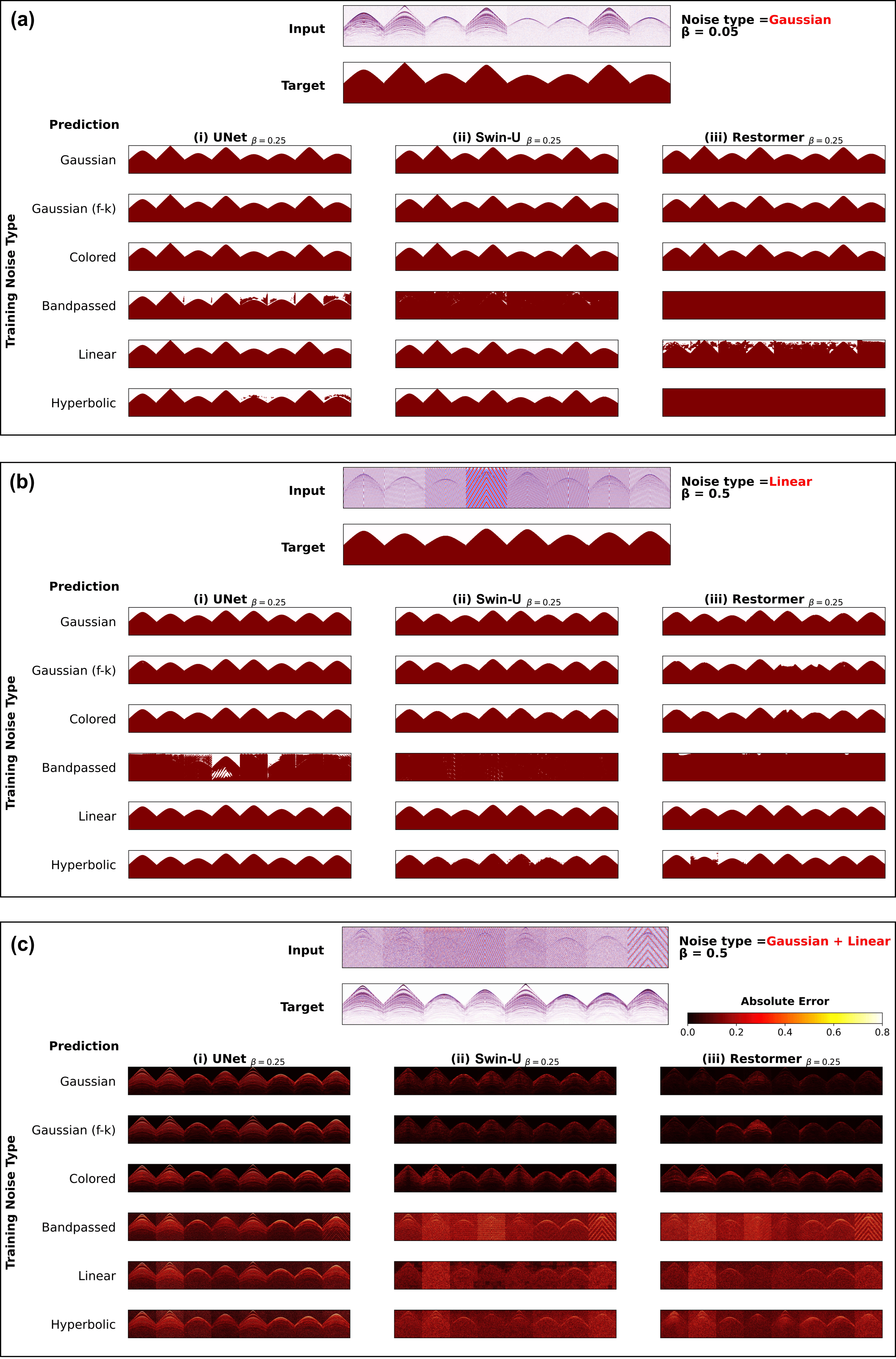}
    \caption{Impact of intermediate-scale ($\beta = 0.25$) single noise training on model predictions (test images). (a) First break picking task - single Gaussian noise at a lower scale; (b) first break picking task - single linear noise at a higher scale; (c) denoising task - compound noise at a higher scale.}
    \label{f24}
\end{figure}

Overall, low-scale noise training enhances generalization across similar noise conditions, but broader robustness is constrained by task complexity, architectural design, and the match between training and test distortions. At this scale, random and colored noise offer the most consistent OOD generalization gains, while highly structured or narrow-spectrum distortions, such as bandpassed and hyperbolic noise, contribute primarily to ID generalization. Features learned at low scale from these structured noises are often sufficient to generalize to higher-scale instances of the same distortion. An exception is linear noise, whose OOD generalization gains can match or even rival those of random and colored noise, particularly in the segmentation task (column (i) in Figure~\ref{f16}e). Nonetheless, training solely at low noise scales does not fully address sensitivity to scale shifts; generalization often breaks down under higher-scale perturbations, indicating that the learned representations remain heavily reliant on visible, low-level features rather than robust, noise-invariant abstractions.
\paragraph{Intermediate noise level}
Training with intermediate noise levels ($\beta = 0.25$) yields broader and more consistent improvements across both segmentation and regression tasks. All architectures improve over both their clean baselines and low-noise ($\beta = 0.05$) counterparts, with average relative gains of $28\%$ in segmentation and $12\%$ in regression. At this noise level, OOD generalization begins to converge across architectures, extending across different noise types at the same scale and slightly higher levels, while preserving strong performance on lower-scale distortions (column (ii) in Figure~\ref{f16}). Although Swin-U remains the top performer in segmentation, Restormer and U-Net achieve higher relative improvements at $\beta = 0.25$, $78\%$ and $89\%$ respectively, substantially narrowing the performance gap observed at lower training scales (panels (a-b) in Figures~\ref{f16_19} and~\ref{f24}).

In regression, while the gains are more modest, training with intermediate noise still enhances robustness over low-noise training (column (ii) in Figure~\ref{f21}). Transformer-based architectures show an average relative improvement of $34\%$, with Restormer maintaining strong performance, particularly when faint signal components remain detectable ($\beta < 0.8$) (Figure~\ref{f24}c). By contrast, models trained at $\beta = 0.05$ perform well only when the signal is clearly visible (Figure~\ref{f16_19}c). U-Net shows slight improvements in amplitude fidelity at higher noise levels but still lags behind transformer architectures in denoising. Training at higher scales (e.g., $\beta = 0.5$) further extends the OOD generalization fronts to more severe noise levels ($\beta = 1.0$). Beyond improved robustness to higher noise levels, models trained at $\beta = 0.25$ also outperform their low-noise counterparts when evaluated on the same lower-scale inputs, demonstrating effective scale-reversal (Figures~\ref{f16_19}a and~\ref{f24}a). This suggests that intermediate-scale training encourages the learning of more robust, noise-invariant representations by exposing models to greater input variability. 

Despite the broader OOD generalization benefits observed at intermediate noise scales, new vulnerabilities begin to emerge. These limitations, largely absent at lower noise scales, manifest as gaps in generalization that are architecture-, task-, and noise-type-specific. At this scale, random and colored noises remain the most effective, supporting OOD generalization across noise types and scales. However, they exhibit opposing generalization gaps at higher intensities, highlighting their complementary characteristics. In Restormer, improvements are largely confined to these two noise types; however, it remains vulnerable to hyperbolic noise across scales and exhibits noticeable performance degradation on clean data when trained with Gaussian (f–k) or colored noise (column (ii) in Figure~\ref{f21}). Consequently, selecting the appropriate noise type involves balancing clean-data fidelity against the goal of enhancing robustness to unseen conditions.

The OOD generalization behavior of structured and bandpassed noise remains limited, primarily enhancing ID performance while failing to improve robustness to unstructured distortions. Although these noise types become increasingly visually disruptive at higher scales, occluding key features and raising task difficulty, their impact does not scale accordingly. Their limited effectiveness appears largely independent of training noise intensity. While linear noise yields modest gains in segmentation, it contributes little to OOD robustness. Its utility is limited mainly to generalizing across similar or lower-scale distortions (Figure~\ref{f24}, a and b), or to higher-scale mixtures containing a structured component, although the latter does not apply in regression (Figure~\ref{f24}c). Other noise types typically generalize better to strong linear distortions. Hyperbolic noise yields small gains for Swin-U in segmentation but fails to generalize in regression.

In summary, training with intermediate-scale noise improves generalization across both tasks and scales, outperforming low-noise training in nearly all cases. Random and colored noise remain the most effective, accounting for $92\%$ of OOD generalization gains, while linear noise contributes modestly in segmentation. While intermediate-scale training improves robustness, it also introduces vulnerabilities driven by increased sensitivity to noise type and structure.
\paragraph{High noise level}
Training with high-amplitude noise ($\beta_{high}$) yields only marginal generalization improvements beyond those achieved at intermediate scales, primarily for random and colored noise. These gains are inconsistent across architectures and tasks, and are typically limited to generalization toward higher-scale distortions. As noise begins to dominate the input, models face increased difficulty extracting weak signals, exposing new vulnerabilities and disrupting prior generalization patterns (column (iii) in Figures~\ref{f16} and~\ref{f21}).

\begin{figure}[h!]
\centering
    \includegraphics[width=\textwidth,height=0.8\textheight, keepaspectratio]{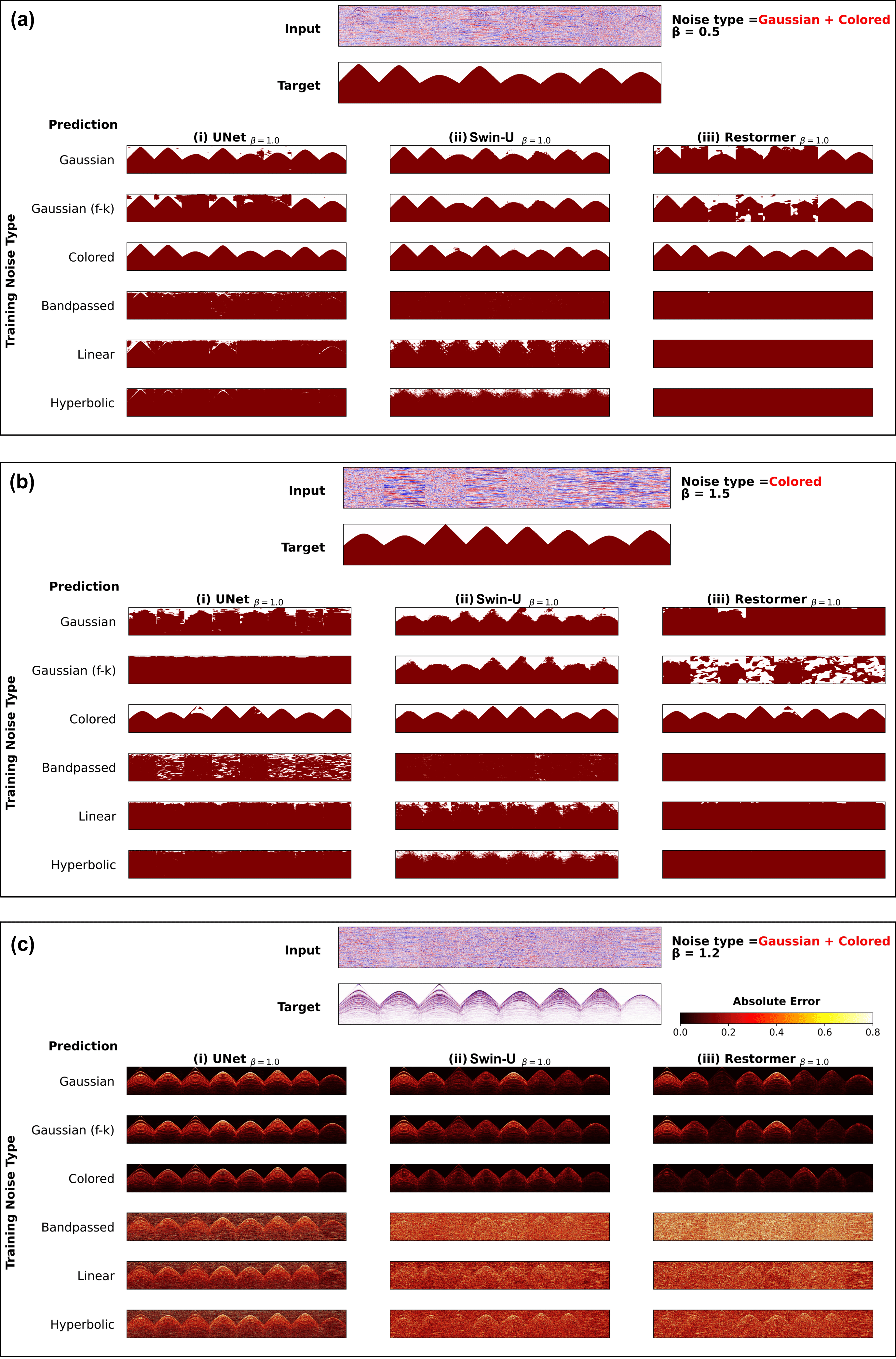}
    \caption{Impact of high-scale ($\beta = 1.0$) single noise training on model predictions (test images). (a) First break picking task - compound (Gaussian + colored) noise at a lower scale; (b) first break picking task - single colored noise at a higher scale; (c) denoising task - compound (Gaussian + colored) noise at a higher scale.}
    \label{f28}
\end{figure}

In segmentation, Swin-U remains the most stable, improving OOD performance by $8\%$ from $\beta_{\text{medium}}$ and maintaining robustness across scales. U-Net and Restormer also benefit from colored noise, achieving performance comparable to Swin-U, yet exhibit increased sensitivity across scales when trained with random and structured distortions (column (iii) in Figure~\ref{f16}). U-Net, in particular, struggles to generalize under Gaussian (f–k) noise, while Restormer, despite stronger gains, remains vulnerable to hyperbolic and colored noise (column (iii) in Figure~\ref{f28}). Bandpassed and structured noise offer minimal gains beyond ID conditions at high scales. Notably, U-Net is the only architecture to show some OOD generalization when trained with linear noise, particularly under lower-scale settings that include structured components. Increasing the noise scale can promote better generalization, but excessively high values may degrade performance or trigger failure modes, even in architectures and tasks that are otherwise well-aligned. For example, although Swin-U is well-suited for segmentation, it still suffers a $3.5\%$ drop in performance when trained with higher-scale noise ($\beta > 1.0$), even with robust noise types (i.e. colored noise). To enable higher-scale training, we recommend cross-task pre-training which helps recover Swin-U’s robustness and reduces Restormer’s sensitivity to low-scale distortions. However, pre-training leads to performance degradation in U-Net, indicating limited transferability of its denoising-specific representations to segmentation. Full details are provided in the \textit{Supporting Information}, Text~\silink{Text7}{S7}. and Figures~\ref{sf2}–~\ref{sf3}.

In regression, similar limitations emerge. While most models generalize to higher scales, performance often falls short of that achieved under lower noise training (Figure~\ref{f21}iii). Restormer again stands out, maintaining high performance with colored noise ($71\%$ improvement over baseline) and showing strong generalization in compound and high intensity conditions (Figure~\ref{f28}c). Training with high-scale noise does not necessarily resolve vulnerabilities that are already present at lower scales. In fact, when such vulnerabilities exist, as in the case of random noise, higher-scale training can amplify them, leading to increased sensitivity to other distortions (e.g., colored noise) and reduced performance on clean or low-noise inputs (Figures~\ref{f21}iii and~\ref{f28}c). Although structured and bandpassed noise are ineffective at high training scales, U-Net continues to be the only architecture that can partially recover signal structure under severe noise, albeit with limited amplitude fidelity (Figure~\ref{f28}c).

Overall, high-noise training offers limited additional generalization and introduces task- and architecture-specific vulnerabilities. Robustness to low-scale noise often degrades, and failure modes intensify. Colored noise continues to yield the most transferable features, whereas structured and bandpassed noise provide minimal gains beyond $\beta_{low}$. Training solely with data manipulations (\textbf{F}) (e.g., low-pass filtering, missing traces) produces consistent trends across noise scales, closely resembling performance under clean training conditions. Despite strong ID performance, these models show limited robustness to additive (\textbf{G}) and compound distortions involving data manipulations (\textbf{G + F}), performing moderately in segmentation under low noise conditions and failing entirely in denoising. In contrast, models trained only with additive noise (\textbf{G}) generalize well to test examples containing data manipulations (\textbf{F}), but struggle with high-scale compound distortions that combine both (\textbf{G + F}), particularly in regression (Figure~\ref{f21}). High-scale noise training further amplifies these vulnerabilities to combined distortions (\textbf{G + F}), with models failing to generalize even to lower-scale variants, despite being trained with robust noise types such as colored noise.
\subsubsection{Noise Complexity}
As noted in earlier sections, training on specific single-source noise types can support generalization to more complex, compound noise scenarios. This capacity extends to higher noise amplitudes when $\beta$ is appropriately tuned. However, the effectiveness of generalization varies by noise type, with certain types either failing to scale or inducing new vulnerabilities under high-noise conditions. To address these weaknesses, we propose training with compound input-source noise. This yields a $21\%$ relative average improvement in OOD generalization across tasks, with Swin-U leading in segmentation and Restormer in denoising. Additive noise from \textbf{G} accounts for the majority of robustness gains ($81\%$ in segmentation, $71\%$ in denoising) and the rest is attributed to manipulations from \textbf{F}.

\begin{figure}[h!]
    \centering\includegraphics[width=\textwidth,height=0.9\textheight, keepaspectratio]{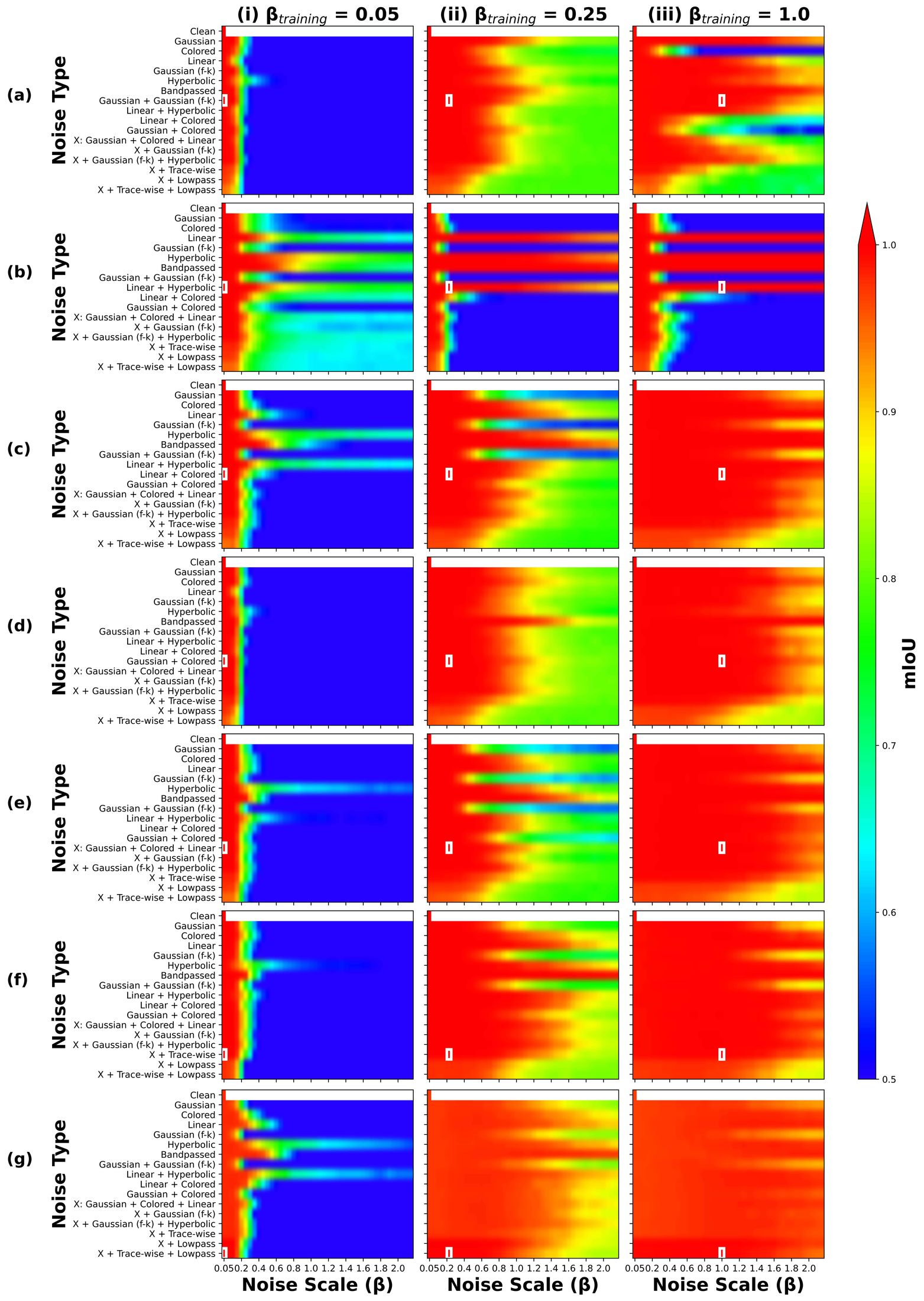} 
    \caption{Robustness matrices for the U-Net architecture trained on the first break picking task with different noise scales, $\beta_{\text{training}} \in \{0.05, 0.25, 1.0\}$, and compound noises: (a) Random, (b) Structured, (c) Linear + Colored, (d) Gaussian + Colored, (e) Gaussian + Colored + Linear, (f) e + Trace-wise (g) f + Low-pass (fixed frequency cutoff). Results are computed using model weights after 50 epochs of training.}
    \label{f32}
\end{figure}

\begin{figure}[h!]
    \centering\includegraphics[width=\textwidth,height=0.9\textheight, keepaspectratio]{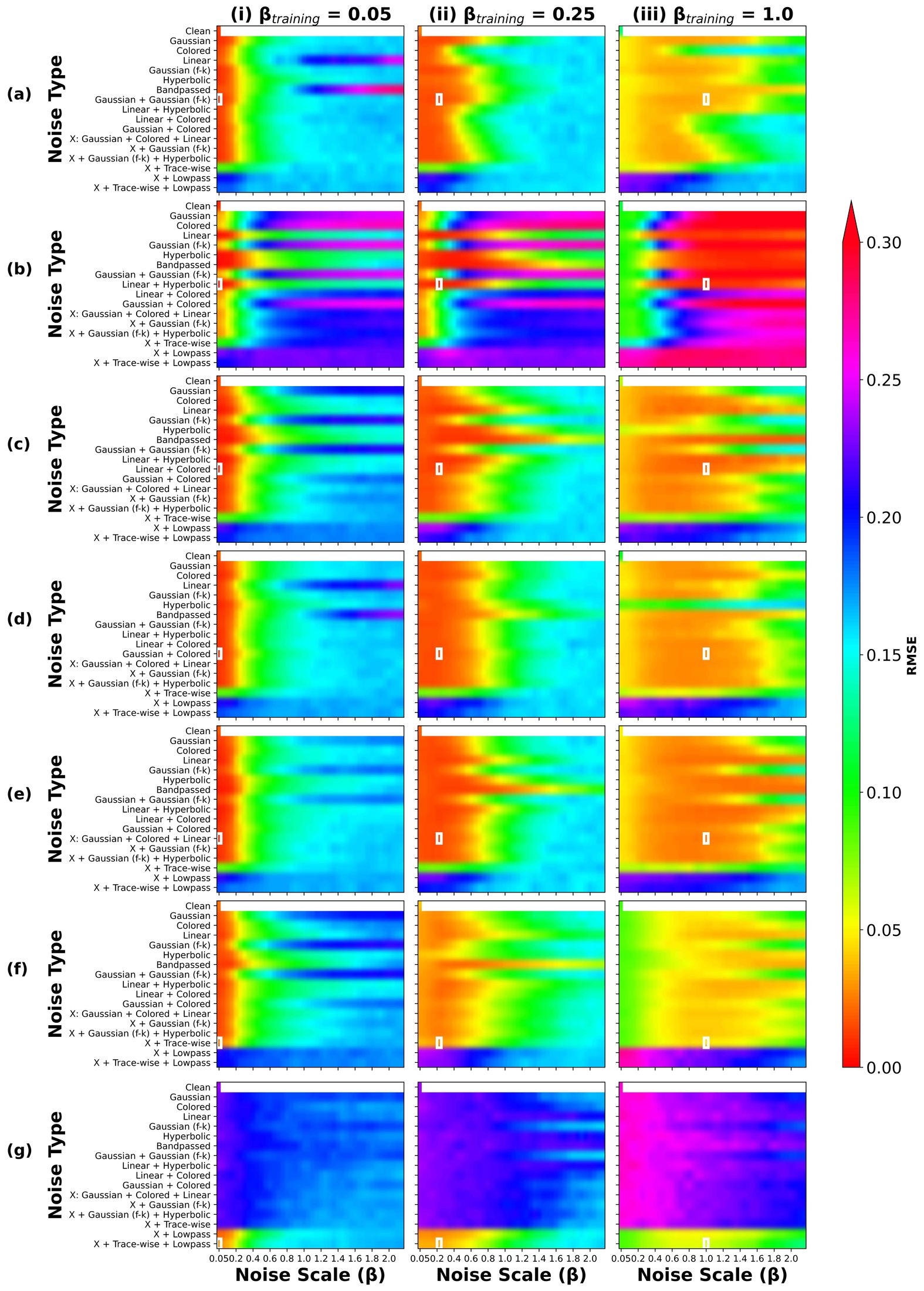} 
    \caption{Robustness matrices for the Restormer architecture trained on the denoising task with different noise scales, $\beta_{\text{training}} \in \{0.05, 0.25, 1.0\}$, and compound noises: (a) Random, (b) Structured, (c) Linear + Colored, (d) Gaussian + Colored, (e) Gaussian + Colored + Linear, (f) e + Trace-wise (g) f + Low-pass (fixed frequency cutoff). Each model Was trained for 100 epochs.}
    \label{f37}
\end{figure}

\begin{figure}[h!]
\centering
    \includegraphics[width=\textwidth,height=0.9\textheight, keepaspectratio]{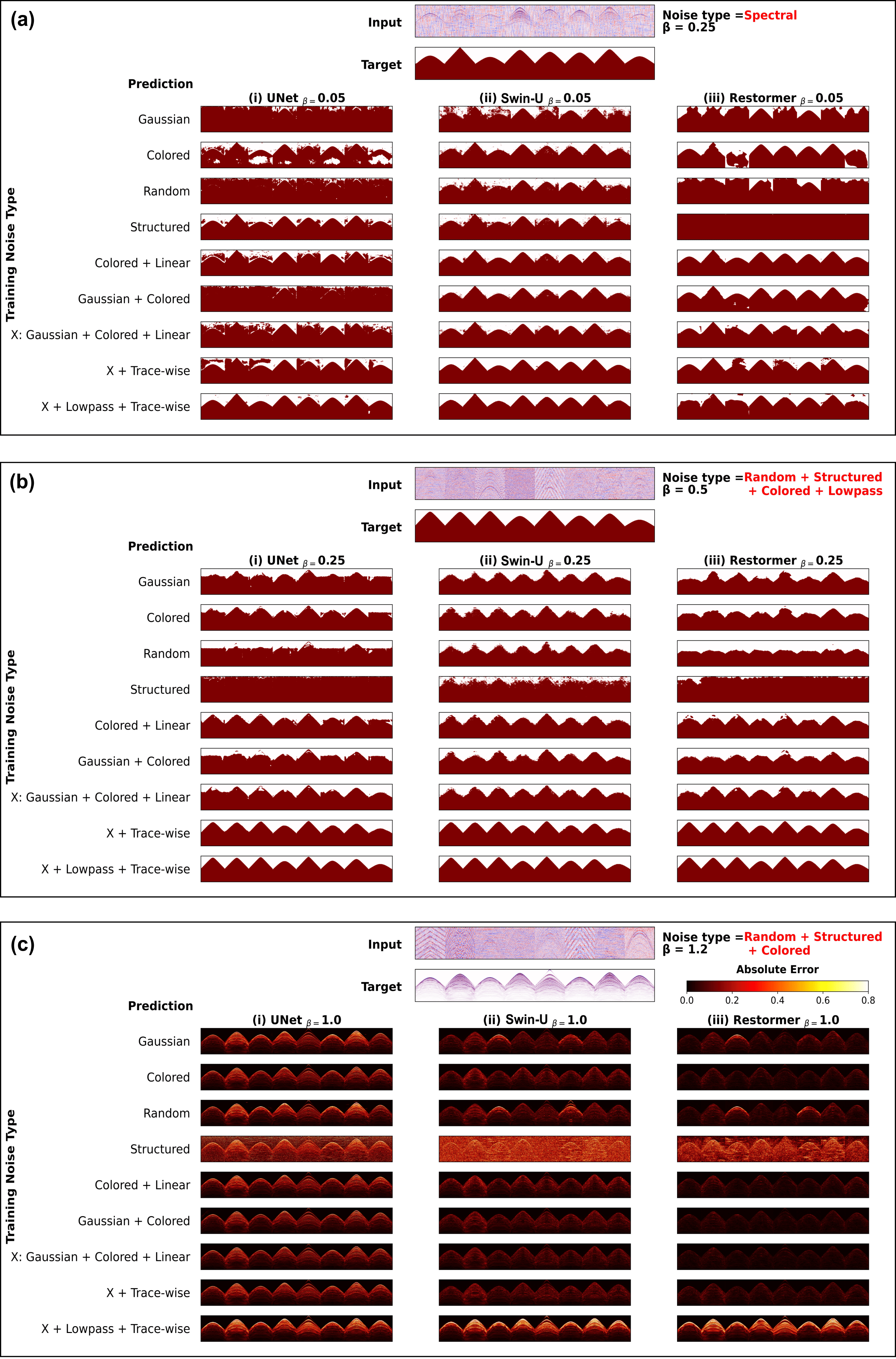}
    \caption{Impact of training with compound noise at different scales ($\beta$) on model predictions under higher-scale compound noise (test images).  (a) First break picking task - Low training scale ($\beta_{0.05}$) - additive compound noise; (b) first break picking task - Medium training scale ($\beta_{0.25}$) - additive compound noise + data manipulation; (c) denoising task - High training scale ($\beta_{1.0}$) - additive compound noise.}
    \label{f38}
\end{figure}

Figures~\ref{f32} and~\ref{f37} illustrate the effect of compound noise training on generalization ($\mathbf{M}$), with noise complexity increasing from (a) to (g) and scale ($\beta$) from low to high across columns. At $\beta_{low}$, single and compound training produce similar results, but compound noise begins to offset the weaknesses of individual noise types. It forms a generalization subspace shaped by its strongest constituent and enhances OOD robustness to higher scales in well-aligned architectures. For example, Restormer improves when low-scale random noise is combined with colored noise (Figures~\ref{f37}d and~\ref{f38}a). Although combining weak types (e.g., structured noise) offers little benefit (Figure~\ref{f37}d), such mixtures do not harm robustness when paired with stronger components (Figures~\ref{f37}c and~\ref{f38}a). Our experiments show that stacking multiple additive noise types does not necessarily guarantee OOD gains and may even exacerbate vulnerabilities when the noise mixture is imbalanced, especially in segmentation (Figure~\ref{f38}a). In contrast, combining data manipulations (e.g., amplitude masking, filtering) with additive noise increases the complexity of the segmentation task, which enhances OOD robustness (Figures~\ref{f32}, f–g and~\ref{f38}a). Robustness in denoising is more sensitive to the type of manipulation. Amplitude masking yields modest benefits when combined with additive noise, whereas filtering disrupts training dynamics, leading to reduced OOD robustness and improvements limited to ID performance (Figure~\ref{f37}, f–g). This highlights the task-specific sensitivity to these interventions.

As with single-noise training, raising $\beta$ to medium or high levels extends the OOD generalization front. Compound additive noise amplifies these effects, pushing the models to generalize to even higher scales (Figures~\ref{f32} and ~\ref{f37}). These generalization fronts improve substantially when strong noise types are combined, resulting in broader generalization profiles than those achievable by individual components alone (Figure~\ref{f38}). Our experiments indicate that in the segmentation task, training with a combination of additive noise and data manipulations (\textbf{G + F}) yields OOD robustness gains comparable to those achieved through higher-scale single-noise training (Figures~\ref{f32}, f-g ;~\ref{f38}a). This suggests that such combinations encourage models to learn transferable representations that remain effective under high noise intensities, representations that are otherwise inaccessible without explicit high-scale training. Although data manipulations enhance OOD robustness to high-scale additive noise, they yield only modest improvements in ID performance, particularly at noise scales beyond those seen during training. Nonetheless, adding data manipulations to the training noise mixture consistently enhances OOD robustness over using additive noise alone (Figure~\ref{f38}b). In denoising, the combination of masking with high-scale additive noise tends to smooth the generalization landscape, leading to a decrease in accuracy on clean or low-noise conditions (column (iii) in Figure ~\ref{f37}f). The absence of convergence in high-noise learning curves (Section~\ref{ID}), suggests further training may be required to fully adapt to such conditions.

\begin{figure}[h!]
\centering
    \includegraphics[width=\textwidth,height=0.8\textheight, keepaspectratio]{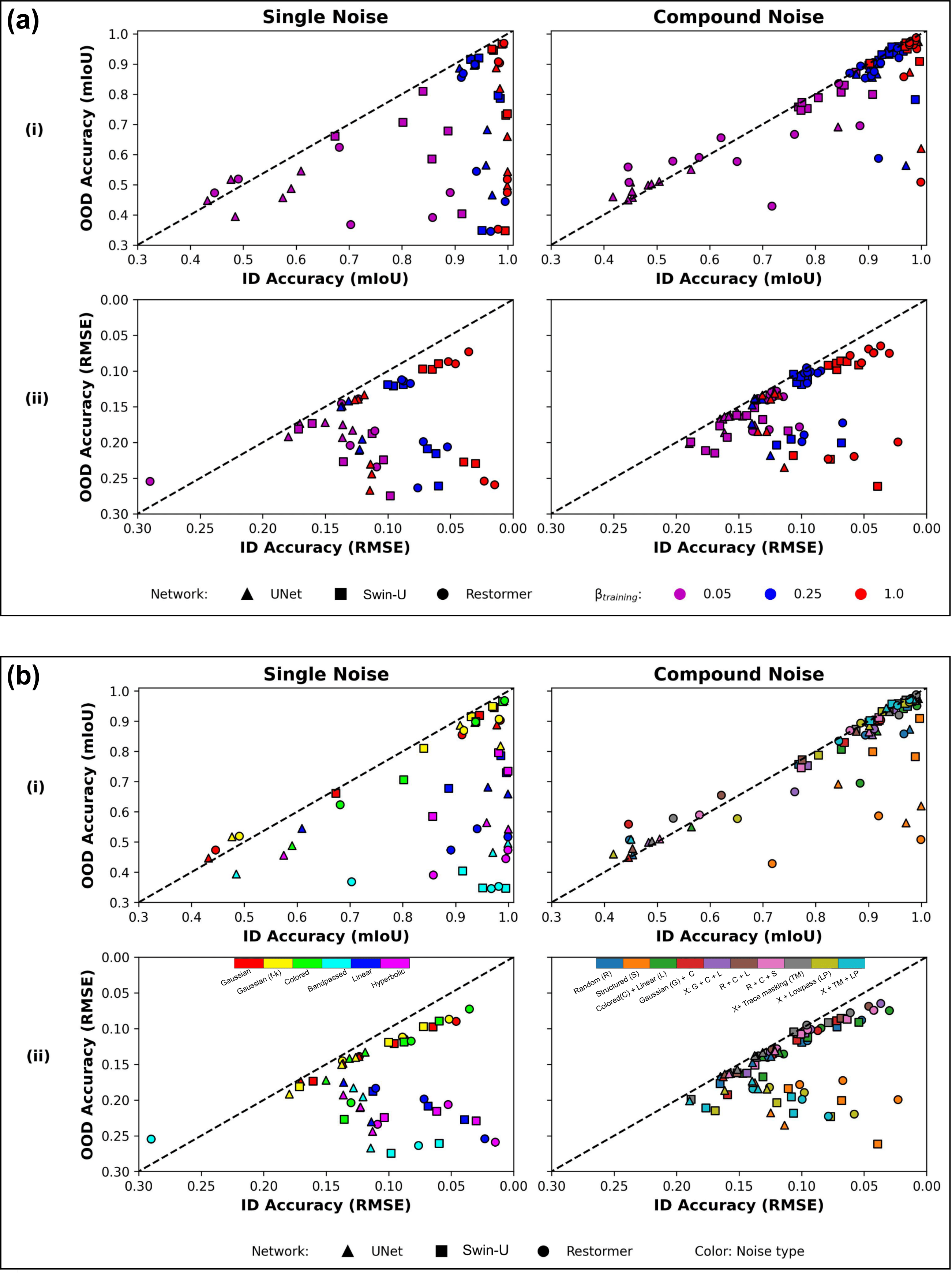}
    \caption{ID Vs. OOD performance on the (i) first break picking, and (ii) denoising tasks. The dashed line denotes the ideal 1:1 performance line. Color code: (a) noise scale, (b) noise type}
    \label{f40}
\end{figure}

A key distinction between single-source and compound noise training lies in their ID–OOD performance relationship. Although ID accuracy remains relatively stable, OOD performance exhibits significantly greater variability. When ID performance is averaged across scales, a linear relationship between ID and OOD performance emerges (Figure~\ref{f40}), with scale acting as a factor that shifts performance along this trend. Higher noise scales consistently shift models upward along the ID–OOD performance trend across tasks, architectures, and noise types (Figure~\ref{f40}a), but the variation in OOD robustness remains primarily driven by the type of noise (Figure~\ref{f40}b). Compound training improves alignment along this trend, with remaining deviations primarily driven by structured or unbalanced noise mixtures. In the segmentation task, models trained on weak single noises (e.g., structured, bandpassed) or their weak combinations (row (i) in Figure~\ref{f40}b, orange) show the greatest divergence from the expected linear relationship. In denoising, this deviation from linearity manifests as a secondary horizontal or even inverse ID–OOD relationship, indicating a breakdown in generalization (row (ii) in Figure~\ref{f40}b). As in segmentation, the strongest deviations in denoising arise from weak single noises or their compound mixtures. However, unlike segmentation, where data manipulation improves robustness, introducing filtering in denoising (row (ii) in Figure~\ref{f40}b) further amplifies the divergence from the ID–OOD trend. These effects persist across noise scales and architectures indicating severe distribution shifts. 

In summary, compound noise training supports broader and more reliable generalization by leveraging the strengths of multiple perturbations. Rather than overfitting to specific distortions, models extract stable features that generalize well across scale and type. Training with compound noise mixtures acts as a form of implicit ensembling, improving ID–OOD consistency without requiring explicit model weighting, particularly when the noise mixtures are well-balanced. However, poorly composed mixtures (e.g., weak or uncomplimentary sources) deviate from this trend. Despite these promising results, further theoretical work is needed to explain the mechanisms driving the observed robustness gains.
 
\clearpage
\section{Conclusion}

Training with input-source noise introduces a strong inductive bias that directly influences a model’s OOD generalization capacity. Specifically, the noise scale ($\beta$) acts as an implicit regularizer, encouraging the learning of noise-robust and transferable representations. This, in turn, extends the model’s generalization front to both lower and higher noise levels. For example, models trained with the noise scale $\beta = 0.25$ generalize effectively to test scenarios with noise levels up to twice the training scale, outperforming models trained under clean or low noise conditions by an average of $25\%$ ($12.2\%$ in denoising and $38.3\%$ in segmentation).

Despite these gains, OOD generalization remains highly dependent on the type of noise used during training, and its alignment with the network architecture and task objective. Our experiments suggest that training with certain single-source noise types (e.g., colored or Gaussian) enables transfer to more complex, compound noise settings, maintaining performance within $\Delta_{\text{mean}} = 6.4\%$ of the best-case scenario. In contrast, other noise types (e.g., bandpassed or structured) exhibit limited transferability, with performance degrading on average by $68.5\%$ under mismatched test conditions, offering little benefit beyond the specific scenarios they were trained for.

While tuning the training noise scale can improve generalization, it can also introduce scale-induced vulnerabilities, even within otherwise robust single noise types. This challenge becomes especially critical in real-world scenarios, such as active testing environments, where noise characteristics are often unknown or highly variable. In such settings, even with a well-chosen noise scale, a mismatch between the model architecture and the dominant noise type can lead to substantial performance degradation or outright failure. As such success depends on selecting noise types that align with the inductive biases of the architecture while accounting for the variability and operational constraints of the deployment environment.

In active monitoring settings, consistency in the processing workflow is critical for reliable baseline–monitoring comparisons. However, such consistency is difficult to maintain when noise is nonstationary and non-repetitive, as is common in field deployments. In SWD, this challenge is further compounded by the progressive weakening of the bit signal with depth, which can make first-break picking highly uncertain at larger offsets. because these conditions cannot be modeled reliably, it is generally more practical to design workflows that learn to suppress or accommodate such variability rather than attempting to explicitly model individual noise mechanisms. In this context, noise-diverse training provides a useful means of stabilizing model behavior across surveys, reducing sensitivity to near-surface noise fluctuations that would otherwise mask or mimic true subsurface changes.

To address these challenges, we propose a compound noise training strategy that incorporates diverse, complementary noise types during training. By exposing networks to a broader family of perturbations, compound noise training improves stability under shifting noise distributions and reduces sensitivity to noise fluctuations that could affect processing workflows. This approach yields significantly improved ID–OOD alignment, reducing performance variance across test conditions by an average of $72.8\%$, and increasing mean OOD performance by $21\%$ relative to single-noise baselines. These improvements stem from a shift in learning dynamics, encouraging models to focus on invariant structural features rather than overfitting to specific noise distributions. By introducing an additional layer of regularization, compound noise training mitigates the risks associated with selecting a single, potentially suboptimal noise source. It also reduces sensitivity to mismatches between architecture and noise type, providing a more reliable foundation for deployment in environments with high noise variability. Ultimately, models trained with compound mixtures are better equipped to generalize under unpredictable or high-variance conditions, making them a robust choice for real-world applications. While no training strategy can guarantee complete robustness to all possible field-noise conditions, our results demonstrate that compound noise training substantially broadens the model’s practical operating range and offers a promising direction for real-world monitoring and processing workflows.

\section{Open Research Section}
The synthetic data, the stochastic noise-generation framework, training and evaluation pipelines, and codes to reproduce all visuals in this study are available on Zenodo~\cite{alsinan2025noiseattackzenodo} and are maintained at the corresponding GitHub repository~\cite{alsinan2025noiseattack}.

\section{Conflict of Interest disclosure}
The authors declare there are no conflicts of interest for this manuscript.

\section{Acknowledgments}
The authors are grateful to the reviewers for their constructive feedback, which
guided the revision of this manuscript, and acknowledge King Abdullah University of
Science and Technology (KAUST) for providing the computational resources and research support that enabled this work.

\bibliographystyle{unsrt}  
\bibliography{references}  

\newpage

\section{Supporting Information}

\begin{figure}[htb]
    \centering\includegraphics[width=0.8\textwidth,height=0.9\textheight, keepaspectratio]{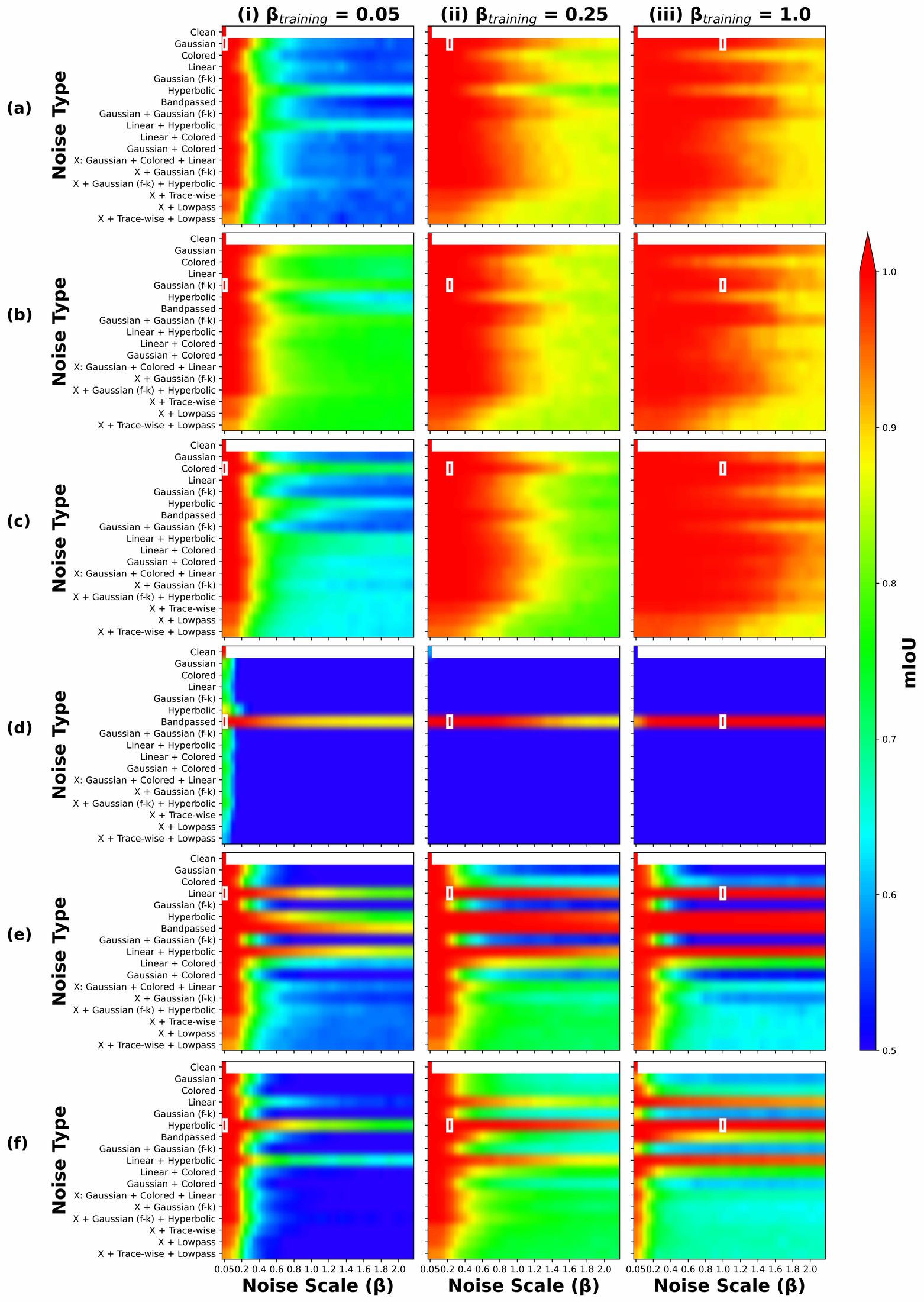} 
    \caption{Robustness matrices for the Swin-U architecture trained on the first break picking task with different noise scales, $\beta_{\text{training}} \in \{0.05, 0.25, 1.0\}$, and single noise types: (a) Gaussian, (b) Gaussian (f-k), (c) Colored, (d) Bandpassed, (e) Linear, and (f) Hyperbolic. Results are computed using model weights after 50 epochs of training.}
    \label{f17}
\end{figure}

\begin{figure}[h]
    \centering\includegraphics[width=0.8\textwidth,height=0.9\textheight, keepaspectratio]{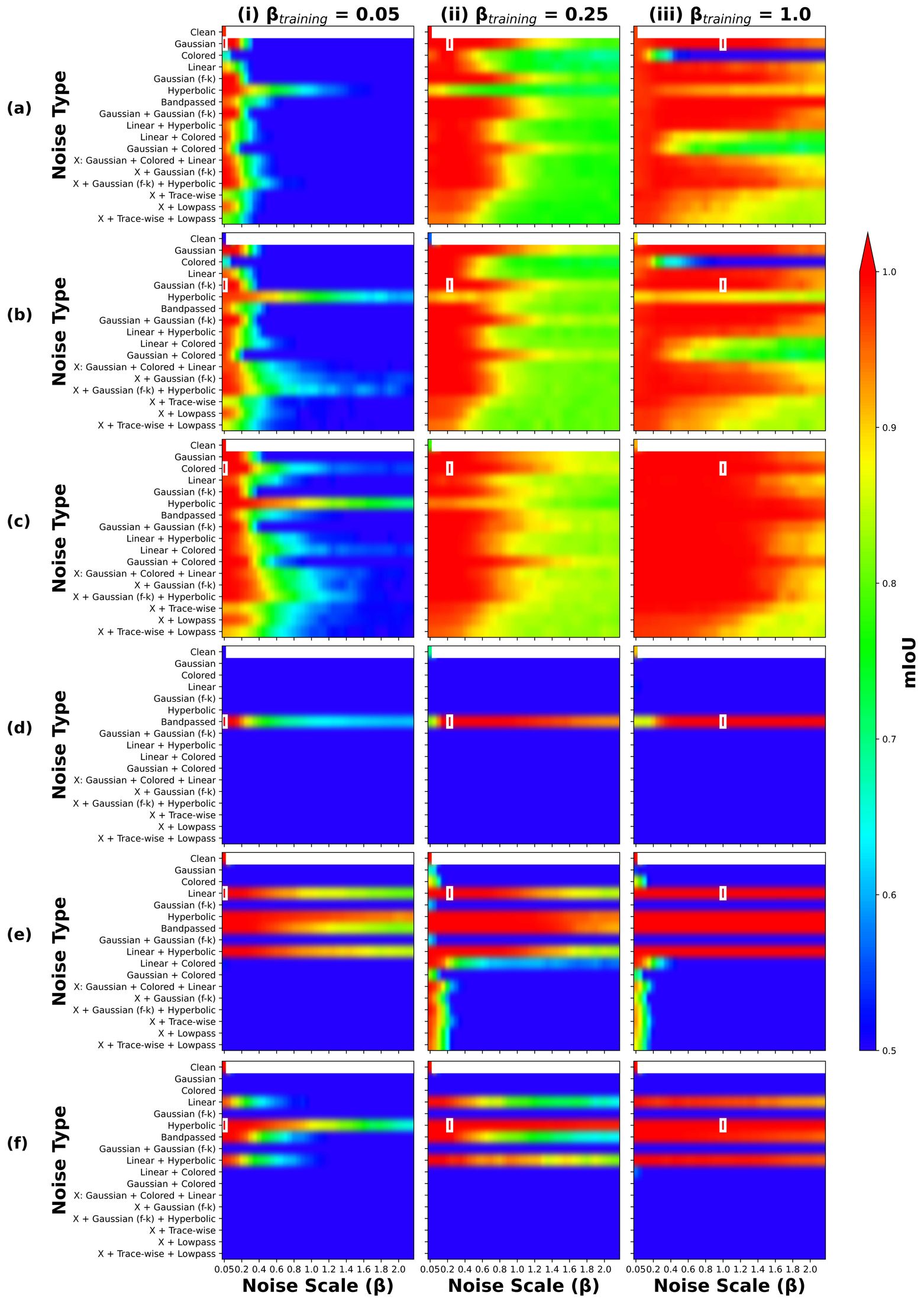} 
    \caption{Robustness matrices for the Restormer architecture trained on the first break picking task with different noise scales, $\beta_{\text{training}} \in \{0.05, 0.25, 1.0\}$, and single noise types: (a) Gaussian, (b) Gaussian (f-k), (c) Colored, (d) Bandpassed, (e) Linear, and (f) Hyperbolic. Results are computed using model weights after 50 epochs of training.}
    \label{f18}
\end{figure}

\begin{figure}[h]
    \centering\includegraphics[width=0.8\textwidth,height=0.9\textheight, keepaspectratio]{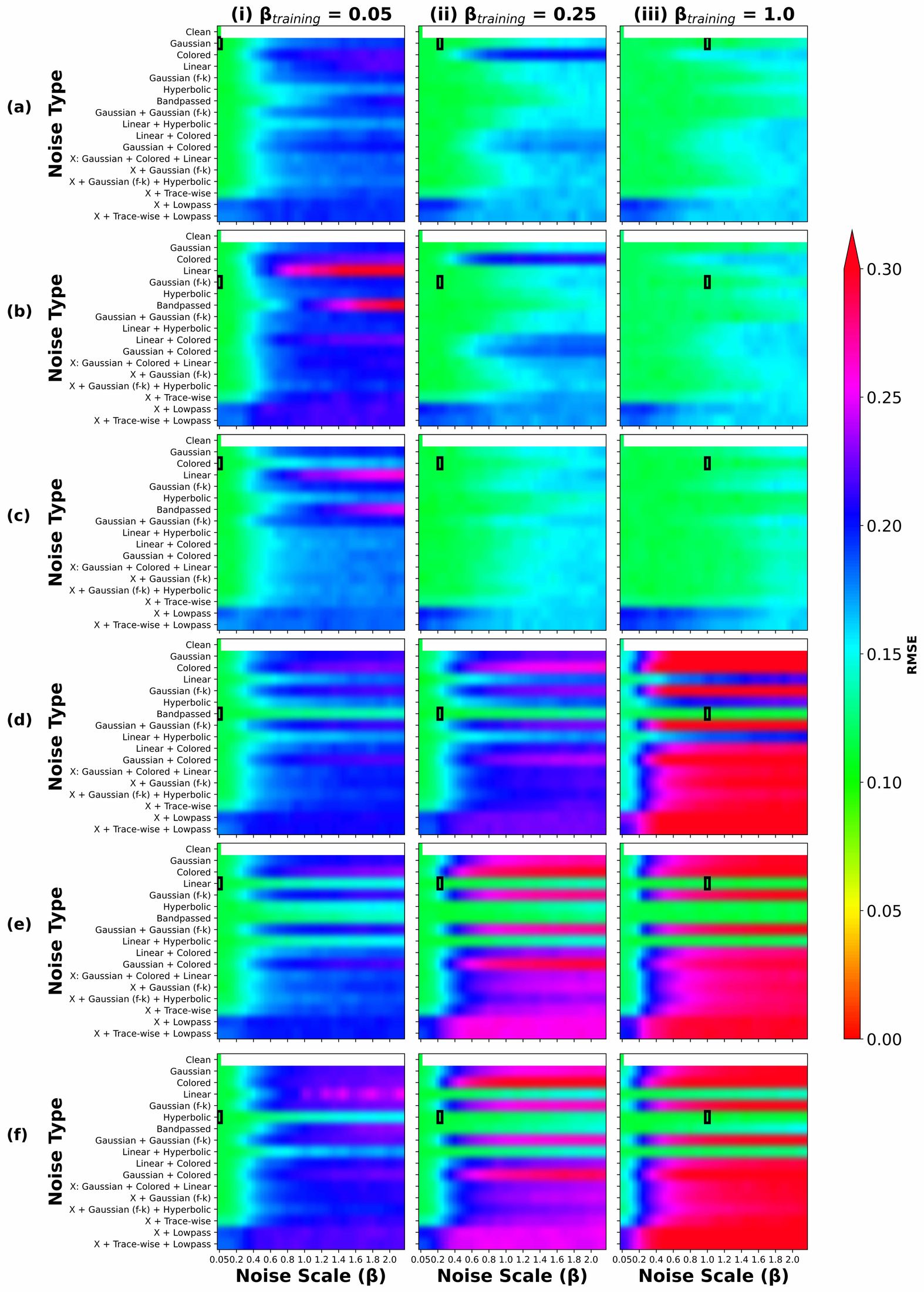} 
    \caption{Robustness matrices for the U-Net architecture trained on the denoising task with different noise scales, $\beta_{\text{training}} \in \{0.05, 0.25, 1.0\}$, and single noise types: (a) Gaussian, (b) Gaussian (f-k), (c) Colored, (d) Bandpassed, (e) Linear, and (f) Hyperbolic. Results are computed using model weights after 100 epochs of training.}
    \label{f19}
\end{figure}

\begin{figure}[h]
    \centering\includegraphics[width=0.8\textwidth,height=0.9\textheight, keepaspectratio]{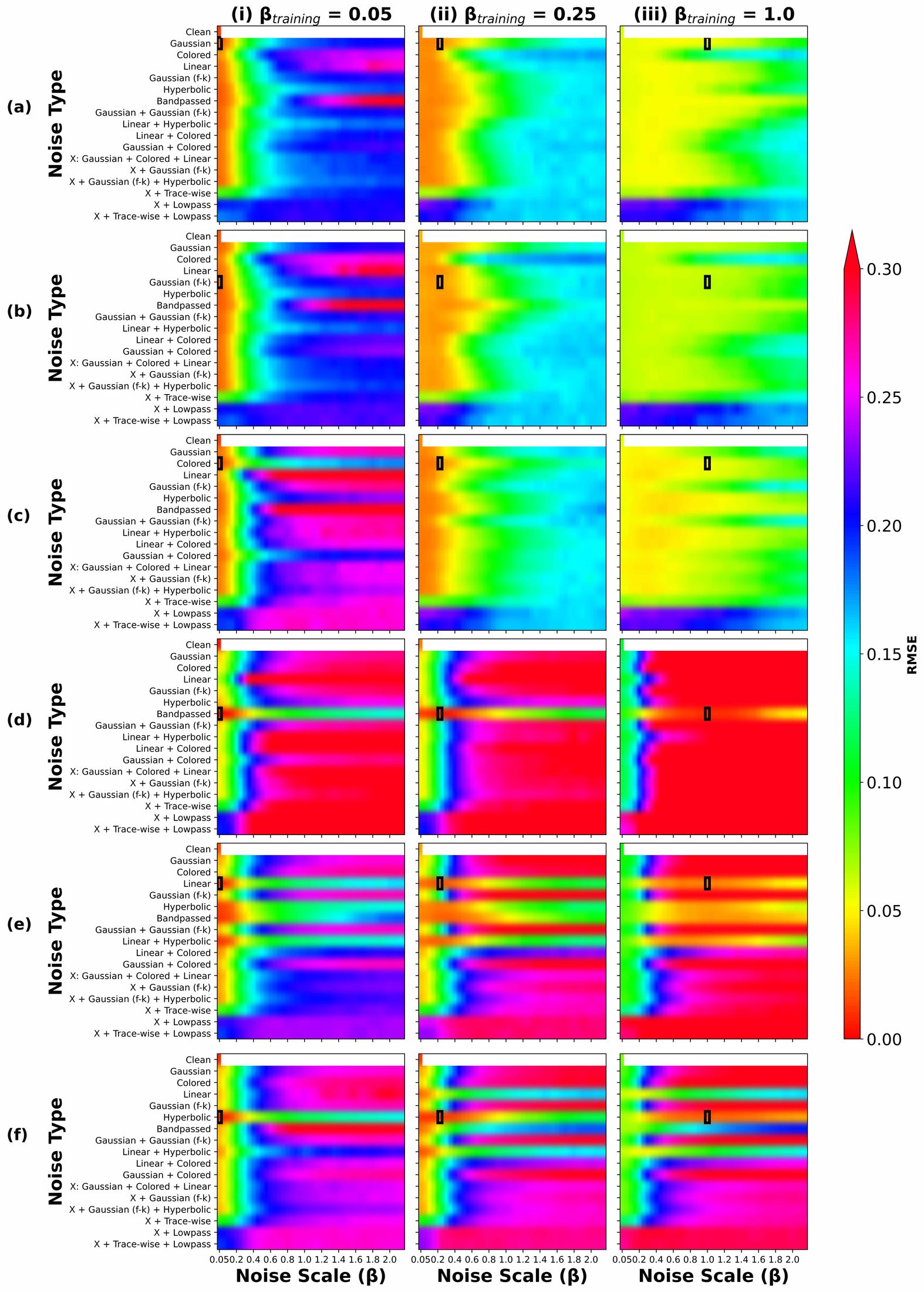} 
    \caption{Robustness matrices for the Swin-U architecture trained on the denoising task with different noise scales, $\beta_{\text{training}} \in \{0.05, 0.25, 1.0\}$, and single noise types: (a) Gaussian, (b) Gaussian (f-k), (c) Colored, (d) Bandpassed, (e) Linear, and (f) Hyperbolic. Results are computed using model weights after 100 epochs of training.}
    \label{f20}
\end{figure}

\begin{figure}[h]
    \centering\includegraphics[width=\textwidth,height=0.8\textheight, keepaspectratio]{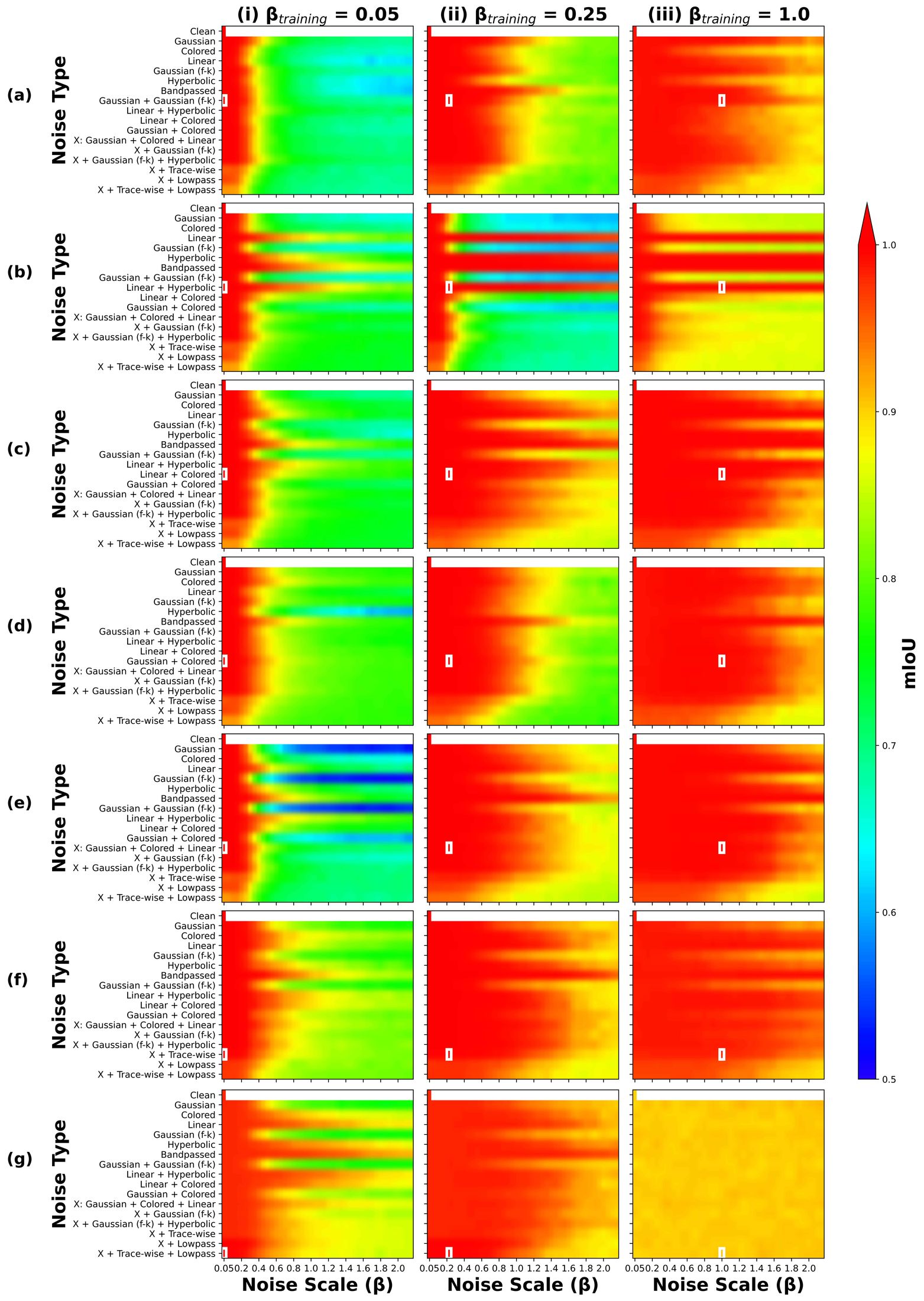} 
    \caption{Robustness matrices for the Swin-U architecture trained on the first break picking task with different noise scales, $\beta_{\text{training}} \in \{0.05, 0.25, 1.0\}$, and compound noises: (a) Random, (b) Structured, (c) Linear + Colored, (d) Gaussian + Colored, (e) Gaussian + Colored + Linear, (f) e + Trace-wise (g) f + Low-pass (fixed frequency cutoff). Results are computed using model weights after 50 epochs of training.}
    \label{f33}
\end{figure}

\begin{figure}[h]
    \centering\includegraphics[width=0.8\textwidth,height=0.9\textheight, keepaspectratio]{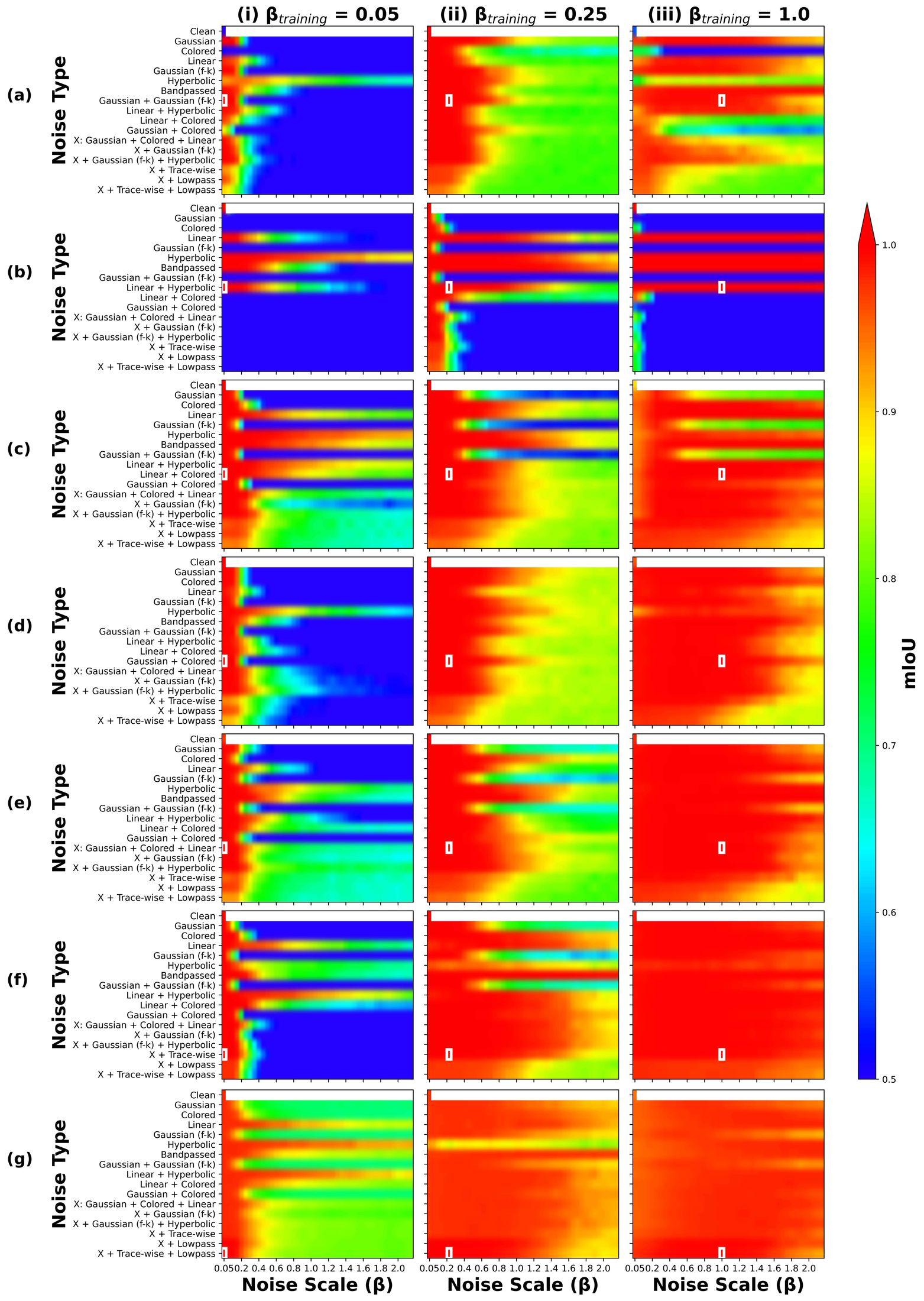} 
    \caption{Robustness matrices for the Restormer architecture trained on the first break picking task with different noise scales, $\beta_{\text{training}} \in \{0.05, 0.25, 1.0\}$, and compound noises: (a) Random, (b) Structured, (c) Linear + Colored, (d) Gaussian + Colored, (e) Gaussian + Colored + Linear, (f) e + Trace-wise (g) f + Low-pass (fixed frequency cutoff). Results are computed using model weights after 50 epochs of training.}
    \label{f34}
\end{figure}

\begin{figure}[h]
    \centering\includegraphics[width=0.8\textwidth,height=0.9\textheight, keepaspectratio]{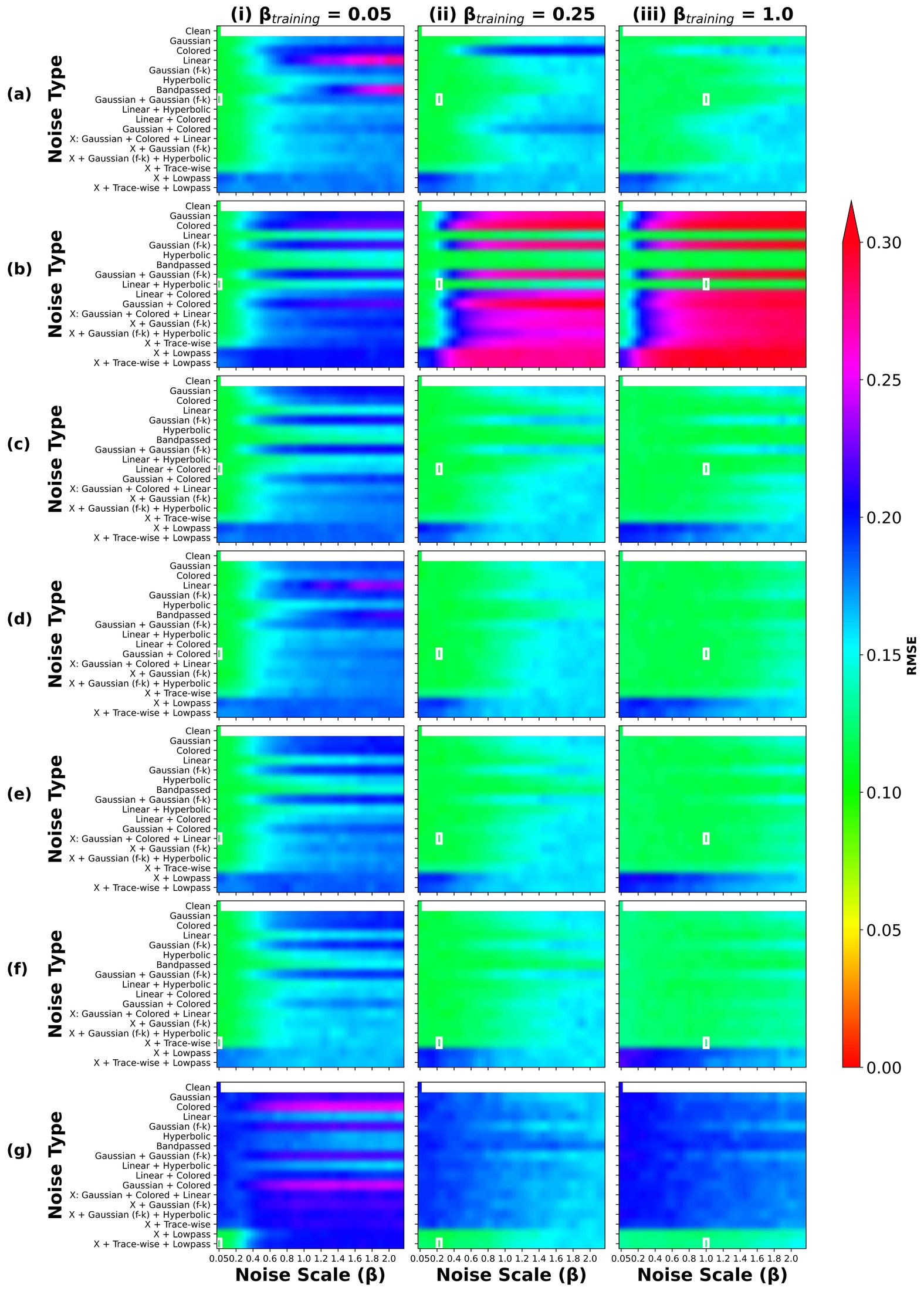} 
    \caption{Robustness matrices for the U-Net architecture trained on the denoising task with different noise scales, $\beta_{\text{training}} \in \{0.05, 0.25, 1.0\}$, and compound noises: (a) Random, (b) Structured, (c) Linear + Colored, (d) Gaussian + Colored, (e) Gaussian + Colored + Linear, (f) e + Trace-wise (g) f + Low-pass (fixed frequency cutoff). Results are computed using model weights after 100 epochs of training.}
    \label{f35}
\end{figure}

\begin{figure}[h]
    \centering\includegraphics[width=0.8\textwidth,height=0.9\textheight, keepaspectratio]{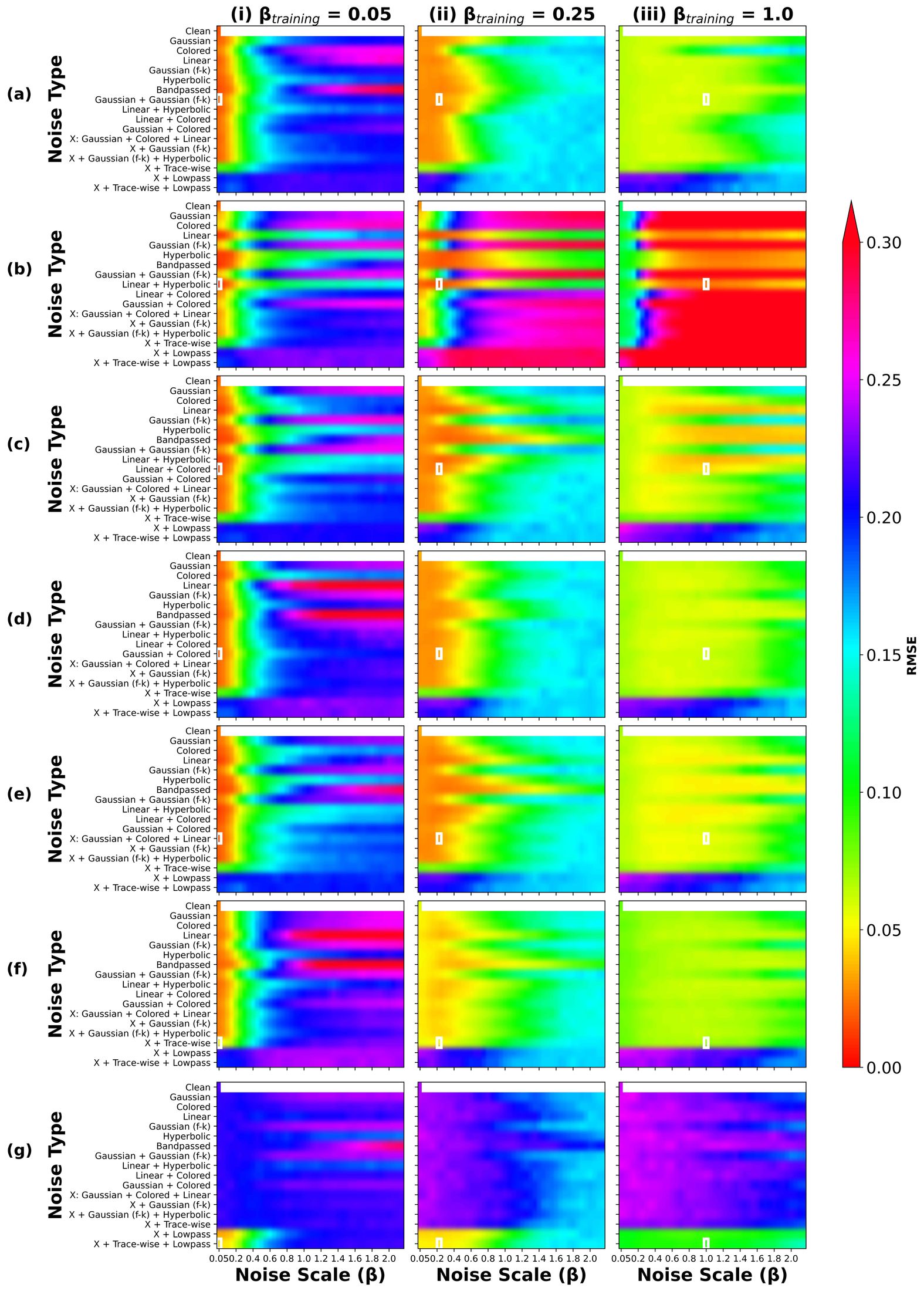} 
    \caption{Robustness matrices for the Swin-U architecture trained on the denoising task with different noise scales, $\beta_{\text{training}} \in \{0.05, 0.25, 1.0\}$, and compound noises: (a) Random, (b) Structured, (c) Linear + Colored, (d) Gaussian + Colored, (e) Gaussian + Colored + Linear, (f) e + Trace-wise (g) f + Low-pass (fixed frequency cutoff). Results are computed using model weights after 100 epochs of training.}
    \label{f36}
\end{figure}

\begin{figure}[h] 
\centering\includegraphics[width=0.5\textwidth]{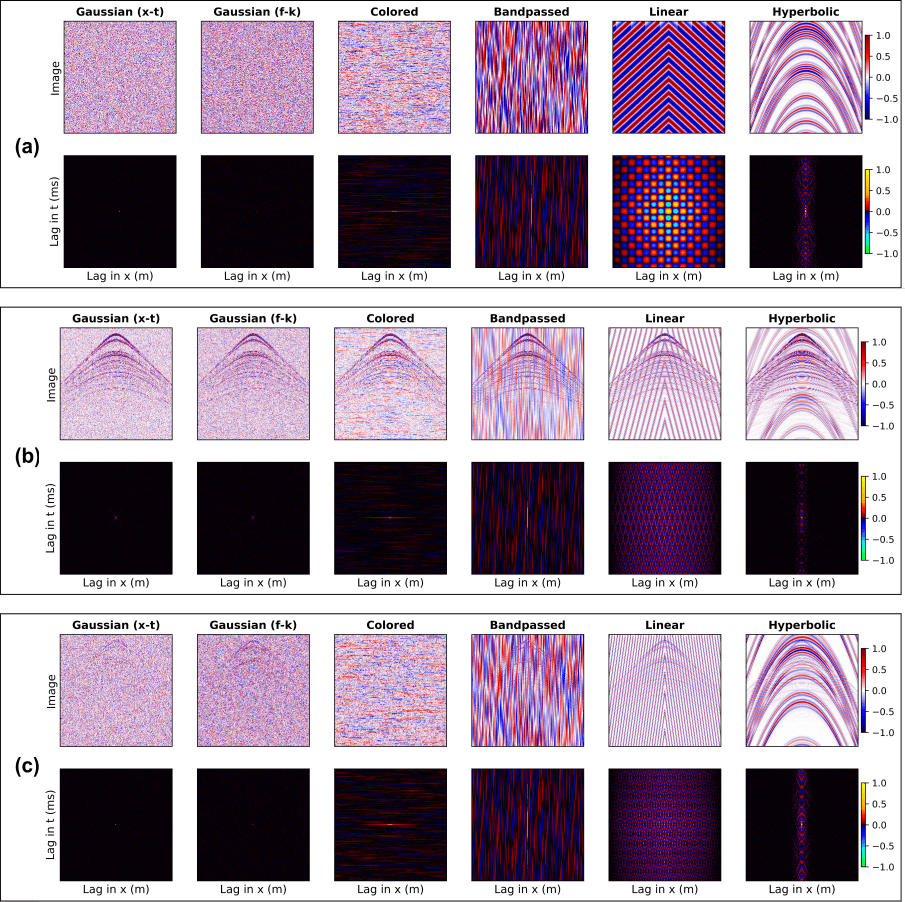}
\caption{2-D autocorrelation function of single noise. (a) Single noise; (b) image contaminated with single noise, $\beta_{0.25}$; (c) image contaminated with single noise, $\beta_{1.0}$.}
\label{sf0}
\end{figure}

\begin{figure}[h] 
\centering\includegraphics[width=0.5\textwidth]{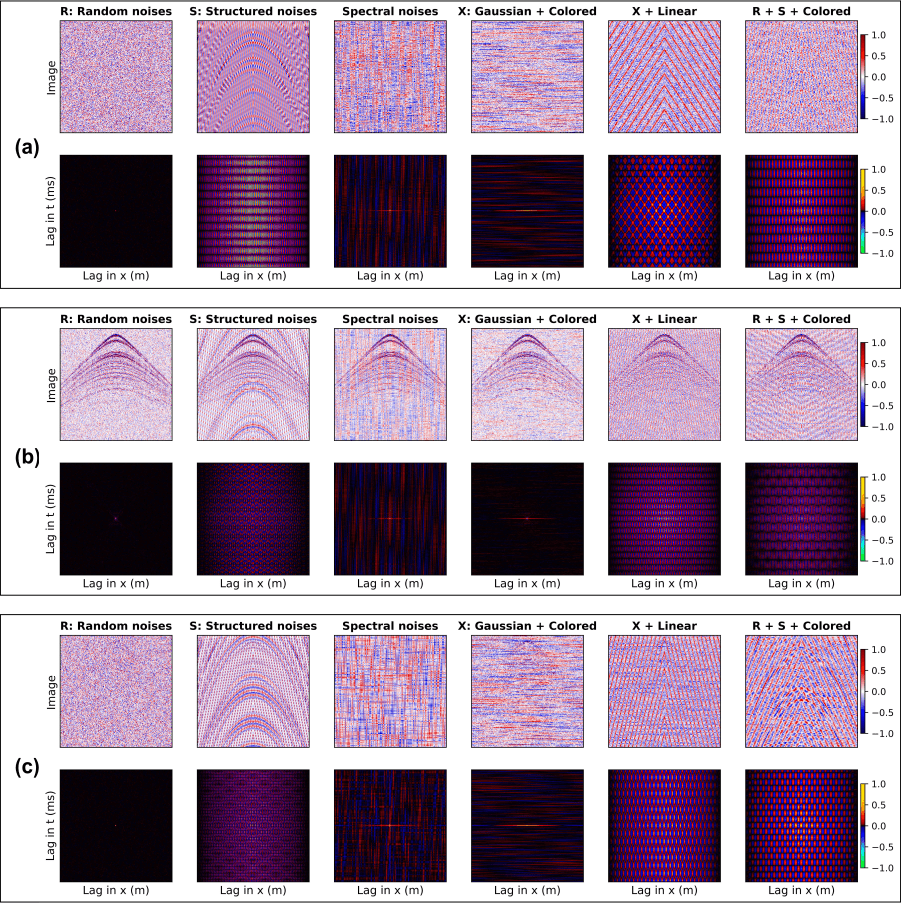}
\caption{2-D autocorrelation function of compound noise. (a) Compound noise; (b) image contaminated with compound noise, $\beta_{0.25}$; (c) image contaminated with compound noise, $\beta_{1.0}$.}
\label{sfx0}
\end{figure}

\begin{figure}[h] 
\centering\includegraphics[width=0.8\textwidth]{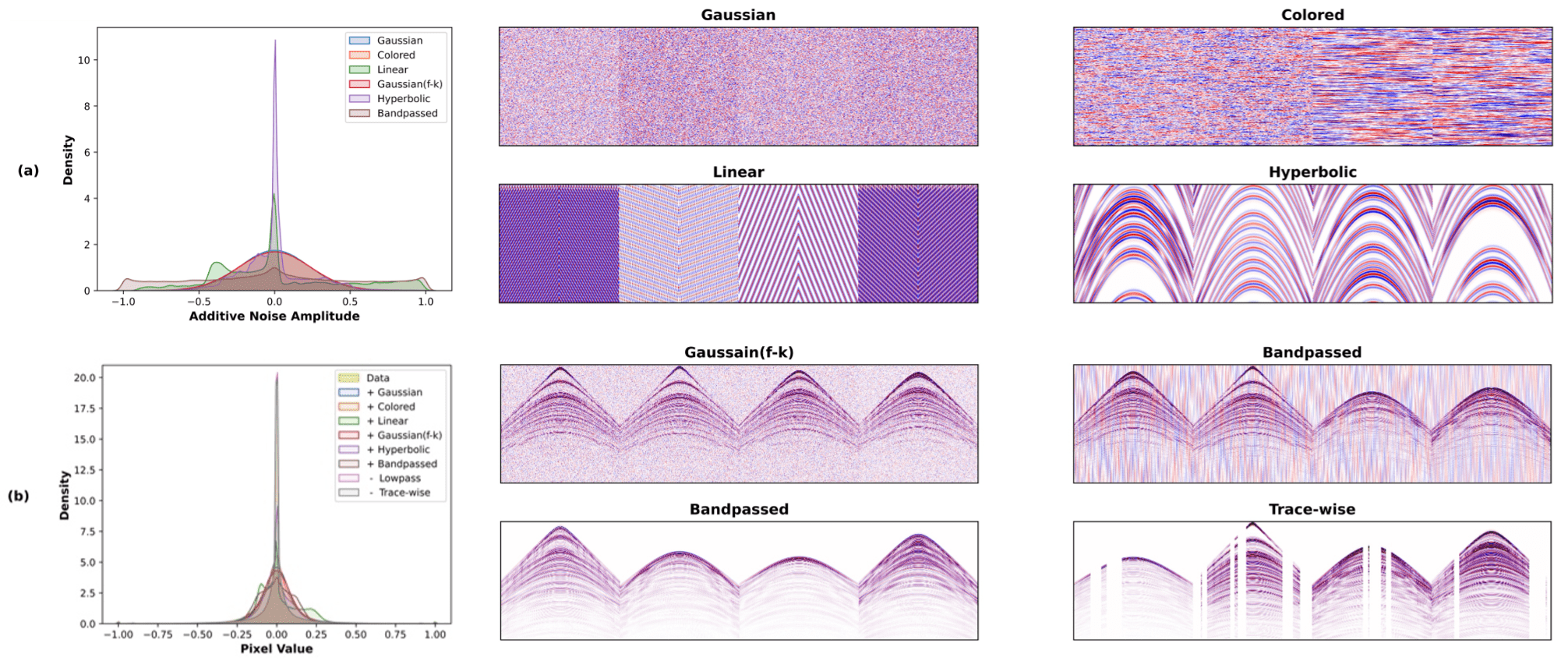}
\caption{Examples of random realizations generated by sampling the distributions of the single noise function n(t) parameters (a) Noise distribution prior to scaling (b) Data combined with noise ($\beta$ =0.1).}
\label{f2}
\end{figure}

\begin{figure}[h] 
\centering\includegraphics[width=0.6\textwidth]{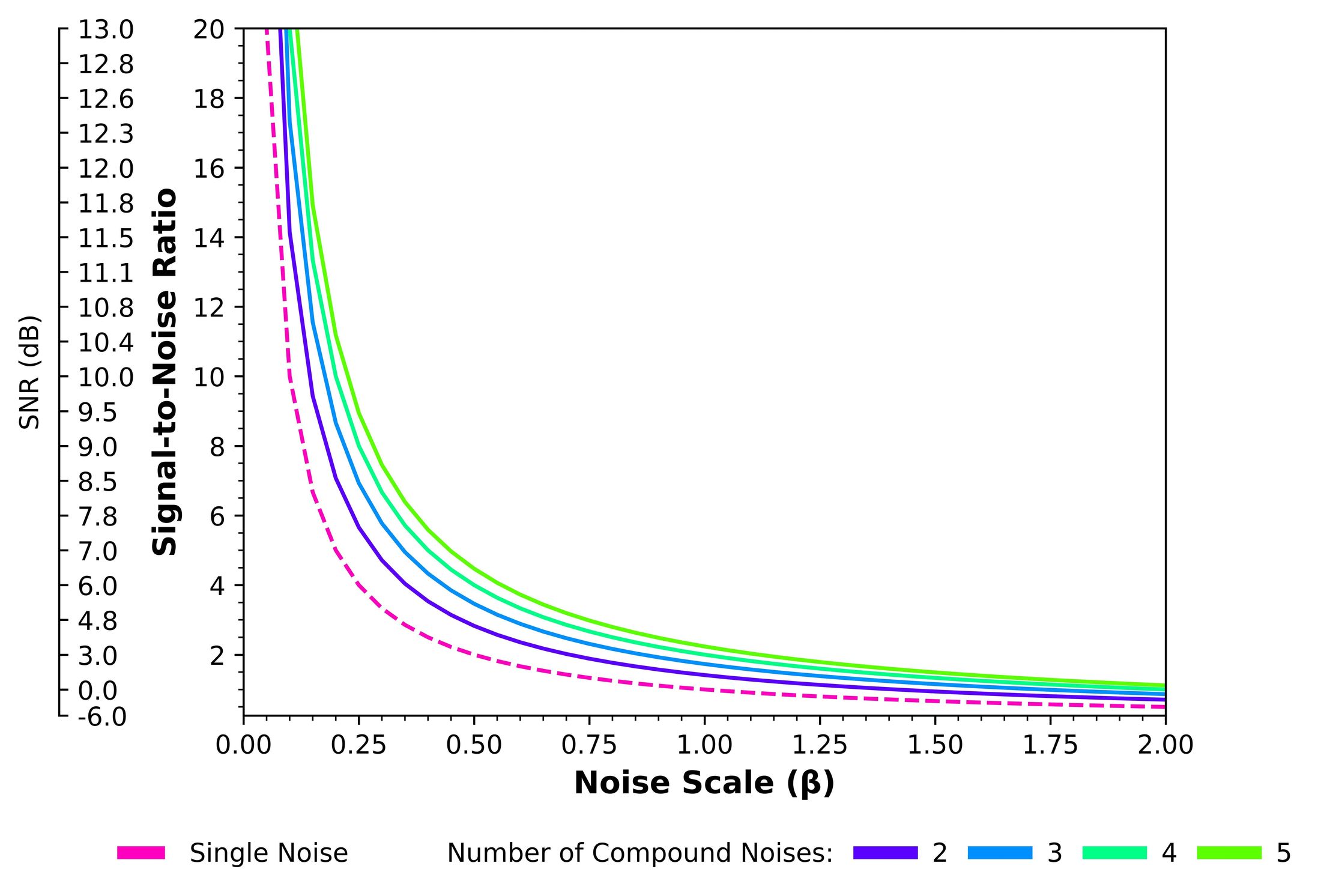}
\caption{Noise scale $\beta$  vs. SNR.}
\label{f3}
\end{figure}

\begin{figure}[h] 
\centering\includegraphics[width=0.75\textwidth]{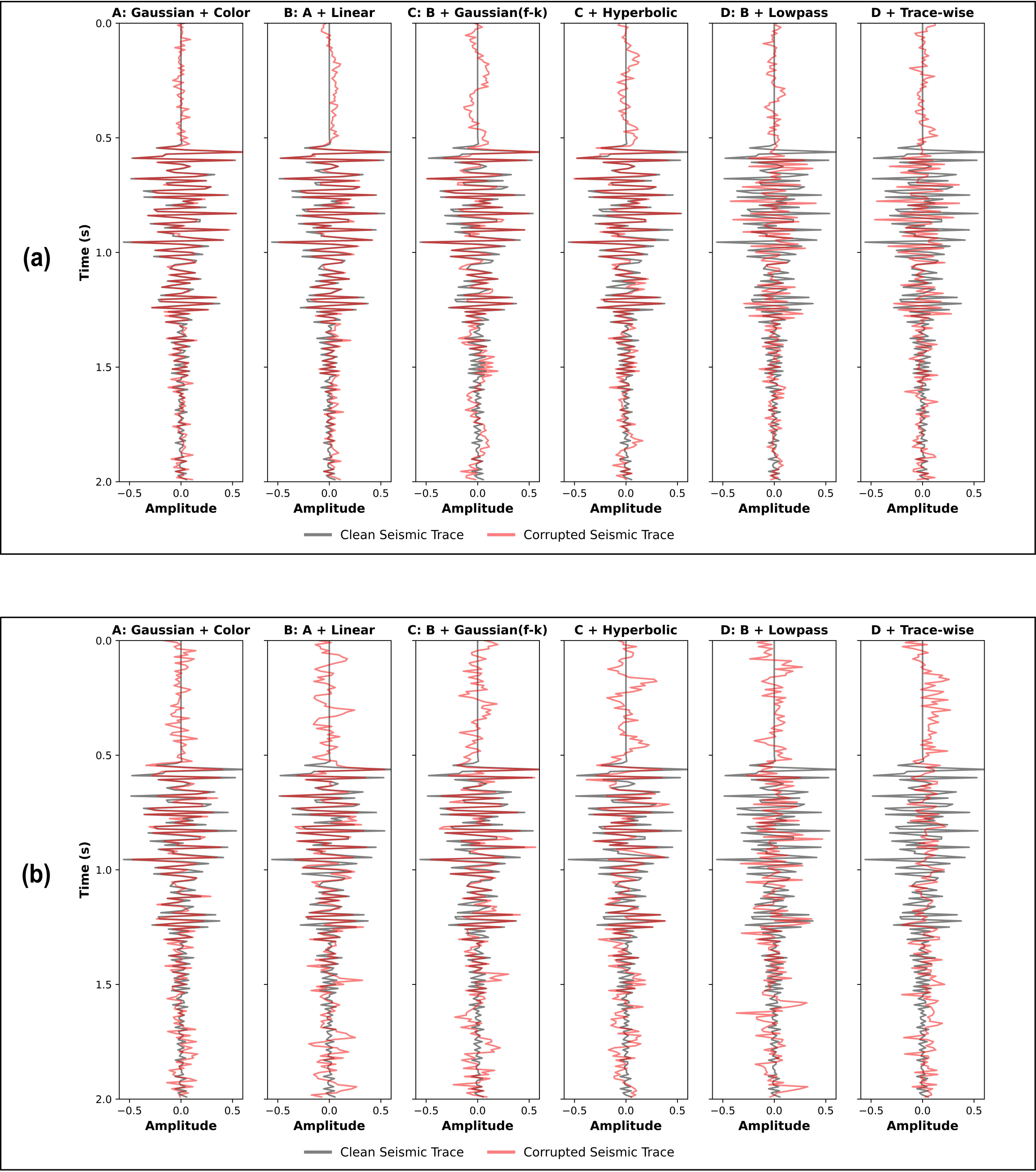}
\caption{1-D view of a random trace after augmenting with compound noises. (a) $\beta =$ 0.05; (b) $\beta=$ 0.1.}
\label{f6}
\end{figure}

\begin{figure}[h]
\centering
    \includegraphics[scale=0.45]{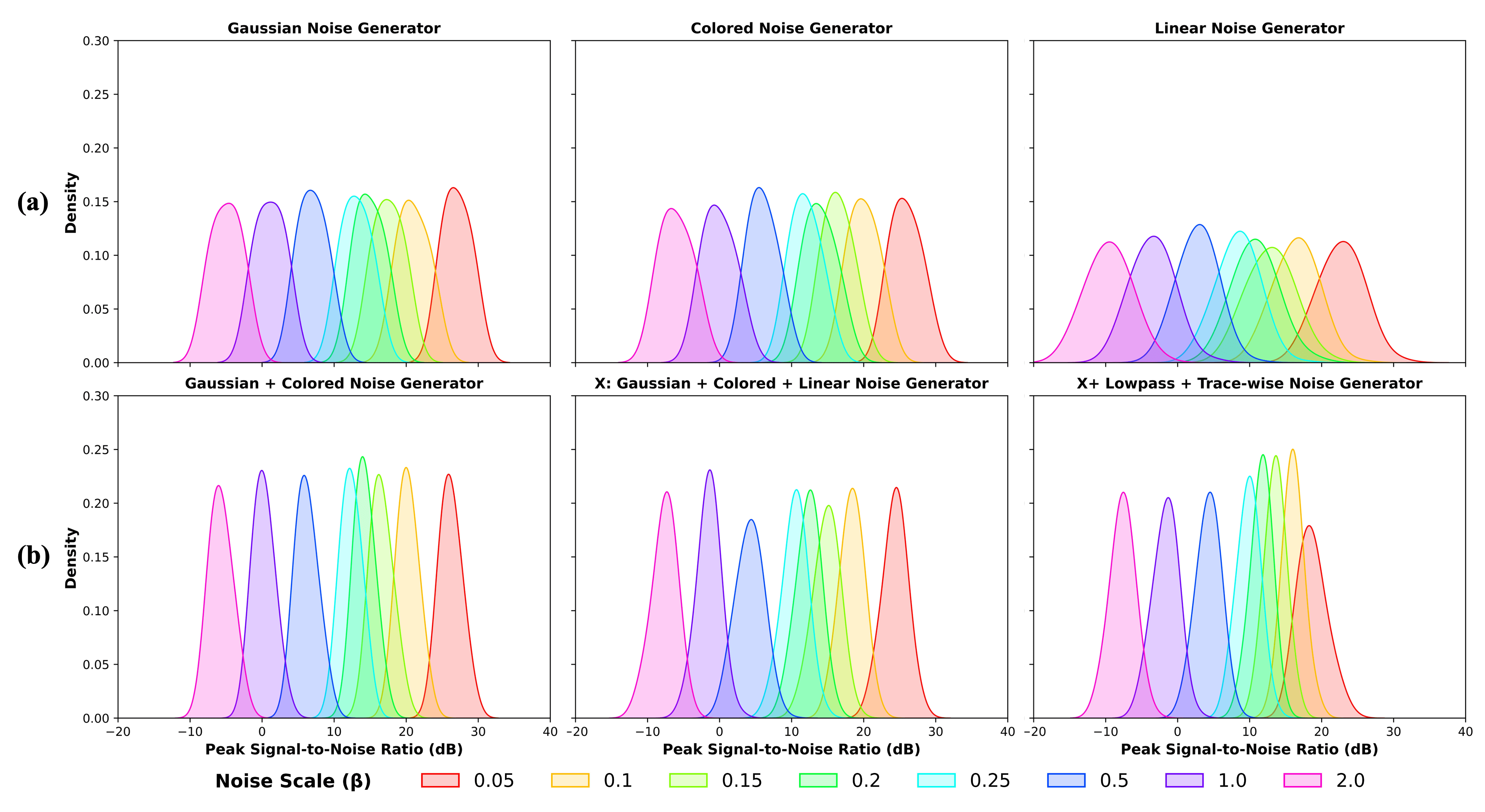} 
    \caption{Image quality (PSNR) with different noise type and scale (500 samples per $\beta$): (a) Single noises (b) Compound noises.}
    \label{f5}
\end{figure}

\begin{figure}[h]
    \centering
    \includegraphics[width=0.75\textwidth]{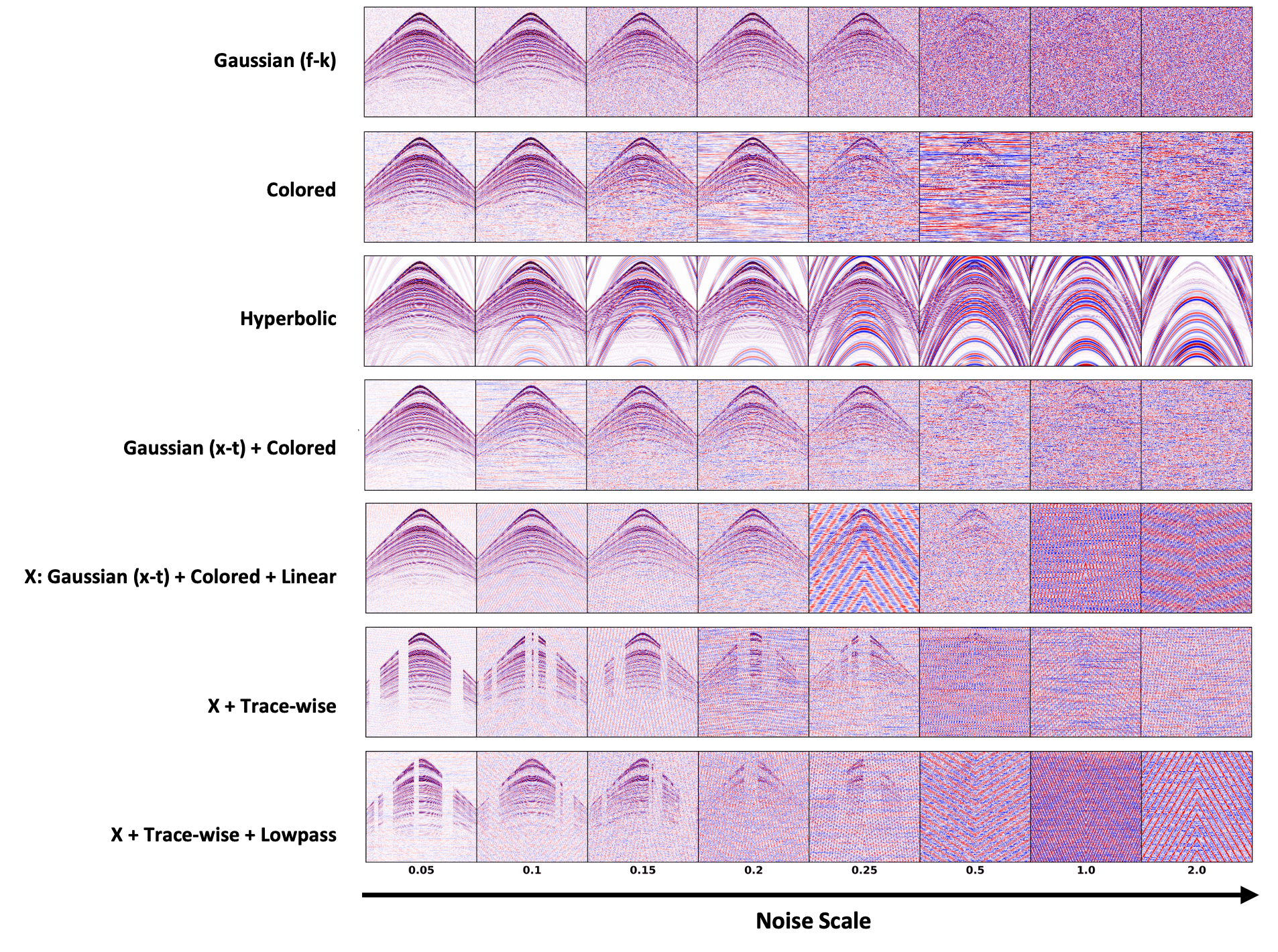} 
    \caption{Visual effect of increasing noise scale ($\beta$) on the same image using random noise profiles (single and compound).}
    \label{f7x}
\end{figure}

\begin{figure}[h]
    \centering
    \includegraphics[width=0.9\textwidth]{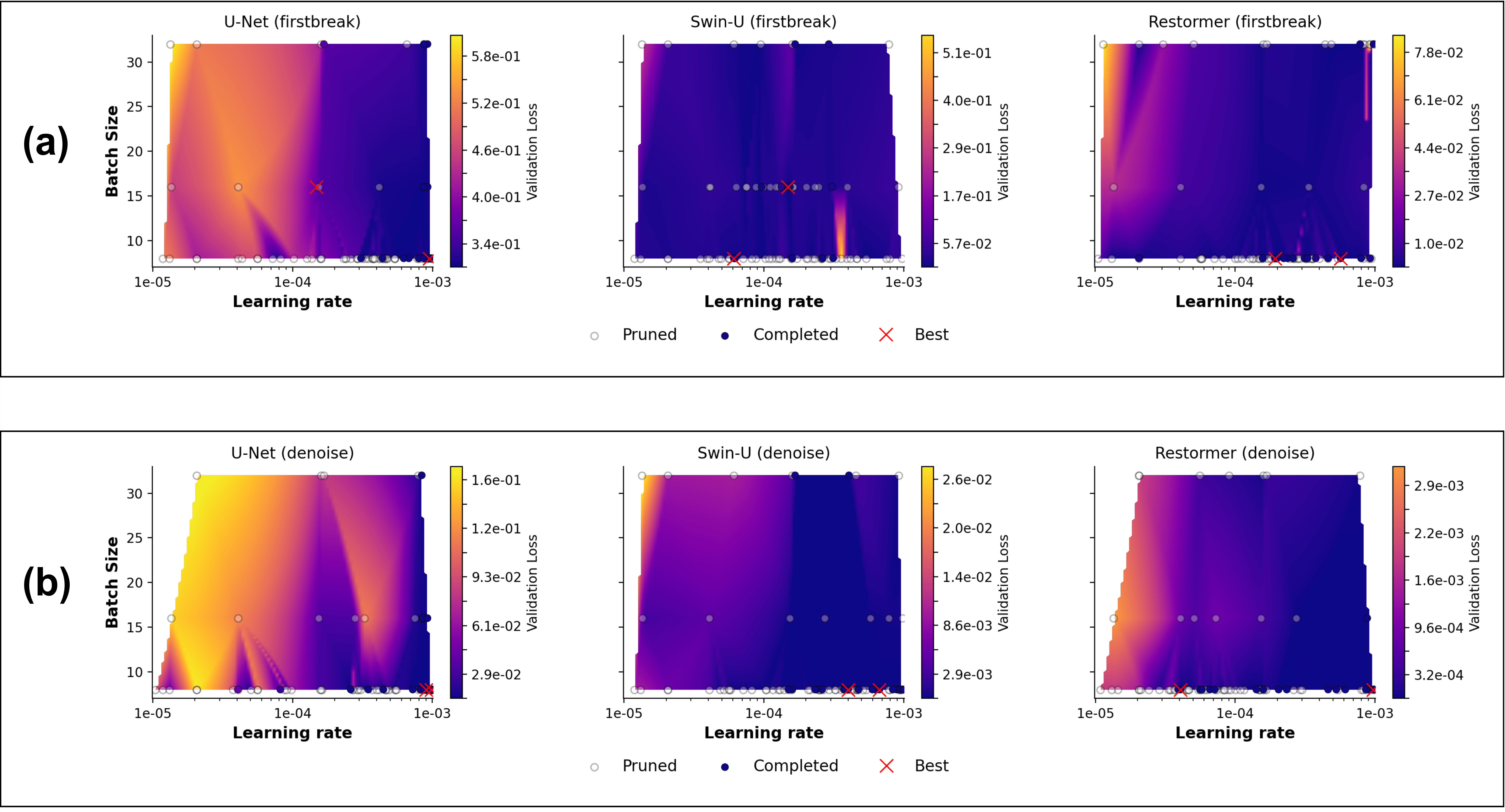} 
    \caption{Contour maps of validation loss as a function of learning rate and batch size for the three network architectures on (a) the first-break picking task and (b) the denoising task. For each task, the reported optimum corresponds to the lowest validation loss achieved among the three models.}
    \label{f41}
\end{figure}

\begin{figure}[h]
    \centering
    \includegraphics[width=0.8\textwidth]{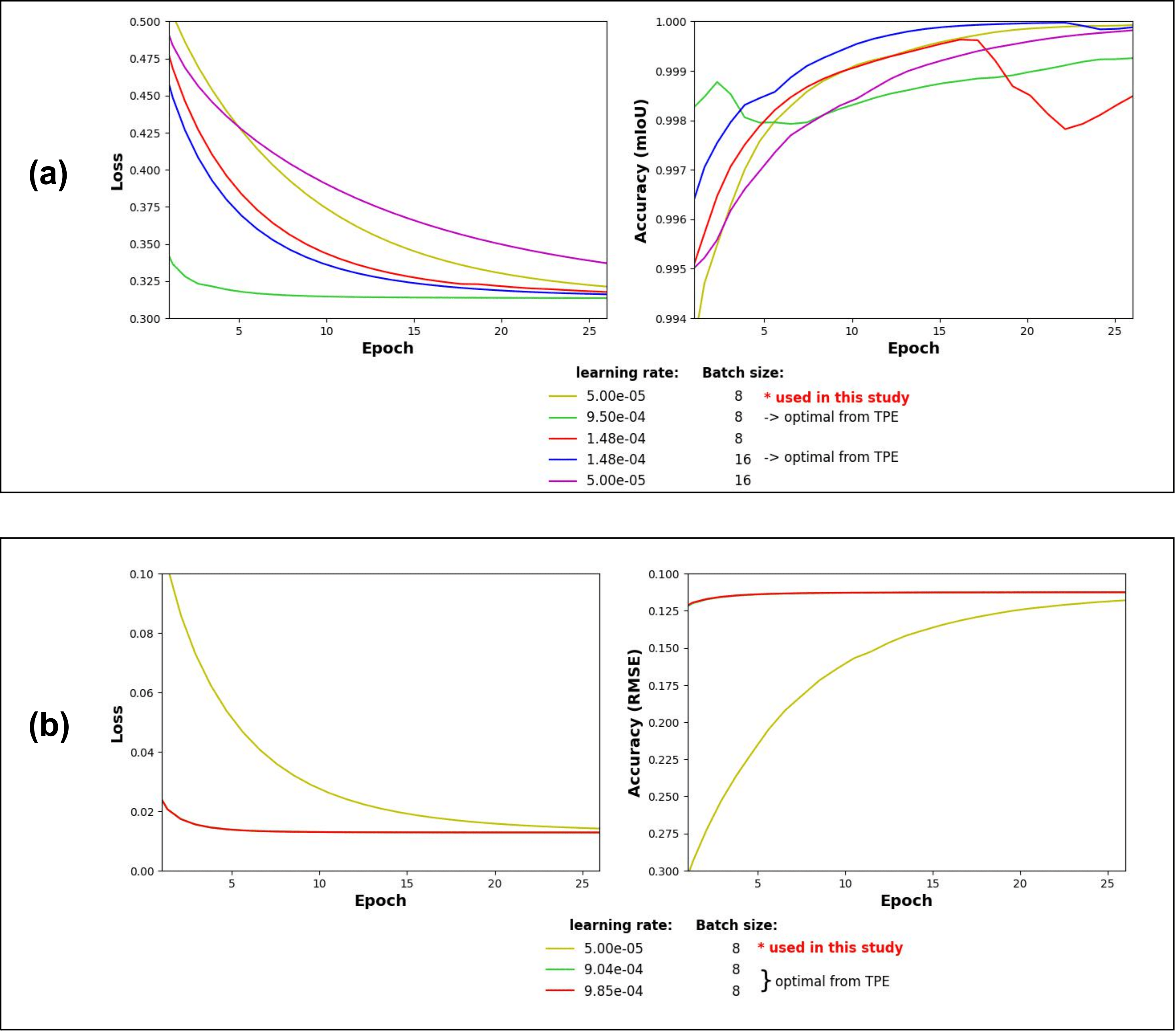} 
    \caption{Validation loss and accuracy of U-Net as a function of learning rate and batch size for (a) first-break picking and (b) denoising tasks.}
    \label{f41x}
\end{figure}

\begin{figure}[h]
\centering\includegraphics[width=0.9\textwidth]{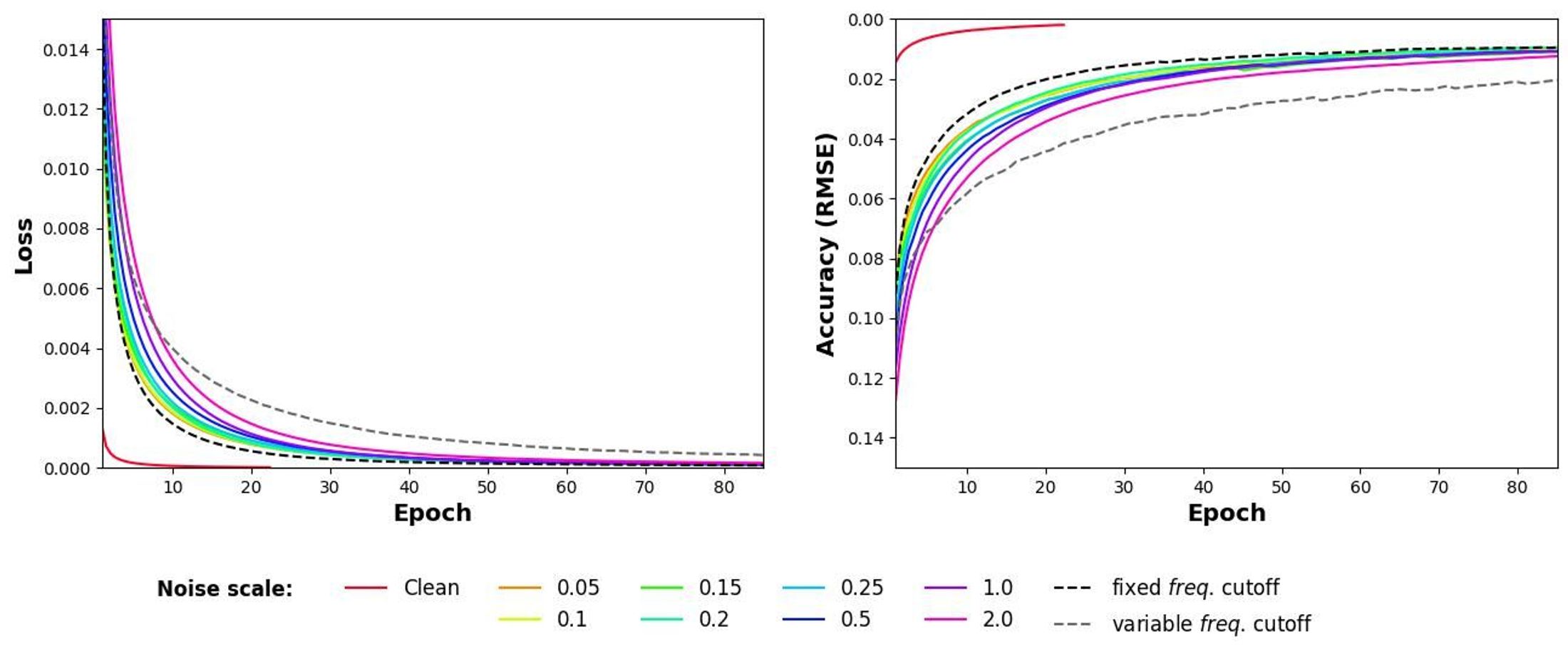} 
    \caption{Validation curves for Restormer models trained with a low pass filter on the denoising task, color coded by noise scale.  Dashed line training using a (black) frequency cutoff centered around $f_{dominant}$, (grey) variable frequency cutoff  across images in the batch: (left) loss, (right) accuracy.}
    \label{f11}
\end{figure}

\begin{figure}[h]
\centering\includegraphics[width=0.9\textwidth]{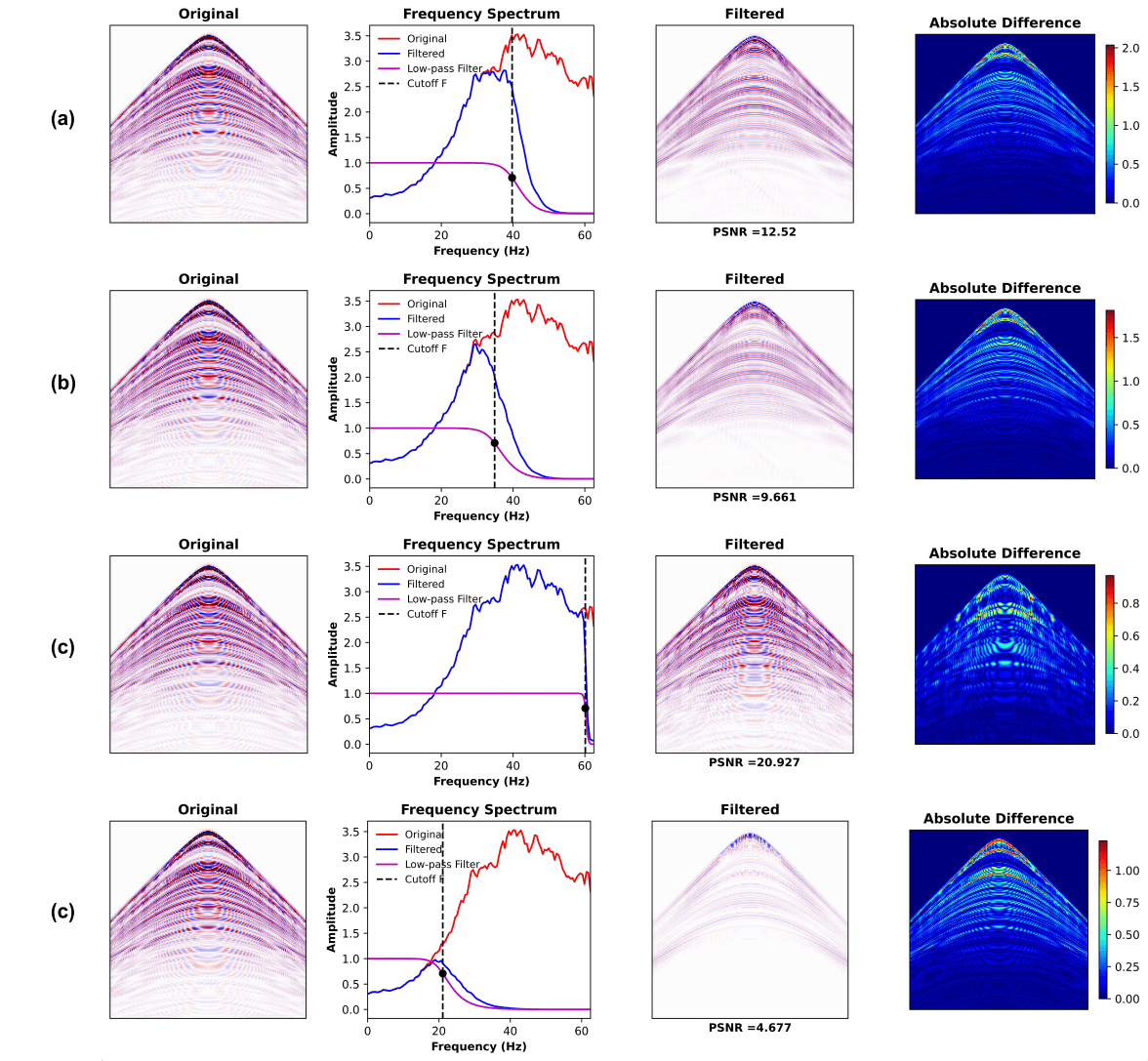} 
    \caption{Visual examples illustrating how different low-pass cutoff choices influence the appearance and smoothness of the seismic gather.(a–b) Cutoffs centered around the dominant frequency; dashed black line in Figure~\ref{f11} (c–d) Cutoffs determined by the noise-scale parameter $\beta$, where larger $\beta$ produces stronger attenuation: (c) $\beta=0.1$, and (d) $\beta=2.0$. }
    \label{sf50}
\end{figure}

\begin{figure}[h]
    \centering
    \includegraphics[width=\textwidth,height=0.9\textheight, keepaspectratio]{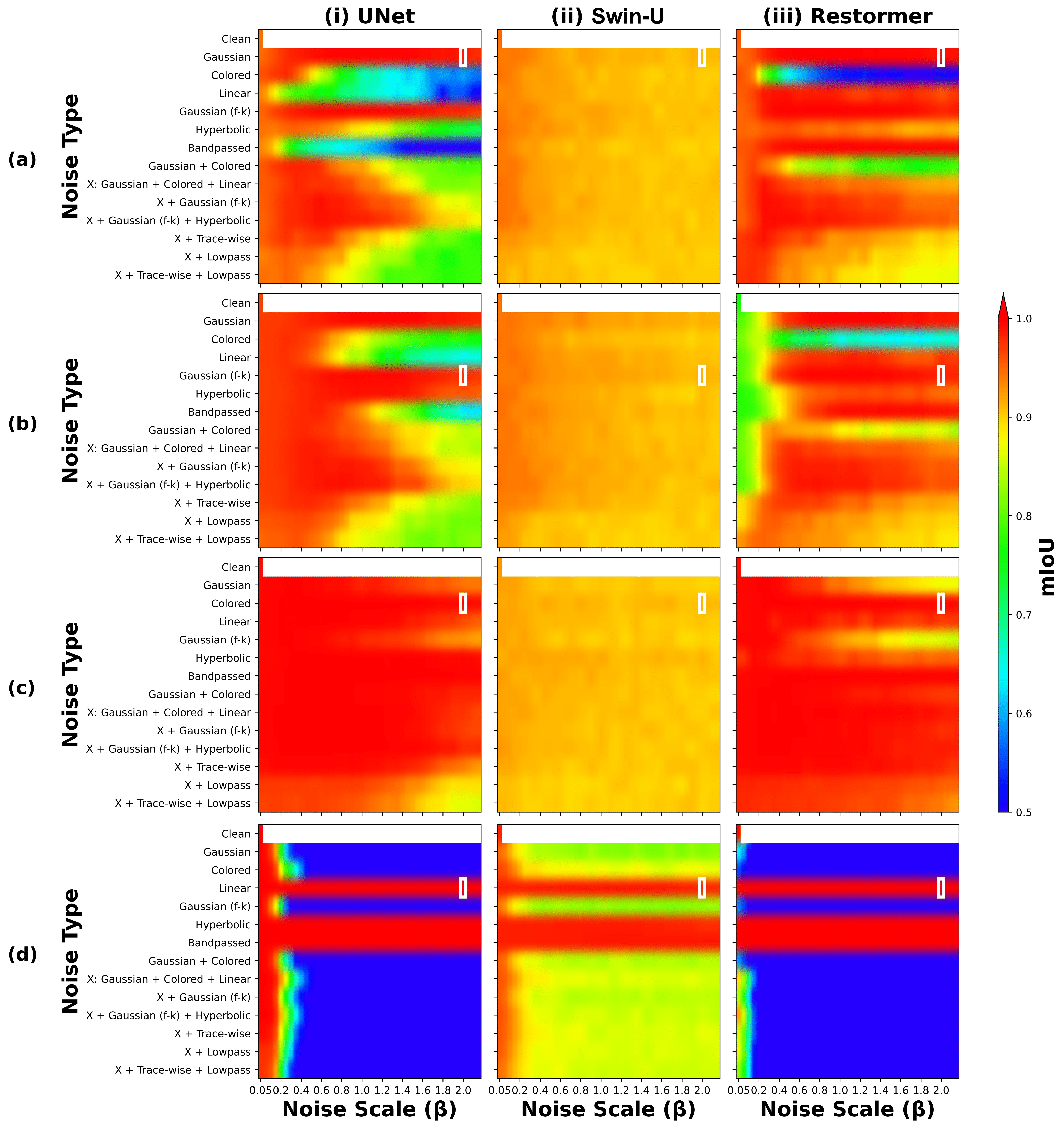} 
    \caption{Robustness matrices $\mathbf{M}_{2.0}$ of (i)U-Net (ii) Swin-U (iii) Restormer trained with random initialization on the first break picking task with various single noise types: (a) Gaussian, (b) Gaussian (f-k), (c) Colored, (d) Linear. Models were trained for 50 epochs.}
    \label{sf2}
\end{figure}

\begin{figure}[h]
    \centering
    \includegraphics[width=\textwidth,height=0.9\textheight, keepaspectratio]{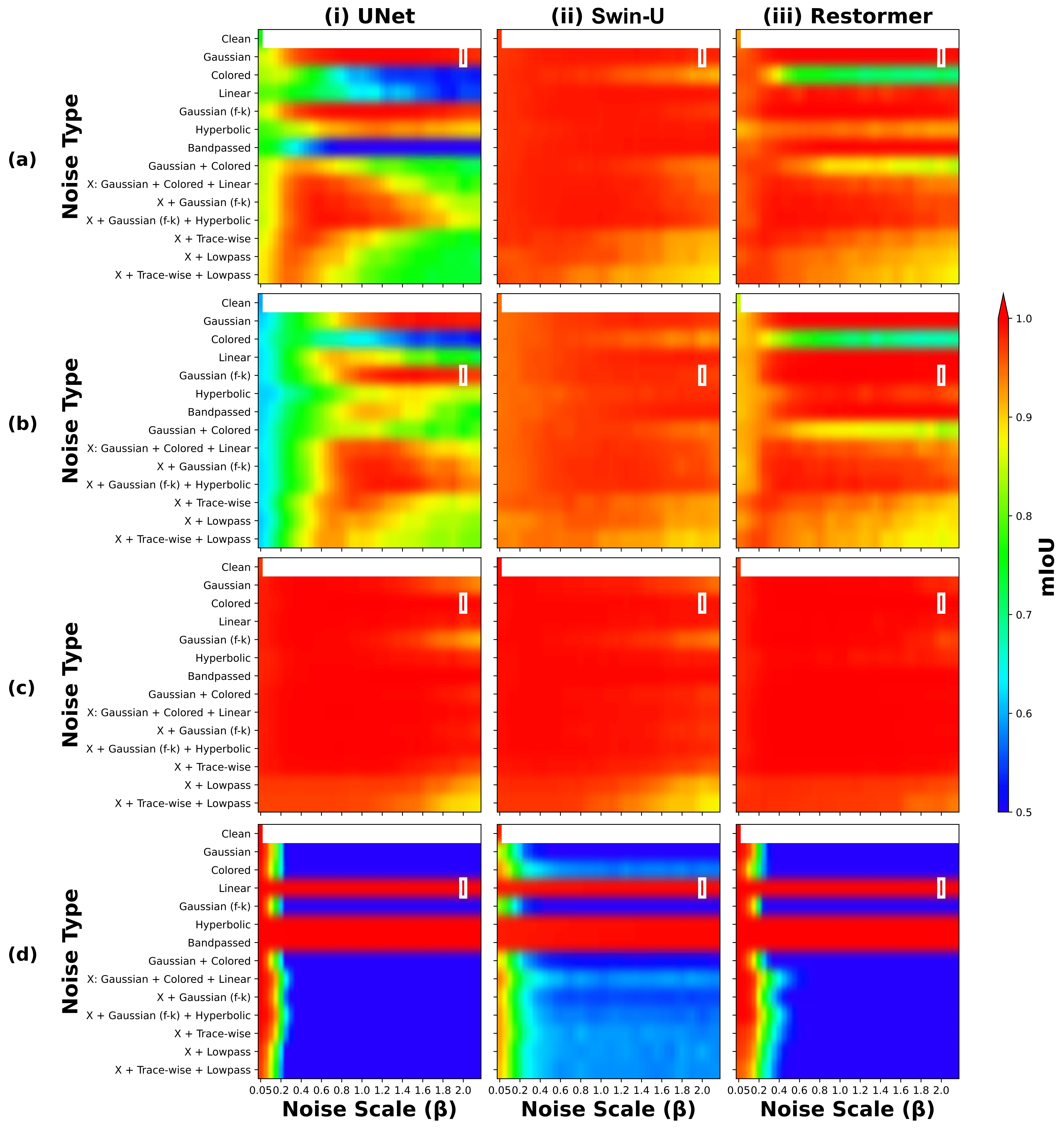} 
    \caption{Robustness matrices $\mathbf{M}_{2.0}$ of (i)U-Net (ii) Swin-U (iii) Restormer initialized using pre-trained weights form the denoising task to train on the first break picking task with various single noise types: (a) Gaussian, (b) Gaussian (f-k), (c) Colored, (d) Linear. Models were trained for 50 epochs.}
    \label{sf2b}
\end{figure}

\begin{figure}[h]
    \centering
    \includegraphics[width=\textwidth,height=0.9\textheight, keepaspectratio]{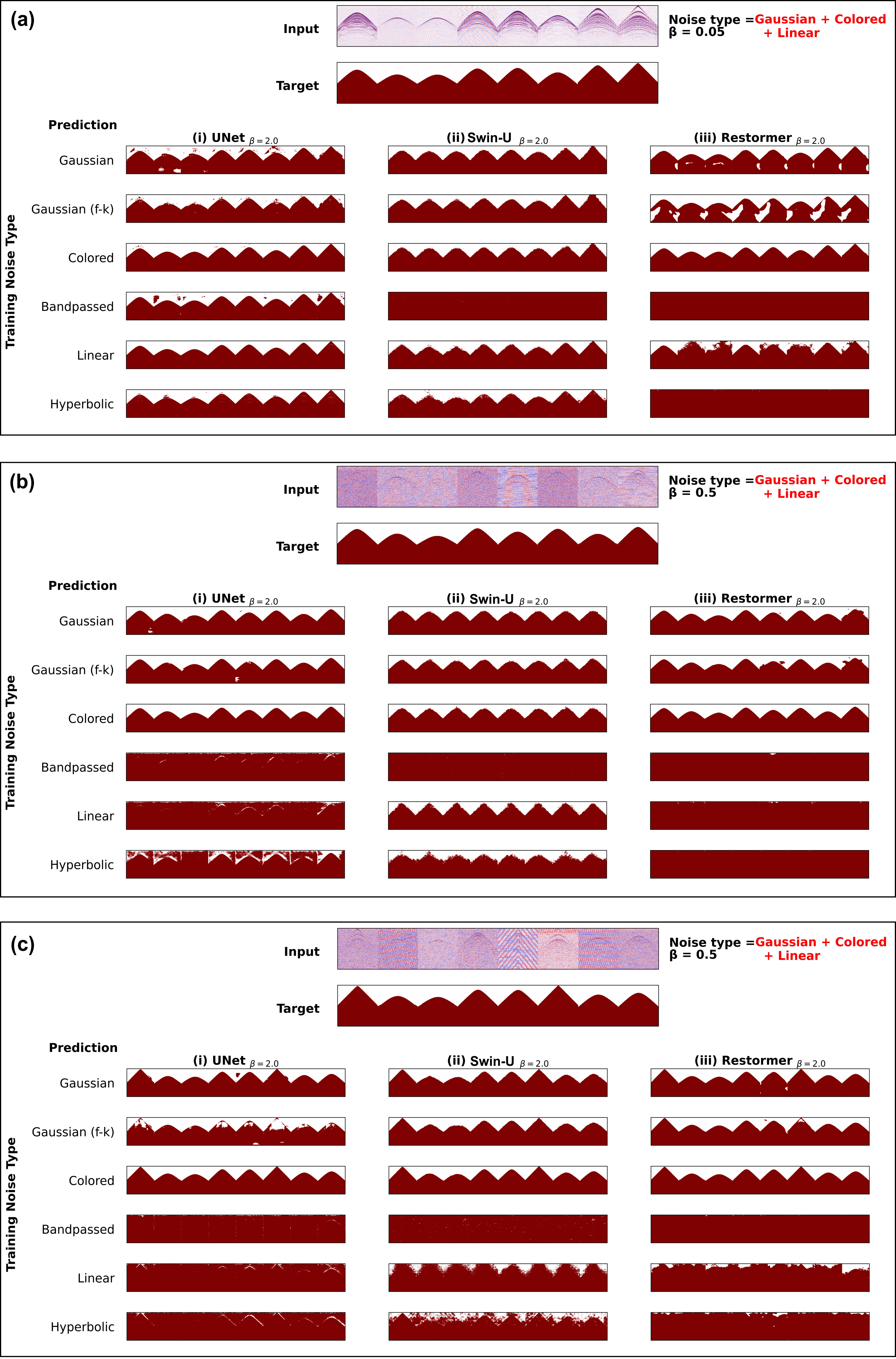} 
    \caption{Effect of high scale ($\beta=2.0$) single noise training on network predictions for the first break picking task. Compound noise (a) at low scale $\beta_{0.05}$, (b) at medium scale $\beta_{0.5}$, (c) at medium scale $\beta_{0.5}$ after pretraining.}
    \label{sf3}
\end{figure}

\begin{figure}[h]
    \centering
    \includegraphics[width=\textwidth]{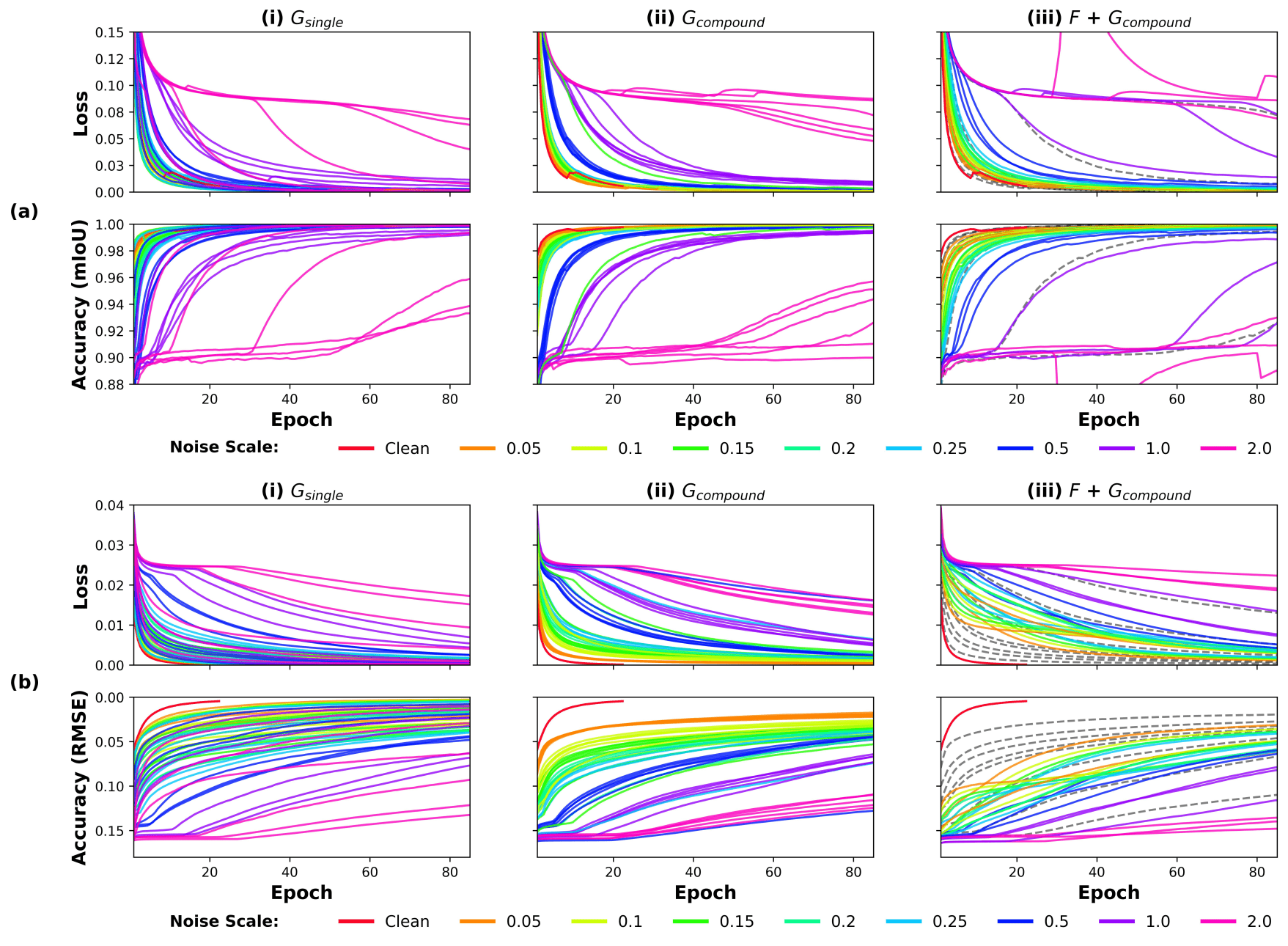} 
    \caption{Validation curves of Swin-U models under varying noise scales for (a) first break picking, and (b) denoising tasks. Noise was generated by (i) $G_{\text{single}}$, (ii) $G_{\text{compound}}$, and (iii) $G_{\text{compound}}$ with additional data manipulation noise from $F$. Dashed gray line in (iii) indicates performance without $F$.}
    \label{fs12}
\end{figure}

\begin{figure}[th]
    \centering
    \includegraphics[width=\textwidth]{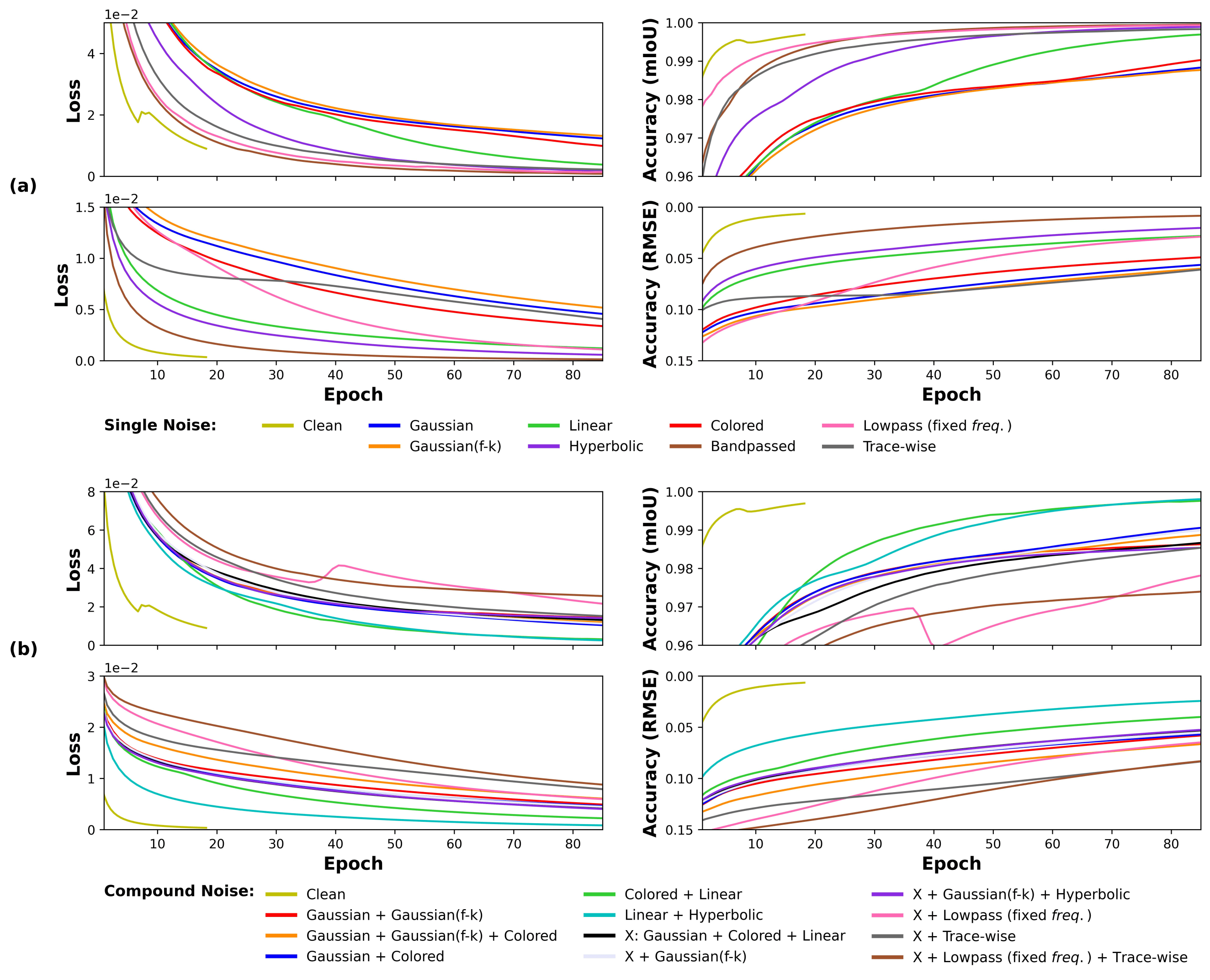} 
    \caption{Mean validation curves of Swin-U models trained on (a) single and (b) compound noise types. Top panels show results for first-break picking task, and bottom panels for denoising tasks. Curves are color-coded by noise type. Left column: loss; right column: accuracy.}
    \label{fs13}
\end{figure}

\begin{figure}[th]
    \centering
    \includegraphics[width=\textwidth]{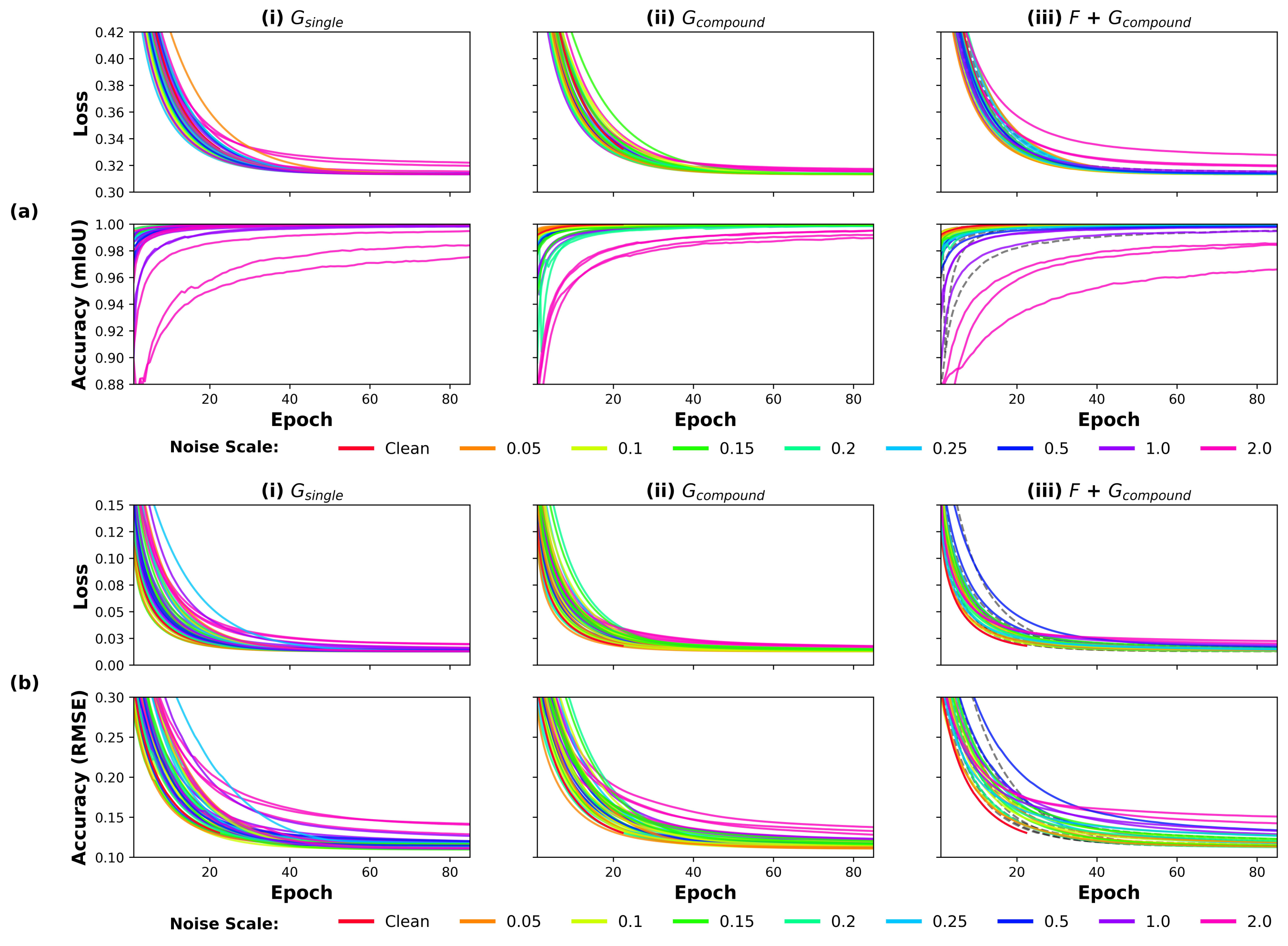} 
    \caption{Validation curves of U-Net models under varying noise scales for (a) first break picking, and (b) denoising tasks. Noise was generated by (i) $G_{\text{single}}$, (ii) $G_{\text{compound}}$, and (iii) $G_{\text{compound}}$ with additional data manipulation noise from $F$. Dashed gray line in (iii) indicates performance without $F$.}
    \label{fs14}
\end{figure}

\begin{figure}[th]
    \centering
    \includegraphics[width=\textwidth]{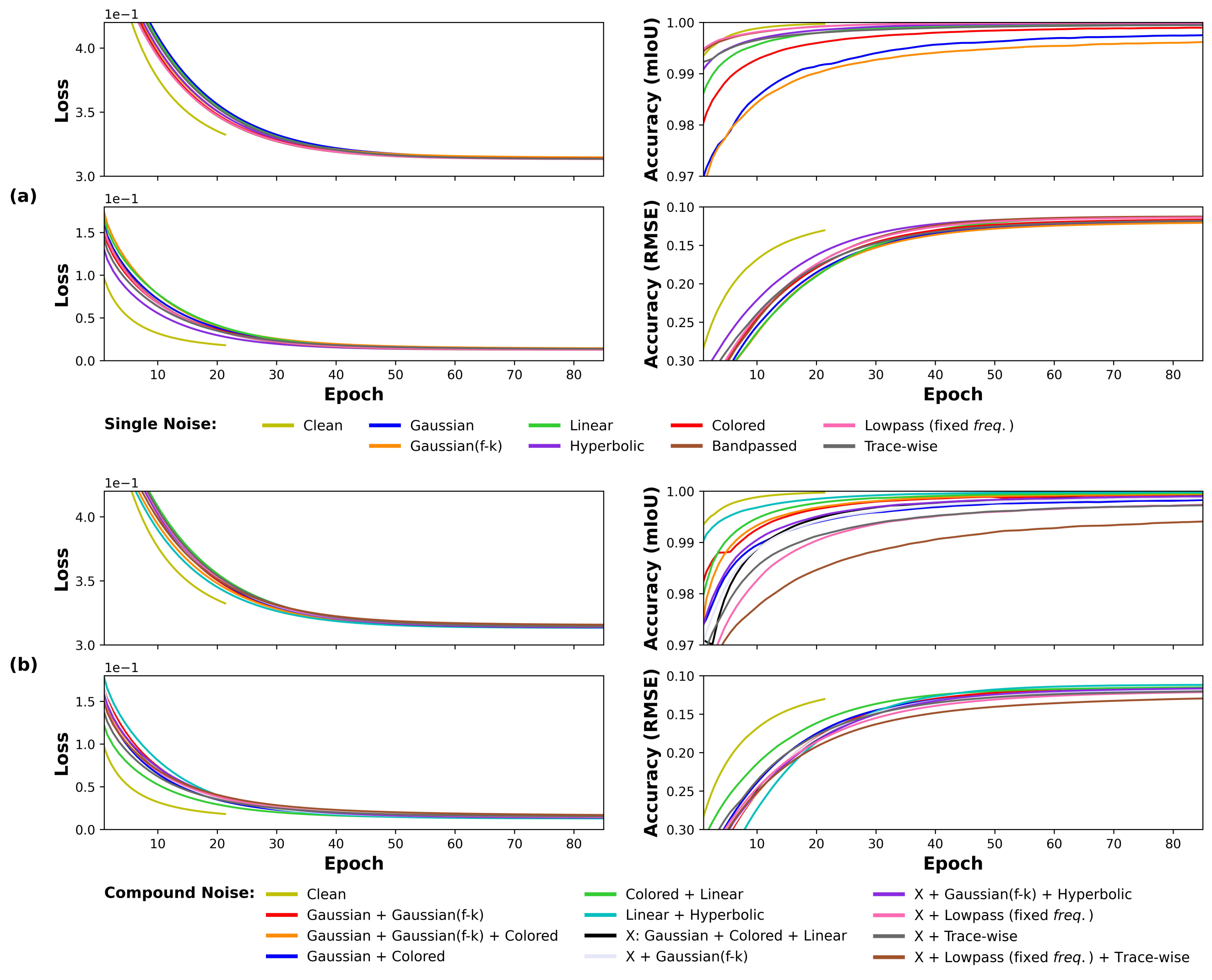} 
    \caption{Mean validation curves of U-Net models trained on (a) single and (b) compound noise types. Top panels show results for first-break picking task, and bottom panels for denoising tasks. Curves are color-coded by noise type. Left column: loss; right column: accuracy.}
    \label{fs15}
\end{figure}

\begin{figure}[ht]
    \centering
    \includegraphics[width=0.7\textwidth]{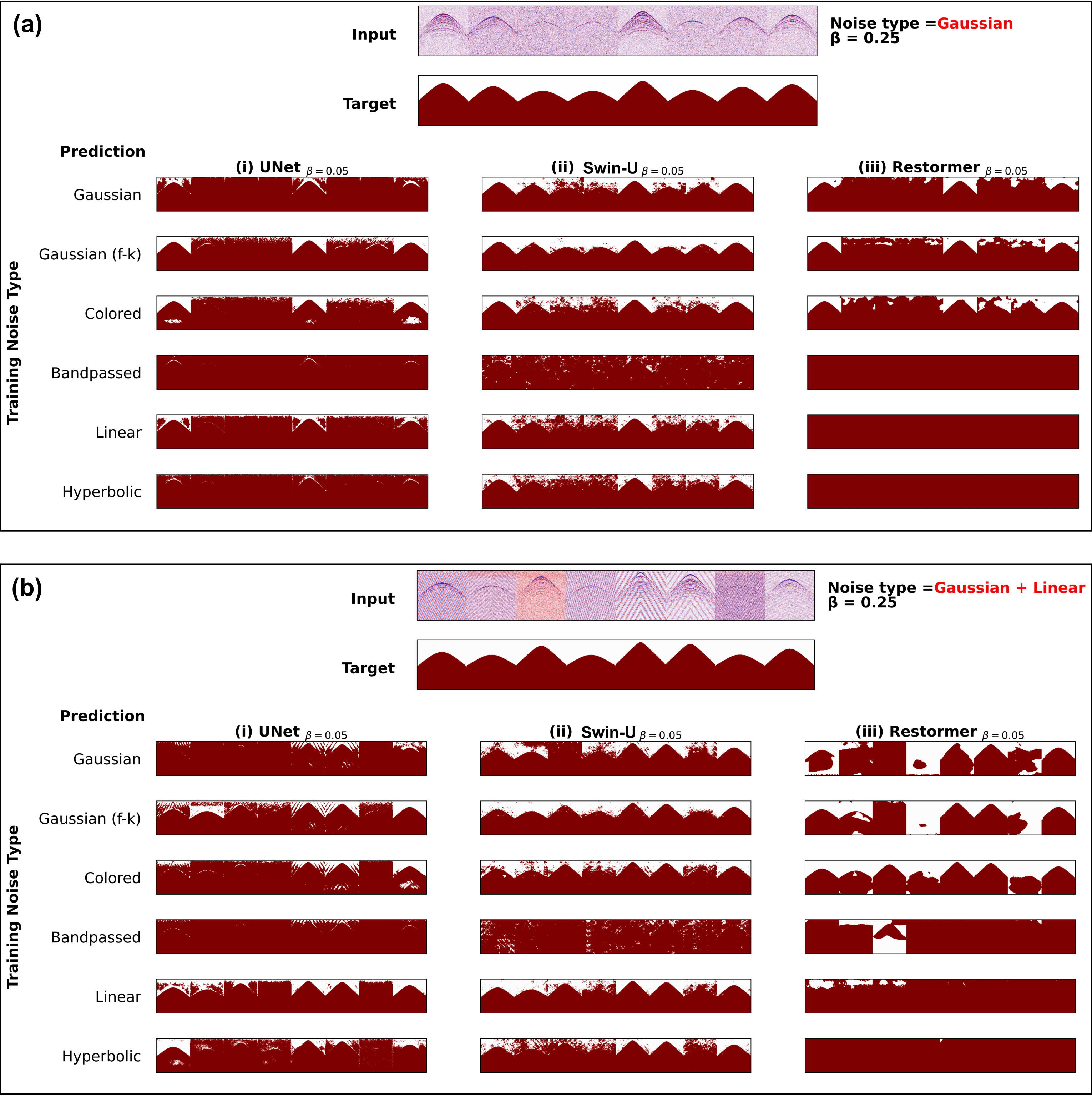} 
    \caption{Effect of low scale ($\beta=0.05$) single noise training on network predictions for the First break picking task. (a) Single Gaussian noise at a higher scale; (b) compound noise at a higher scale.}
    \label{f20_22}
\end{figure}

\begin{figure}[h]
    \centering
    \includegraphics[width=0.7\textwidth]{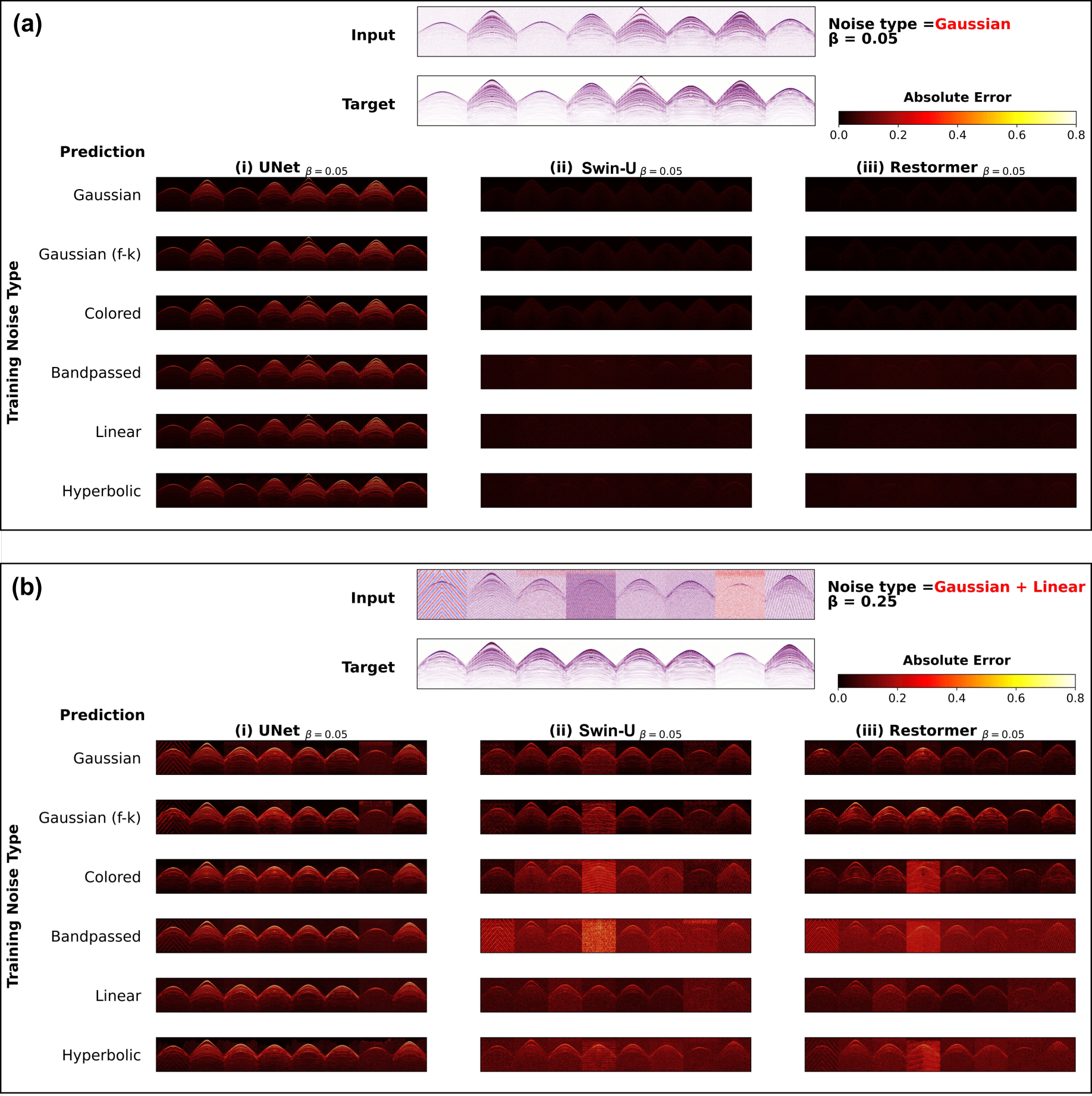} 
    \caption{Effect of low scale ($\beta=0.05$) single noise training on network predictions for the denoising task. (a) Single Gaussian noise at a higher scale; (b) compound noise at a higher scale.}
    \label{f20_22x}
\end{figure}

\begin{figure}[th]
    \centering
    \includegraphics[width=0.7\textwidth]{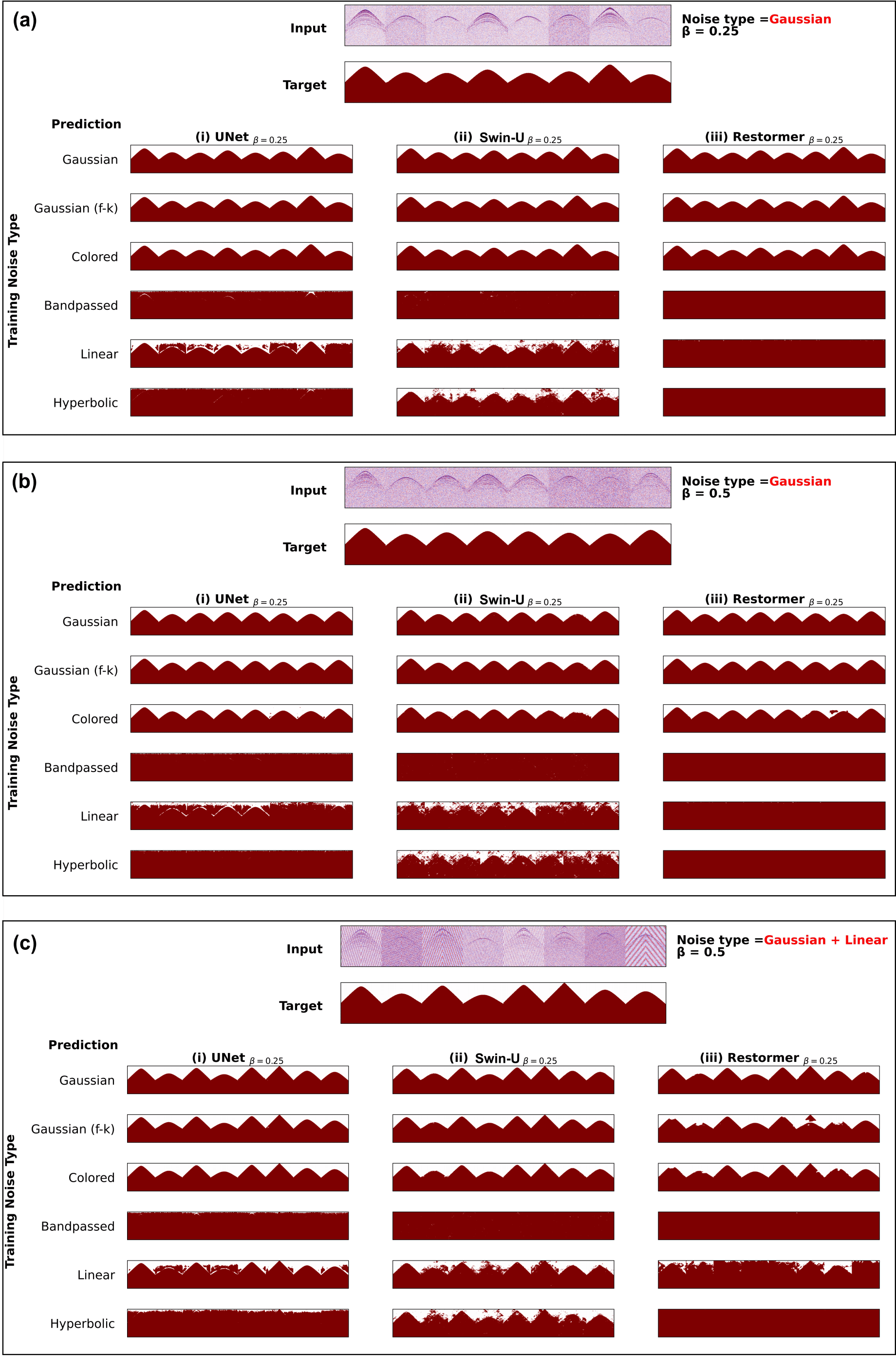} 
    \caption{Effect of medium scale ($\beta=0.25$) single noise training on network predictions on the first break picking task. (a) Single Gaussian noise at the same training scale, (b) single Gaussian noise at a higher scale, (c) compound noise at a higher scale.}
    \label{f24s}
\end{figure}

\begin{figure}[h]
    \centering
    \includegraphics[width=0.7\textwidth]{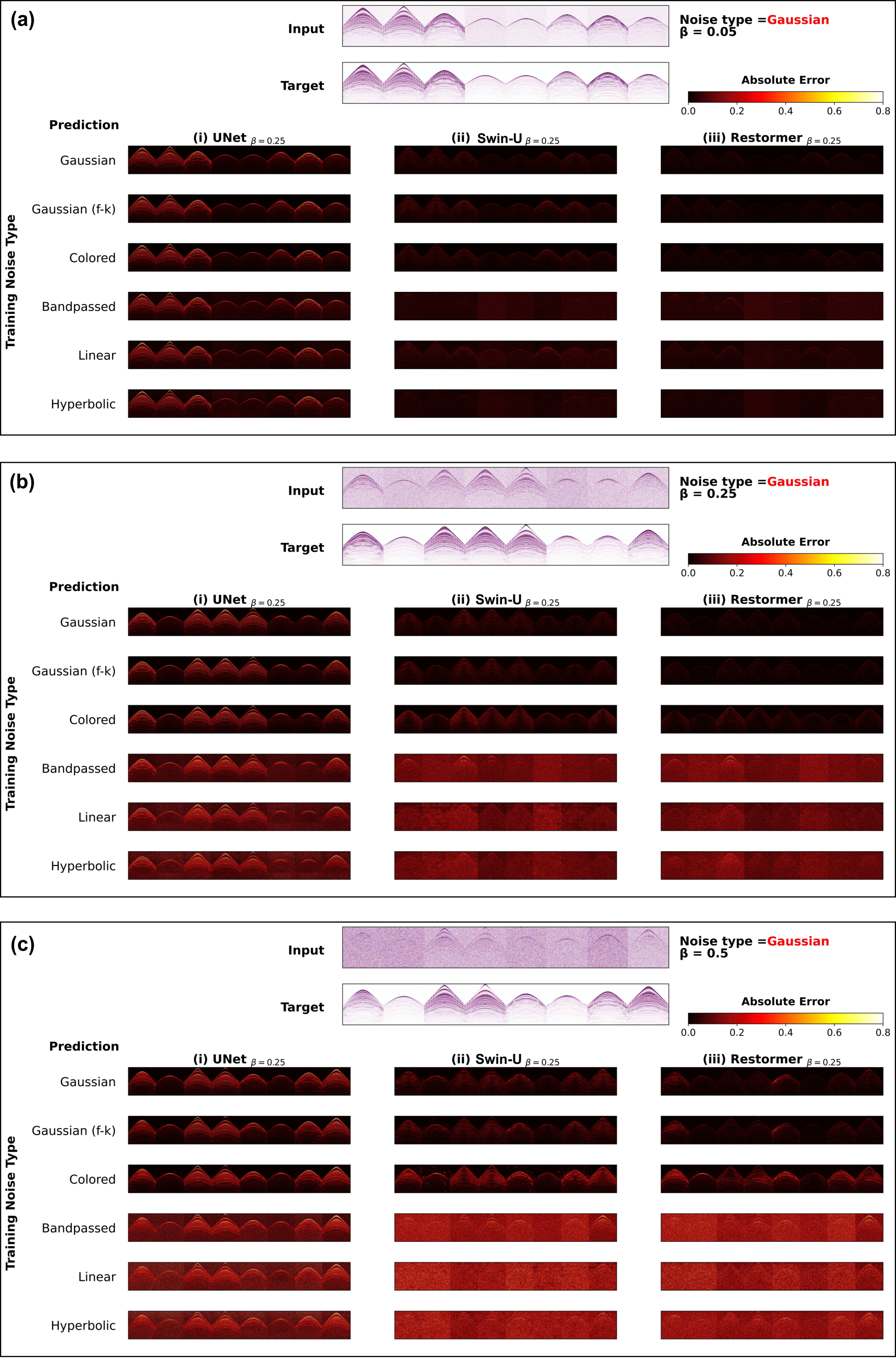} 
    \caption{Effect of medium scale ($\beta=0.25$) single noise training on network predictions for the denoising task. Single Gaussian noise at (a) a lower scale, (b) the same training scale, (c) a higher scale.}
    \label{f26s}
\end{figure}

\begin{figure}[h]
    \centering
    \includegraphics[width=0.7\textwidth]{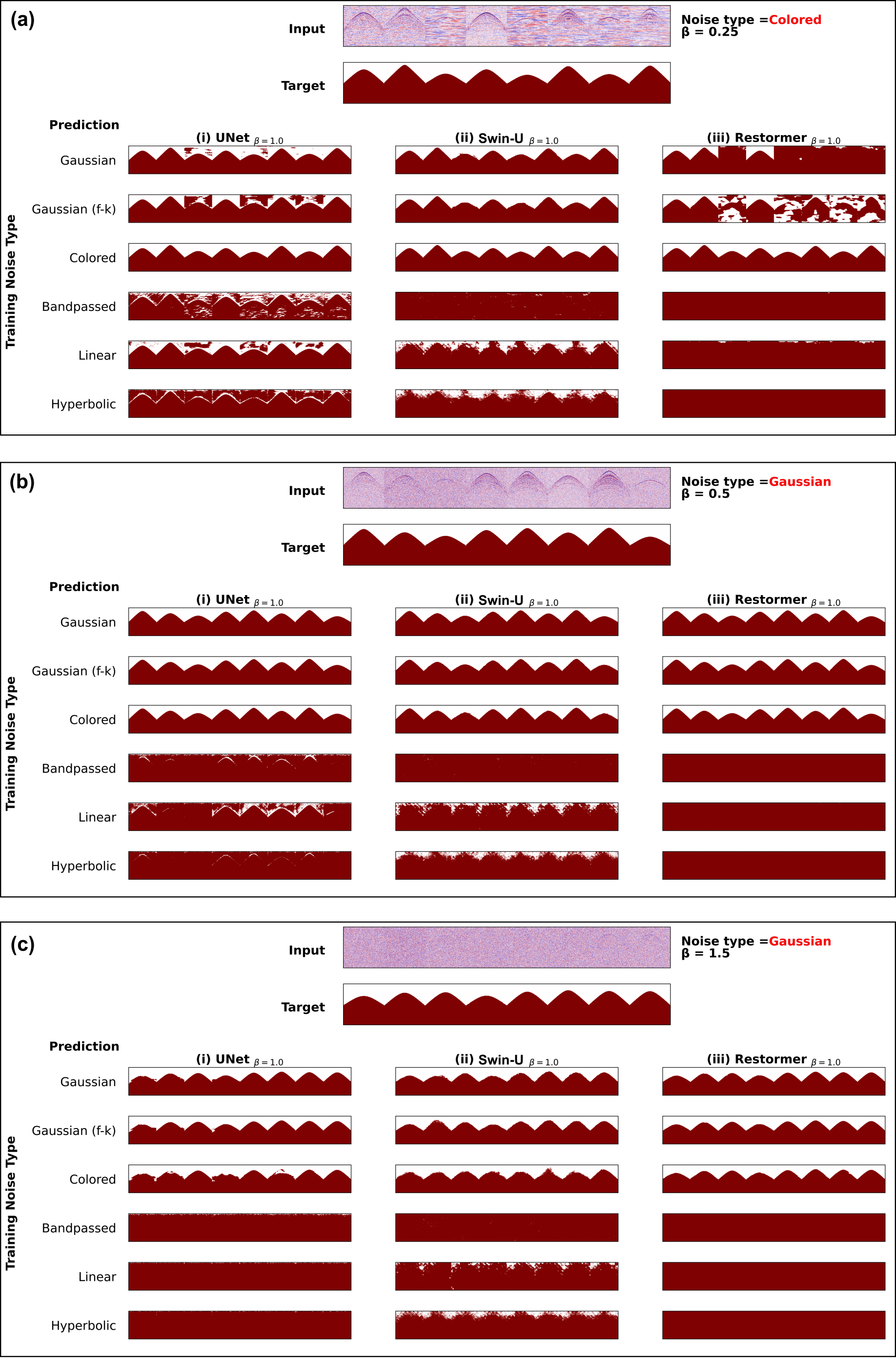} 
     \caption{Effect of high scale ($\beta=1.0$) single noise training on network predictions for the first break picking task. Single Gaussian noise at (a) a lower scale $\beta_{0.25}$, (b) a lower scale $\beta_{0.5}$, (c) a higher scale $\beta_{1.5}$. }
     \label{f29s}
\end{figure}

\begin{figure}[h]
    \centering
    \includegraphics[width=0.7\textwidth]{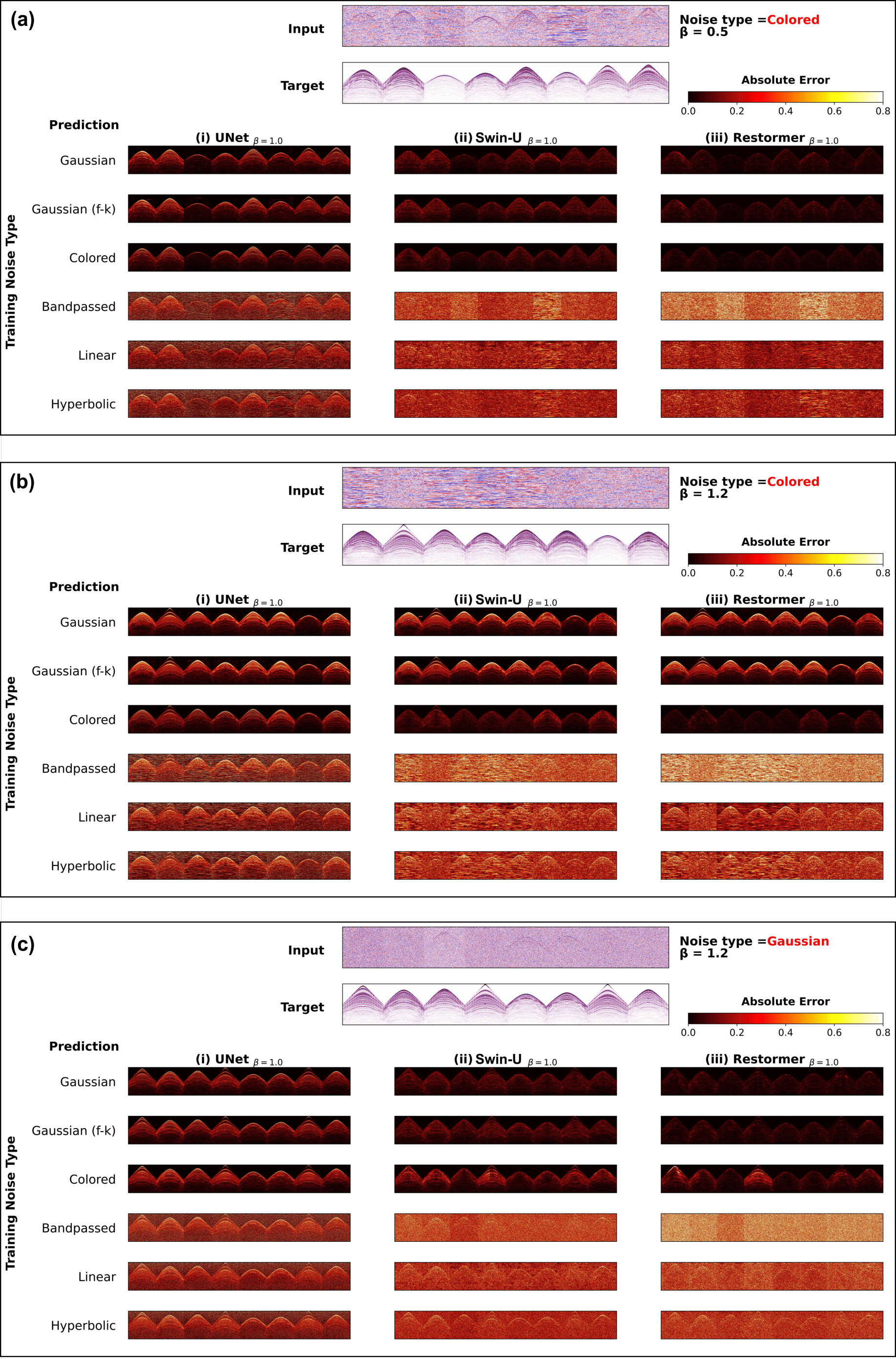} 
     \caption{Effect of high scale ($\beta=1.0$) single noise training on network predictions for the denoising task. (a) a lower scale $\beta_{0.25}$, (b) a lower scale $\beta_{0.5}$, (c) a higher scale $\beta_{1.5}$. Single (a) colored noise at lower scale $\beta_{0.5}$, (b) colored noise at a higher scale $\beta_{1.2}$, (c) Gaussian noise at a higher scale - $\beta_{1.2}$.}
     \label{f30s}
\end{figure}

\begin{figure}[h]
    \centering
    \includegraphics[width=0.7\textwidth]{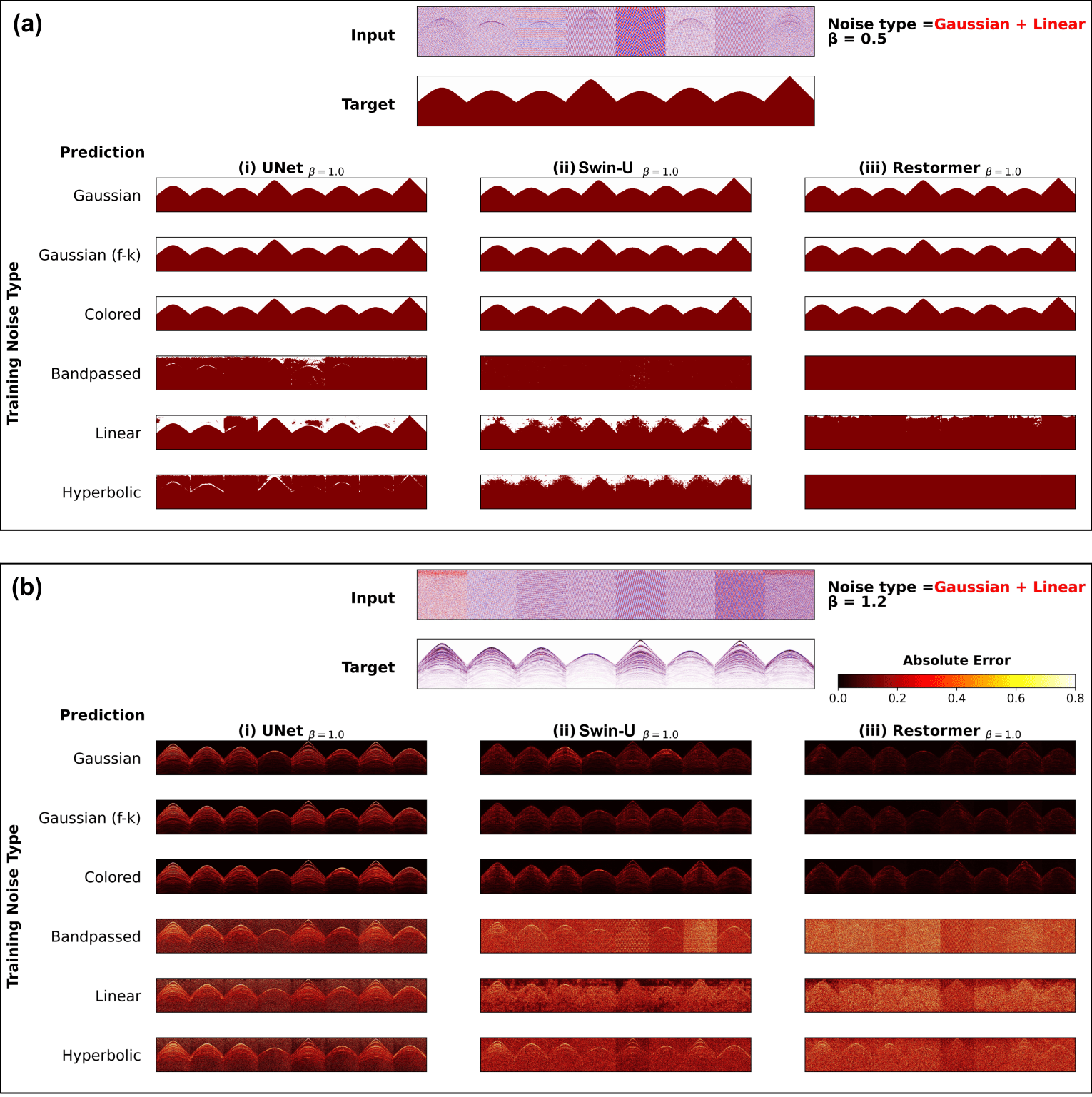} 
    \caption{Effect of high-scale ($\beta = 1.0$) single-noise training on predictions with a structured noise component present in the compound noise. Compound (Gaussian + linear) (a) at a lower scale $\beta_{0.5}$ on the first break picking task , (b) at a higher scale $\beta_{1.2}$ on the denoising task.}
     \label{f31s}
\end{figure}

\begin{figure}[h]
    \centering
    \includegraphics[width=0.7\textwidth]{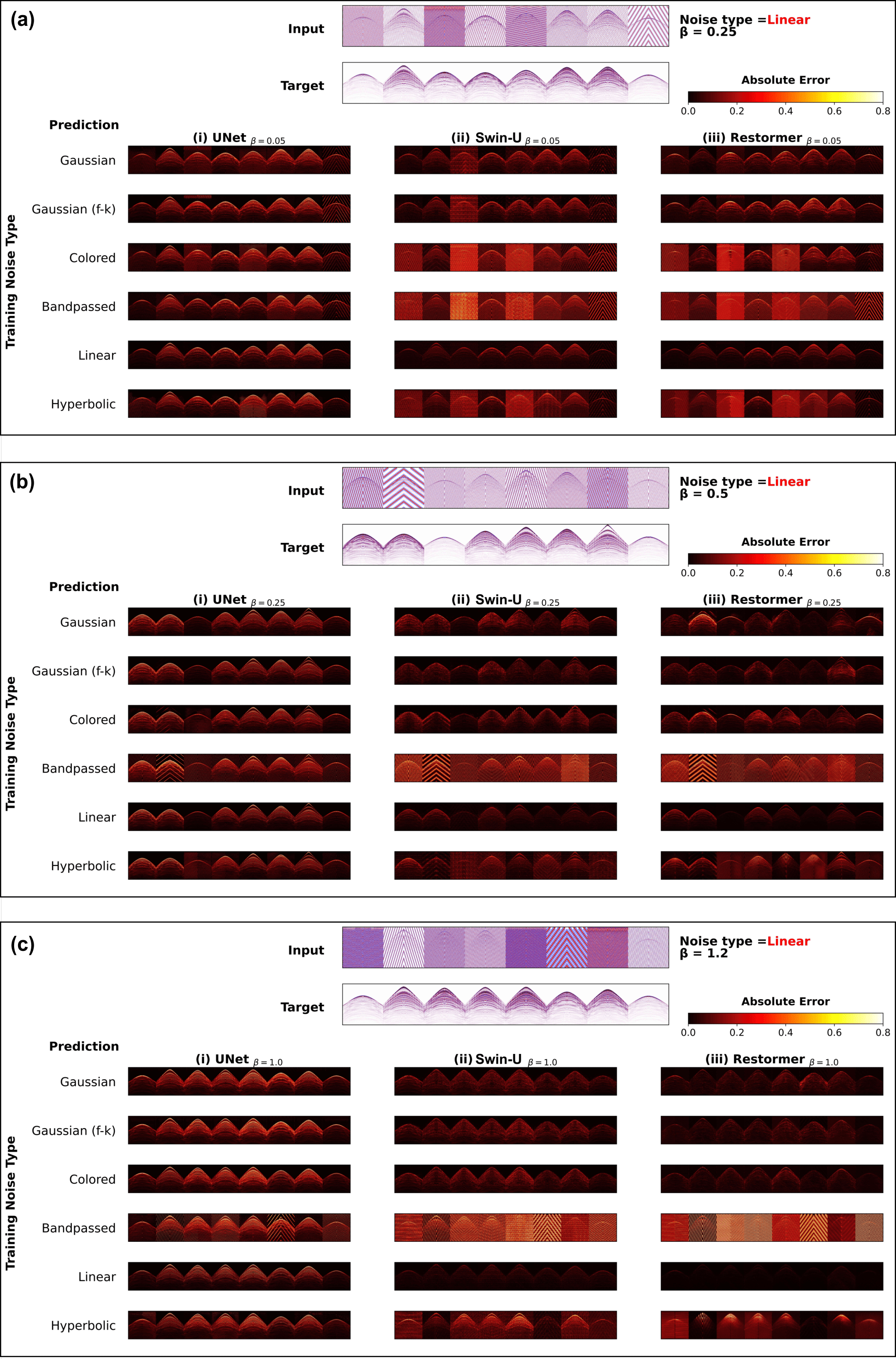} 
     \caption{Impact of training scale ($\beta$) on the performance on higher scale linear noises for the denoising task. (a) Low training scale ($\beta_{0.05}$); (b) medium training scale ($\beta_{0.25}$); (c) high training scale $\beta_{1.0}$}
     \label{f27s}
\end{figure}

\begin{figure}[h]
    \centering
    \includegraphics[width=0.65\textwidth]{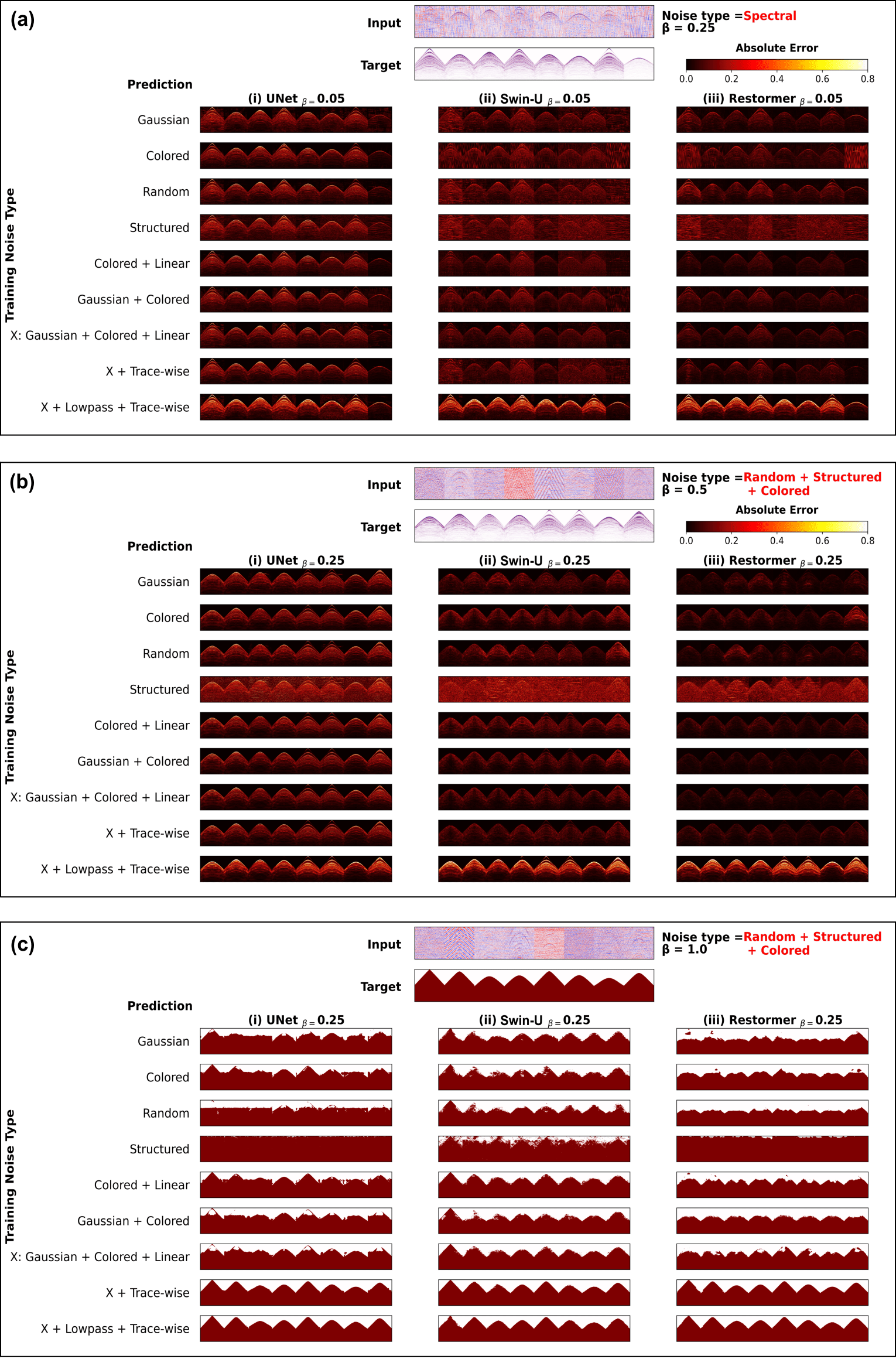} 
     \caption{Effect of training with compound noise at different scales ($\beta$) on predictions under higher-scale compound noise. Tasks: (a-b) denoising, (c) first break picking. (a) Low training scale ($\beta_{0.05}$) - additive compound noise; (b) medium training scale ($\beta_{0.25}$) additive compound noise; (c) medium training scale ($\beta_{0.25}$) additive compound noise. }
     \label{f39s}
\end{figure}

\begin{figure}[h]
    \centering
    \includegraphics[width=0.65\textwidth]{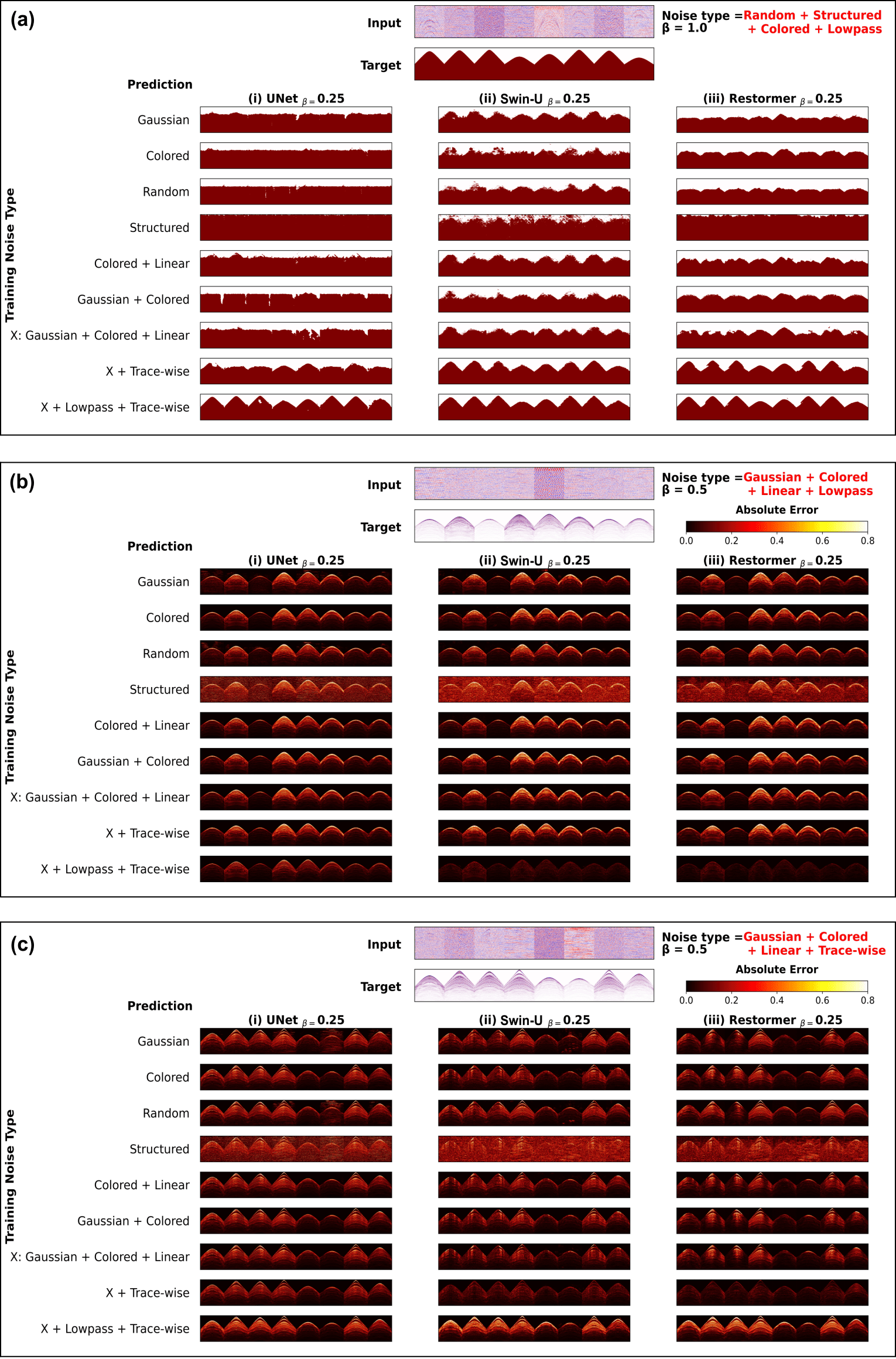} 
     \caption{Effect of training with compound noise at ($\beta =0.25$) on predictions under higher-scale compound noise (additive + data manipulations). Tasks: (a) first break picking, (b) denoising lowpass (c) denoising missing traces.}
     \label{40s}
\end{figure}

\textbf{Contents:}
\begin{itemize}
\item Text \silink{Text1}{S1} to \silink{Text7}{S7}.
\item Figures~\ref{f17} to~\ref{40s}
\item Tables~\ref{T1} to~\ref{T5}
\end{itemize}

This Supporting Information provides the extended methodology, additional analyses, and visual results that accompany the main manuscript. Its purpose is to ensure full reproducibility and to offer deeper insight into how noise variability influences neural network robustness and learning behavior. The file is structured into seven supporting text sections, followed by figures and tables. All results in this Supporting Information were generated under the same dataset, seeds, and training configuration described in the main manuscript. noise sampling remains stochastic by design; sample-level variation is expected, but no instability or anomalies were observed, and all trends remain consistent with the conclusions presented in the paper.

\textbf{Appendix A: Text Sections (S1–S7)}

\begin{itemize}
    \item[\silink{Text1}{S1}.] Synthetic dataset and preprocessing workflow.
    \item[\silink{Text2}{S2}.] Spatial noise characterization using 2-D autocorrelation analysis.
    \item[\silink{Text3}{S3}.] PSNR behavior of the noise forward models across scales and types.
    \item[\silink{Text4}{S4}.] Network architectures, implementation details, and training configuration.
    \item[\silink{Text5}{S5}.] Hyperparameter search (TPE + Hyperband), search space exploration, and selected final settings.
    \item[\silink{Text6}{S6}.] Low-pass filtering strategy and its effect on training dynamics.
    \item[\silink{Text7}{S7}.] Cross-task pre-training for first-break picking.
\end{itemize}

\textbf{Appendix B: Figures (\ref{f17}–~\ref{40s})} 

The figures included in this document illustrate and expand upon the core results presented in the manuscript, and are organized into groups:
\begin{itemize}
    \item[\ref{f17}–~\ref{f36}]: Full robustness matrices extending the main manuscript results for all architectures under single- and compound-noise training across both tasks.
    \item[\ref{sf0}-\ref{sfx0}]: Spatial noise structure visualizations and 2-D autocorrelation maps.
    \item[\ref{f2}–~\ref{f7x}]: PSNR distributions for single- and compound-noise forward models across noise scales.
    \item[\ref{f41}–~\ref{f41x}]: Hyperparameter optimization landscapes and comparative training curves.
    \item[\ref{f11}–~\ref{sf50}]: Low-pass filtering behavior and variability effects during training.
    \item[\ref{sf2}–~\ref{sf3}]: Cross-task pre-training results for first-break picking.
    \item[\ref{fs12}–\ref{fs15}]: Extended in-distribution training and performance visualizations for the remaining architectures across noise regimes and tasks.
    \item[\ref{f20_22x}-~\ref{40s}]: Additional qualitative predictions for all architectures across noise regimes and tasks.
\end{itemize}

\textbf{Appendix C: Tables (T1–T5)}

\begin{itemize}
    \item[T\ref{T1}.]  Dataset statistical profile (raw vs processed).
    \item[T\ref{T2}.]  Optimal hyperparameters from TPE search.
    \item[T\ref{T3}.]  Mean in-distribution (ID) performance across noise parameters, networks, and tasks.
    \item[T\ref{T4}.]  Mean robustness of networks trained on a single noise type across both tasks.
    \item[T\ref{T5}.]  Mean robustness of networks trained on compound noise type across both tasks.
\end{itemize} 

\clearpage

\subsection{Appendix A: Text Sections}

\subsubsection{Text S1: Data Preprocessing Pipeline} ~\label{Text1}
The experiments in this study were conducted using a clean synthetic seismic dataset consisting of 2121 common-shot gathers of size $241\times1001$, generated from acoustic forward modeling. The raw data were stored in SEG-Y format and converted to NumPy arrays for efficient processing. The amplitude distribution is highly peaked around zero, with most samples concentrated within the narrow interval $[\minus0.14,0.14]$, with rare long-tailed outliers ($\pm1.6$). The binary segmentation masks were constructed from the first break arrival times (ASCII format). For each trace, the first break pick was normalized to the temporal sampling of the data, and a binary label was generated by marking all samples preceding the arrival time as 1 and all remaining samples as 0, resulting in masks of shape $ 2121\times241\times1001$. 
To prepare the data for training, both inputs and labels were spatially resized to $2121\times224\times224$ using area-preserving interpolation, corresponding to down-sampling factors of 1.08 (height) and 4.47 (width). The resized seismic data were then amplitude-clipped to $\pm 4$ to suppress extreme outliers while retaining meaningful dynamic range, followed by normalization to the range $[\minus1,1]$. After preprocessing, the data exhibited a tightly concentrated distribution around zero with interquartile bounds of approximately $[\minus0.078, 0.074]$, while extreme values ($\pm0.83$) appeared only rarely (Table~\ref{T1}). Each gather and its corresponding label were finally stored as individual 2D NumPy slices for efficient loading during training.
The dataset was split into training/validation and test sets using a fixed random seed (911), allocating $10\%$ of the samples to testing and reserving the remaining $90\%$ for model development. The final training/validation split was performed dynamically within the data loader using the same fixed seed to ensure reproducibility. The data and code repository can be found in the \textit{Open Research Section}. 

\subsubsection{Text S2: Noise Spatial Characteristics}~\label{Text2}
The noise generated in this study exhibits distinct spatial characteristics, which influence the structural patterns in the observed images. To quantitatively assess these effects, we analyze the 2D autocorrelation function of the images under various noise conditions. This analysis provides insight into the degree of spatial dependency introduced or disrupted by each noise type, and helps to distinguish between structured and unstructured noise behaviors.

Figure~\ref{sf0} show the spatial autocorrelation of single noises. Sub-figure (a) presents the raw noise patterns, while (b) and (c) display the corresponding spatial correlations at scales $\beta=0.25$ and $\beta=1.0$, respectively. Random noises (Gaussian (x-t) and Gaussian (f-k)) exhibit no spatial correlation, while spectral noises (colored and bandpassed) display structured correlations, often appearing as linear features. Structured noises (linear and hyperbolic), show distinct and coherent patterns across the image, reflecting their underlying design. 

Figure~\ref{sfx0} illustrates the visual appearance (top row) and corresponding 2D autocorrelation functions (bottom row) of seismic images corrupted with different compound noises. Random noise (R) shows no discernible spatial structure, with its autocorrelation limited to a central spike, indicating an absence of spatial correlation. In contrast, structured noise (S) and spectral noise exhibit strong and anisotropic spatial correlations, visible as repeating horizontal or vertical patterns in their autocorrelation maps. The combination of Gaussian and colored noise (X) introduces mild correlation, while adding linear trends further enhances directional structure. Notably, the compound noise (R + S + Colored) results in complex and coherent autocorrelation patterns, even at low intensity. This demonstrates that compound noise is not merely an additive sum of independent sources, but produces emergent spatial structure due to the interaction between multiple noise types. The compound noises cause image distortions that vary across different SNR levels, exhibiting an inverse relationship with the noise scale $\beta$ (Figure~\ref{f3}).

\subsubsection{Text S3: Analysis of PSNR of Noise Forward Models}~\label{Text3}
The noise generators in this work provide diverse realizations of each noise type, which are then scaled by a noise amplitude factor $\beta$ and added to the output of the data model $F$ (Figure~\ref{f2}). The parameter $\beta$ offers additional control over the noise level, allowing for systematic variation of the output's SNR. The resulting distributions remain sharply peaked around zero, indicating that the noise perturbations are relatively minor. Equation \ref{eq4} relates the noise scale $\beta$ to SNR, where an increases in $\beta$ reduces the SNR exponentially (Figure~\ref{f3}).
\begin{linenomath*}
    \begin{equation} \label{eq4}
        SNR\ \left(dB\right)=10\log_{10}{\left(\ \frac{1}{\beta\ }\right)} \ .
    \end{equation}
\end{linenomath*}

\begin{figure}[h] 
\centering\includegraphics[width=0.75\textwidth]{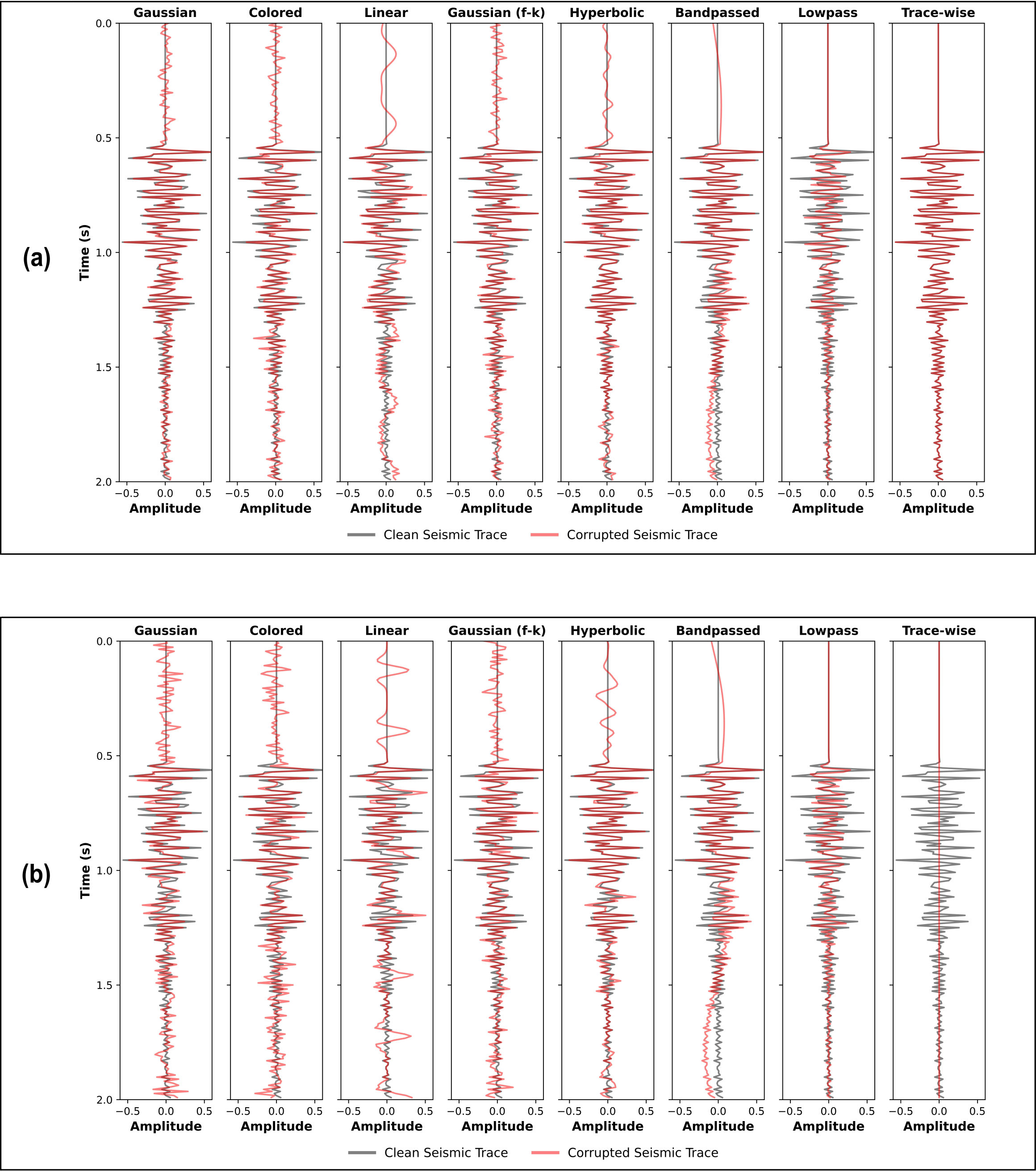}
\caption{1-D view of a random trace after augmenting with single noises. (a) $\beta =$ 0.05; (b) $\beta=$ 0.1.}
\label{f4}
\end{figure}

Figure~\ref{f4} showcases the effect of varying $\beta$ on the amplitude of the single noise perturbation applied on a random trace. At low noise levels ($\beta$=0.05) random noise appear mainly as pre-event disturbances without drastically changing the signal. In contrast, coherent noises, such as linear noise, seems to slightly alter the first arrival of the P-wave at low noise levels (Figure~\ref{f4}a). As for data manipulations produced by $F$ (last two columns) they primarily effect the amplitude of the signal but do not impact the P-wave arrival. By increasing $\beta$ to 0.1 (Figure~\ref{f4}b), the characteristic of the signal begins to change, adding complexity to the tasks of removing noise, reconstructing the original amplitudes, and identifying the first arrivals.

Figure~\ref{f5} showcase the distribution in image quality, measured in Peak Signal-to-Noise Ratio (PSNR) (Equation~\ref{eq7}), for 500 random noise realizations (per $\beta$) applied on the same shot image. As $\beta$ increases, the quality of the images reduces. The Gaussian distribution of PSNR indicates that at each noise level $\beta$, the generator produces variable noise patterns and levels of distortion. There are six distinct $\beta$ levels, each corresponding to a defined range of image distortion ($\beta$ =0.05, 0.1, 0.25, 0.5, 1.0, 2.0). The overlap between classes at low noise scales suggests that the noise introduced by these generators have a similar impact on signal quality. Consequently, we expect that networks trained with a specific $\beta$ will generalize to the overlapping classes that display similar noise characteristics. The introduction of data manipulations by F, such as removing some of the frequency components of the signal or masking random traces (as depicted in the last column in  Figure~\ref{f5}b) leads to a greater reduction in the PSNR of the image at low noise ($\beta$ =0.05) and shifts the peaks of low $\beta$ classes closer. However, at higher noise levels, the missing traces become indistinguishable as the additive noise effectively fills the gaps created by the mask.  These unique $\beta$ classes will be utilized as assessment intervals for evaluating network performance and generalizability.
 
\begin{linenomath*}
    \begin{equation} \label{eq7}
        PSNR(x,\Tilde{x})=\ 10\log_{10}{\frac{ \Vert x \Vert _2 ^2}{\Vert x - \Tilde{x} \Vert_2 ^2}} \ \ ,  \Tilde{x} = x + \xi \ .
    \end{equation}
\end{linenomath*}

The 1D view in Figure~\ref{f6} demonstrates the effect of compound noises in altering the P-wave arrival and signal amplitude. Even though the number of additive noises increases progressively from one column to the next, the effect of noise type is less pronounced compared to single noises (Figure~\ref{f4}). This can be attributed to the interference between noises, leading to effects that are less intense or pronounced than they would be if the noises were acting independently. In comparison to single noises (Figure~\ref{f4}), compound noises generate more complex signal distortion patterns, especially when combined with data manipulations by $F$.  The implementation of a masking strategy is known to increase the difficulty in the training task, encouraging the network to learn contextual information and enhancing their weights \cite{Vinvenet, pathak2016contextencodersfeaturelearning,devlin2019, Park_2019,he2021,chen2024}, which can be beneficial when training on simpler tasks such as first break picking. Figure~\ref{f7x} displays the effect of increasing $\beta$ on some of these intrinsic single and compound noise augmentations. As $\beta$ increases, the mages becomes more occluded by noise.

\subsubsection{Text S4: Architecture and Training Parameters}~\label{Text4}
All components of the proposed framework—including data preprocessing, model architectures, and training procedures—were implemented using the PyTorch library. Below we outline any modifications done on the original architectures. 

Our U-Net follows the PyTorch implementation of \cite{buda2019association}, where we retain the same feature-map configuration as in Buda’s code, while adapting the model to our seismic setting by using single-channel inputs and either two output channels for first-break segmentation or one output channel for denoising. All architectural components, including padded 3$\times$3 convolutions, Batch Normalization, ReLU activations, and the skip-connection structure, are kept unchanged.

The only modification we introduce is the removal of the hard-coded sigmoid activation from the final 1$\times$1 convolution layer present in the reference implementation. This activation is not required because the PyTorch loss used for segmentation \textit{nn.CrossEntropyLoss} internally applies the appropriate log-softmax over classes. Accordingly, the output layer in our implementation uses an identity mapping.

For the Swin-Unet architecture, we utilize the publicly available Swin-Unet implementation by \cite{cao2021swinunetunetlikepuretransformer}. We use the same encoder–decoder architecture and Swin-Transformer blocks, but adapt it to single-channel seismic inputs and task-specific output heads. In our experiments, we use a reduced patch-embedding dimension of 48 (instead of the original 96) to obtain a lighter Swin-T–style backbone \cite{liu2021swintransformerhierarchicalvision} while preserving the hierarchical structure and shifted-window self-attention. 

For the Restormer architectures, we utilize the code implementation by 
\cite{zamir2022restormerefficienttransformerhighresolution}, but reduce the embedding dimension from 48 to 24, yielding a lighter 24–48–96–192 hierarchy instead of the original 48–96–192–384 configuration, while keeping all architectural components unchanged. Restormer includes two Layer Normalization variants (BiasFree and WithBias) used differently across tasks in the original implementation. In our experiments, we adopt the WithBias configuration, which differs from the BiasFree variant typically used for denoising, but remains consistent with the configurations employed for other restoration tasks (e.g., de-raining and de-blurring).

For Swin-UNet, the linear layers are initialized with truncated normal weights (std $= 0.02$) and zero bias, and the LayerNorm parameters are initialized with unit scale and zero bias, following the standard Swin Transformer practice \cite{liu2021swintransformerhierarchicalvision}.  In contrast, U-Net and Restormer were left with the default PyTorch weight initialization, which uses Kaiming (He) initialization ~\cite{he2015delvingdeeprectifierssurpassing} for convolutional and linear layers, and sets BatchNorm and LayerNorm parameters to unit scale and zero bias.

\subsubsection{Text S5: Hyperparameter Search}~\label{Text5}
To identify suitable optimization hyperparameters, we performed an automated search using the Tree-Structured Parzen Estimator (TPE) \cite{bergstra} algorithm implemented in Optuna \cite{optuna_2019}. The TPE sampler (seed = 42) was run for 40 optimization trials, with multivariate sampling enabled to capture interactions between hyperparameters and the constant-liar strategy used to ensure stable, parallel-safe trial generation. To accelerate the search and enable efficient early stopping of underperforming trials, we employed a Hyperband pruner configured with a minimum resource of 1, a maximum resource of 30 (equal to the total number of training epochs), a reduction factor of 3, and a bootstrap count of 1. Each trial was trained for up to 30 epochs using mixed-precision (AMP) and early stopping with a patience of 3 epochs.

The search was conducted independently for each network and task combination. The learning rate was optimized over a logarithmic range lr$\in\{1e^{-5}, 1e^{-3}\}$, and, when enabled, the batch size was selected from the discrete set $\{8,16,32\}$ . All other training hyperparameters, including the optimizer (Adam)\cite{kingma2017adammethodstochasticoptimization}, weight decay ($\beta_1=0.9, \ \beta_2=0.999$), and the default architectural nonlinearities were kept fixed according to the original network specifications.

The objective function minimized the validation loss for the corresponding task and reported intermediate results to the TPE sampler at every epoch. Trials whose intermediate performance fell below the Hyperband pruning threshold were terminated early. Figure~\ref{f41} illustrates the validation loss across the explored search space, where both completed and pruned trials are displayed to provide full coverage. However, only completed trials were used when selecting the final hyperparameters for training.

For each architecture–task pair, the optimal hyperparameters were first identified as those producing the lowest validation loss among completed trials. For consistency in subsequent training experiments in this paper, we then selected a unified set of hyperparameters per task by choosing the minimum-loss configuration across all three architectures. This procedure yielded an average learning rate of $5e^{-5}$, which performed effectively for both tasks. Transformer-based architectures (Swin-U and Restormer) generally favor lower learning rates for stability, whereas U-Net can typically accommodate higher learning rates. For U-Net, training with the unified learning rate resulted in a final accuracy of 0.116 at epoch 30, compared with 0.113 when using the architecture-specific optimal learning rate on the denoising task (Figure~\ref{f41x}b). While the optimal learning rate identified by TPE converges more rapidly during the initial training epochs, the resulting performance plateau is nearly identical to that of the unified learning rate. Thus, the practical benefit of the optimal setting is confined to training speed rather than any substantive improvement in model accuracy or generalization.

Similarly, while the hyperparameter search suggested that a batch size of 16 yielded a slightly lower validation loss for U-Net, the improvement was marginal and did translate into meaningful performance gains (Figure~\ref{f41x}a). Although larger batch sizes can in principle be advantageous, using a batch size of 16 consistently across all architectures was not practical due to the considerably higher memory requirements of the Restormer model. Therefore, to ensure consistency, reproducibility, and compatibility with hardware constraints, we adopted a unified batch size of 8 for all final experiments, which provided stable and competitive performance across all networks and tasks.

The hyperparameter search was performed on the clean synthetic data described in this paper, without applying any additional stochastic noise augmentation. A complete listing of the optimal hyperparameter configurations obtained for each architecture–task pair is provided in Table~\ref{T2}.

\subsubsection{Text S6: Low-pass Filter}~\label{Text6}
Training with variable low-pass filters introduces persistent difficulty throughout the learning process, leading to slower convergence and slightly worse final performance compared to training with a fixed filter (Figure~\ref{f11}). However, this variability serves an important purpose as it prevents the model from overfitting to a particular low-pass characteristic and thereby improves generalization to unseen filtering conditions \cite{geirhos2020generalisationhumansdeepneural,Sulun_2021}. This highlights an inherent trade-off between accuracy and generalizability, illustrating how noise variability influences the learning dynamics of neural networks.

A visual illustration of the effects of different low-pass filters is provided in Figure~\ref{sf50}. In all cases, the cutoff frequency is nominally fixed but stochastically resampled from a narrow interval for each image, introducing mild variability while remaining within the same spectral neighborhood (Figure~\ref{sf50}a–b). The variable-filter strategy (gray dashed curve in Figure~\ref{f11}) further amplifies this effect; the cutoff frequency is randomly sampled for each batch (with the option to resample per image), exposing the network to a wider range of filtering conditions and consequently increasing training difficulty. 

\subsubsection{Text S7: Cross Task Pre-training (First break picking)}~\label{Text7}
We explore the benefits of initializing the training process with pretrained weights, as opposed to starting from scratch (random weights) on the desired task. Specifically, we utilize pretrained weights from the regression task to enhance the segmentation task. Our results reveal that initializing the first break task with weights obtained from the denoising task significantly improves the learning process. This enhancement is likely due to the complexity of the denoising task, which yields more informative training and weight updates compared to the first break picking task. This effect is particularly evident when using the Swin-U architecture, which contributes to stabilizing the training process and minimizes the risk of performance degradation, especially at high noise scales ($\beta>1.0$). Results of the ablation tests can be found Figures~\ref{sf2} to~\ref{sf3}.
\begin{notation}
\notitem{\(\beta\):}{noise scale.}
\notitem{ID:}{in-distribution.}
\notitem{OOD:}{out-of-distribution.}
\notitem{SNR:}{signal-to-noise ratio.}
\notitem{PSNR:}{peak signal-to-noise ratio.}
\notitem{\(f\)-\(k\):}{frequency--wavenumber domain.}
\end{notation}
\newpage
\subsection{Appendix B: Figures}

\subsubsection{Additional Generalization Figures}
\noindent\textit{Single Noise:}
\clearpage
\clearpage
\clearpage
\clearpage
\noindent\textit{Compound Noise:}
\clearpage
\subsubsection{Noise Spatial Characteristics}

\clearpage
\subsubsection{Analysis of PSNR of Noise Forward Models}
\clearpage
\subsubsection{Hyperparameter Search}
\clearpage
\subsubsection{Low-pass Filter}
\clearpage
\subsubsection{Cross Task Pre-training (First break picking)}
\clearpage
\subsubsection{Additional ID Performance Figures}
\clearpage
\subsubsection{Visual Results}
\noindent\textit{Single Noise:}
\clearpage
\noindent\textit{Compound Noise:}
\clearpage
\subsection{Appendix C: Tables}
\begin{table}[ht]
\centering
\caption{Statistical summary of the synthetic dataset before and after preprocessing.}
\begin{tabular}{lcc}
\hline
\textbf{Statistic} & \textbf{Raw Data} & \textbf{$^\star$Processed Data} \\
\hline
Size & $2121\times241\times1001$ & $2121\times224\times224$ \\
Min & $-407.97$ & $-404.44$ \\
Max & $1411.20$ & $404.44$ \\
Mean & $5.71$ & $5.90\times 10^{-5}$ \\
Std. Dev. & $1.70$ & $0.920$ \\
5th Percentile & $-1.603$ & $-0.836$ \\
25th Percentile (Q1) & $-0.142$ & $-0.0785$ \\
Median (Q2) & $0.0$ & $0.0$ \\
75th Percentile (Q3/MAD) & $0.135$ & $0.074$ \\
95th percentile (Q4) & $1.593$ & $0.828$ \\
\hline
\multicolumn{3}{l}{\footnotesize $\star$ Prior to amplitude clipping and normalization.}
\end{tabular}
\label{T1}
\end{table}

\begin{sidewaystable}[ht]
\centering
\caption{Best hyperparameters selected from completed TPE optimization trials for each architecture–task pair.}

\begin{tabular}{llllll}
\hline
\textbf{Problem} & \textbf{Model} & \textbf{Loss} & \textbf{Learning rate} & \textbf{Batch size} & \textbf{Optimized hyperparameters} \\
\hline
        & U\textsc{-}Net   & $1.29\times10^{-2}$  & $9.85\times10^{-4}$ & 8  & learning rate, batch size \\
        & U\textsc{-}Net   & $1.29\times10^{-2}$  & $9.04\times10^{-4}$ & 8  & learning rate \\[1mm]

Denoise & Swin-U           & $8.58\times10^{-6}$  & $6.71\times10^{-4}$ & 8  & learning rate, batch size \\
        & Swin-U           & $1.18\times10^{-5}$  & $4.00\times10^{-4}$ & 8  & learning rate \\[1mm]

        & Restormer        & $\boldsymbol{1.57\times10^{-7}}$  & $\boldsymbol{4.06\times10^{-5}}$ & \textbf{8}  & learning rate, batch size \\
        & Restormer        & $3.77\times10^{-7}$  & $9.71\times10^{-4}$ & 8  & learning rate \\
\hline
        & U\textsc{-}Net  & $1.74\times10^{-3}$  & $1.48\times10^{-4}$ & 16 & learning rate, batch size \\
        & U\textsc{-}Net  & $3.13\times10^{-1}$  & $9.51\times10^{-4}$ & 8  & learning rate \\[1mm]
First break & Swin-U          & $1.74\times10^{-3}$  & $1.48\times10^{-4}$ & 16 & learning rate, batch size \\
        & Swin-U          & $\boldsymbol{1.50\times10^{-3}}$  & $\boldsymbol{6.09\times10^{-5}}$ & \textbf{8}  & learning rate \\[1mm]
        & Restormer       & $1.95\times10^{-3}$  & $5.69\times10^{-4}$ & 8  & learning rate, batch size \\
        & Restormer       & $1.76\times10^{-3}$  & $1.93\times10^{-4}$ & 8  & learning rate \\
\hline
\multicolumn{6}{l}{\footnotesize Unified parameters used in training experiments: learning rate $5\times10^{-5}$, batch size $8$.}
\end{tabular}
\label{T2}
\end{sidewaystable}

\begin{sidewaystable}[htb]
\centering
\caption{Mean ID performance metrics on the validation set}
\resizebox{\textwidth}{!}{
    \begin{tabular}{cc|ccc|ccc}
        \hline
        \multicolumn{2}{c|}{} & \multicolumn{3}{c|}{\textbf{First break Picking$^{*}$ (mIoU)}} & \multicolumn{3}{c}{\textbf{Denoising$^{**}$ (RMSE)}} \\
        \multicolumn{1}{c}{Noise} & \multicolumn{1}{c|}{Network} & $\beta_{Low}$ & $\beta_{Medium}$ & $\beta_{High}$ & $\beta_{Low}$ & $\beta_{Medium}$ & $\beta_{High}$ \\
        \hline
        \multicolumn{1}{c|}{} & U-Net   & $99.98\pm0.02$ & $99.97\pm0.02$ & $99.61\pm0.55$ & $0.114\pm0.001$ & $0.114\pm0.001$ & $0.118\pm0.007$ \\
        \multicolumn{1}{c|}{\textbf{Single}} & Swin-U    & $99.88\pm0.07$ & $99.85\pm0.06$ & $97.85\pm2.58$ & $0.021\pm0.012$ & $0.026\pm0.014$ & $0.048\pm0.031$ \\
        \multicolumn{1}{c|}{} & Restormer & $99.96\pm0.02$ & $99.94\pm0.02$ & $99.66\pm0.44$ & \textbf{$0.009\pm0.006$} & \textbf{$0.012\pm0.007$} & \textbf{$0.025\pm0.021$} \\
        \hline
        \multicolumn{1}{c|}{} & U-Net & $99.96\pm0.03$ & $99.93\pm0.05$ & $99.11\pm0.77$ & $0.115\pm0.002$ & $0.117\pm0.003$ & $0.128\pm0.006$ \\
        \multicolumn{1}{c|}{\textbf{Compound}} & Swin-U   &  $99.84\pm0.09$ & $99.76\pm0.14$ & $94.95\pm2.16$ & $0.030\pm0.011$ & $0.037\pm0.009$ & $0.086\pm0.024$ \\
        \multicolumn{1}{c|}{} & Restormer & $99.93\pm0.04$ & $99.90\pm0.05$ & $99.21\pm0.51$ & \textbf{$0.016\pm0.008$} & \textbf{$0.021\pm0.009$} & \textbf{$0.049\pm0.020$} \\
        \hline
        \multicolumn{8}{l}{$\beta_{Low} =[0.05\text{-}0.15]$, $\beta_{Medium} =[0.2\text{-}0.5]$, $\beta_{High} =[1.0\text{-}2.0]$} \\
        \multicolumn{8}{l}{$^{*}50$ Epochs, $^{**}100$ Epochs} \\
    \end{tabular}
}
\label{T3}
\end{sidewaystable}

\begin{sidewaystable}
\centering
    \caption{Mean Robustness ($\mathbf{M}$) for Networks Trained on a Single Noise Type}
    \resizebox{\textwidth}{!}{
     \begin{tabular}{c|ccc|ccc|ccc}
        \multicolumn{10}{c}{\textbf{(a) First break picking (mIoU)$^{*}$}} \\
        \hline
        \multicolumn{1}{c}{Network} & \multicolumn{3}{c}{\textbf{U-Net}} & \multicolumn{3}{c}{\textbf{Swin-U}} & \multicolumn{3}{c}{\textbf{Restormer}} \\
        \multicolumn{1}{c}{Clean baseline} & \multicolumn{3}{c}{$38.88\pm5.24$} & \multicolumn{3}{c}{$50.20\pm12.07$} & \multicolumn{3}{c}{$37.61\pm4.63$} \\
        \hline
        \multicolumn{1}{r|}{Noise Scale ($\boldsymbol{\beta}$)} & 0.05 & 0.25 & 1.0 & 0.05 & 0.25 & 1.0 & 0.05 & 0.25 & 1.0 \\
        \hline
        \multicolumn{1}{r|}{Gaussian} & $44.62\pm18.63$ & $87.55\pm10.62$ & $89.33\pm\phantom{0}9.56$ & $66.18\pm14.36$ & \textbf{92.14$\pm\phantom{0}$5.68} & \textbf{94.68$\pm\phantom{0}$4.45} & $47.16\pm18.52$ & $85.88\pm\phantom{0}9.74$ & $90.84\pm13.88$ \\
        \multicolumn{1}{r|}{Gaussian (f-k)} & $51.53\pm20.06$ & \textbf{88.80$\pm\phantom{0}$9.00} & $82.90\pm16.79$ & \textbf{81.18$\pm\phantom{0}$9.15} & \textbf{91.63$\pm\phantom{0}$5.98} & \textbf{95.03$\pm\phantom{0}$4.22} & $51.75\pm20.65$ & $87.19\pm\phantom{0}8.14$ & $91.22\pm12.36$ \\
        \hline
        \multicolumn{1}{r|}{Colored} & $49.39\pm19.12$ & \textbf{89.81$\pm\phantom{0}$8.56} & \textbf{97.04$\pm\phantom{0}$4.44} & $71.23\pm13.09$ & \textbf{90.13$\pm\phantom{0}$7.66} & \textbf{96.74$\pm\phantom{0}$3.46} & \textbf{62.77$\pm$19.71} & \textbf{90.07$\pm\phantom{0}$6.56} & \textbf{97.01$\pm\phantom{0}$4.20} \\
        \multicolumn{1}{r|}{Bandpassed} & $40.00\pm12.16$ & $49.70\pm19.03$ & $52.78\pm16.39$ & $43.59\pm16.32$ & $38.57\pm16.17$ & $38.78\pm17.23$ & $38.91\pm\phantom{0}9.97$ & $38.44\pm16.72$ & $39.20\pm17.22$ \\
        \hline
        \multicolumn{1}{r|}{Linear} & \textbf{54.90$\pm$19.75} & $69.91\pm21.11$ & $68.08\pm26.05$ & $69.12\pm18.01$ & \textbf{79.84$\pm$15.71} & $74.68\pm18.74$ & $49.99\pm24.32$ & $56.88\pm26.28$ & $54.80\pm27.63$ \\
        \multicolumn{1}{r|}{Hyperbolic} & $46.39\pm14.50$ & $58.90\pm23.59$ & $57.16\pm25.80$ & $60.21\pm16.92$ & \textbf{80.80$\pm$11.35} & $75.16\pm13.48$ & $42.04\pm17.36$ & $47.92\pm23.18$ & $50.67\pm26.99$ \\
        \hline
        \multicolumn{10}{l}{\footnotesize $^{*}$ 50 Epochs}  
    \end{tabular}
    }
    \vspace{1cm} 

    \resizebox{\textwidth}{!}{
    \begin{tabular}{c|ccc|ccc|ccc}
        \multicolumn{10}{c}{\textbf{(b) Denoising (RMSE)$^{*}$ }} \\
        \hline
        \multicolumn{1}{c}{Network} & \multicolumn{3}{c}{\textbf{U-Net}} & \multicolumn{3}{c}{\textbf{Swin-U}} & \multicolumn{3}{c}{\textbf{Restormer}} \\
        \multicolumn{1}{c}{Clean baseline} & \multicolumn{3}{c}{$0.195\pm0.031$} & \multicolumn{3}{c}{$0.252\pm0.076$} & \multicolumn{3}{c}{$0.254\pm0.079$} \\
        \hline
        \multicolumn{1}{r|}{Noise Scale ($\boldsymbol{\beta}$)} & 0.05 & 0.25 & 1.0 & 0.05 & 0.25 & 1.0 & 0.05 & 0.25 & 1.0 \\
        \hline
        \multicolumn{1}{r|}{Gaussian} & $0.173\pm0.028$ & $0.148\pm0.024$ & $0.138\pm0.019$ & \textbf{0.173$\pm$0.060} & $0.119\pm0.052$ & \textbf{0.096$\pm$0.048} & \textbf{0.138$\pm$0.048} & \textbf{0.112$\pm$0.057} & \textbf{0.087$\pm$0.055} \\
        \multicolumn{1}{r|}{Gaussian (f-k)} & $0.191\pm0.041$ & $0.149\pm0.026$ & $0.139\pm0.018$ & $0.181\pm0.065$ & $0.118\pm0.052$ & \textbf{0.096$\pm$0.045} & $0.145\pm0.047$ & \textbf{0.111$\pm$0.057} & \textbf{0.085$\pm$0.054} \\
        \hline
        \multicolumn{1}{r|}{Colored} & \textbf{0.171$\pm$0.030} & \textbf{0.141$\pm$0.019} & \textbf{0.132$\pm$0.020} & $0.221\pm0.086$ & \textbf{0.117$\pm$0.050} & \textbf{0.088$\pm$0.050} & $0.199\pm0.086$ & \textbf{0.115$\pm$0.056} & \textbf{0.070$\pm$0.058} \\
        \multicolumn{1}{r|}{Bandpassed} & $0.180\pm0.031$ & $0.191\pm0.039$ & $0.258\pm0.072$ & $0.264\pm0.100$ & $0.248\pm0.090$ & $0.324\pm0.113$ & $0.257\pm0.079$ & $0.252\pm0.084$ & $0.462\pm0.158$ \\
        \hline
        \multicolumn{1}{r|}{Linear} & \textbf{0.173$\pm$0.033} & $0.205\pm0.062$ & $0.223\pm0.076$ & $0.183\pm0.069$ & $0.200\pm0.100$ & $0.216\pm0.115$ & $0.179\pm0.069$ & $0.191\pm0.096$ & $0.240\pm0.151$ \\
        \multicolumn{1}{r|}{Hyperbolic} & $0.190\pm0.036$ & $0.204\pm0.059$ & $0.236\pm0.076$ & $0.217\pm0.073$ & $0.206\pm0.085$ & $0.217\pm0.090$ & $0.226\pm0.080$ & $0.197\pm0.076$ & $0.244\pm0.118$ \\
        \hline
        \multicolumn{10}{l}{\footnotesize $^{*}$ 100 Epochs}  
    \end{tabular}
    }
\label{T4}
\end{sidewaystable}

\begin{sidewaystable}[ph!]
\centering
    \caption{Mean Robustness ($\mathbf{M}$) for Networks Trained on a Compound Noises}
    \resizebox{\textwidth}{!}{
    \begin{tabular}{c|ccc|ccc|ccc}
        \multicolumn{10}{c}{\textbf{(a) First break picking (mIoU)$^{*}$}} \\
        \hline
        \multicolumn{1}{c}{Network} & \multicolumn{3}{c}{\textbf{U-Net}} & \multicolumn{3}{c}{\textbf{Swin-U}} & \multicolumn{3}{c}{\textbf{Restormer}} \\
        \multicolumn{1}{c}{Clean baseline} & \multicolumn{3}{c}{$38.88\pm5.24$} & \multicolumn{3}{c}{$50.20\pm12.07$} & \multicolumn{3}{c}{$37.61\pm4.63$} \\
        \hline
        \multicolumn{1}{r|}{Noise Scale ($\boldsymbol{\beta}$)} & 0.05 & 0.25 & 1.0 & 0.05 & 0.25 & 1.0 & 0.05 & 0.25 & 1.0 \\
        \hline
        \multicolumn{1}{r|}{Random} & $45.74\pm18.32$ & $88.14\pm\phantom{0}8.77$ & $87.98\pm15.11$ & $75.84\pm10.87$ & $89.77\pm\phantom{0}7.55$ & \textbf{95.06$\pm\phantom{0}$3.95} & $50.39\pm19.24$ & $85.65\pm\phantom{0}9.11$ & $86.48\pm14.94$ \\
        \multicolumn{1}{r|}{Structured} & \textbf{70.11$\pm$16.47} & $58.88\pm27.74$ & $64.33\pm27.53$ & $80.63\pm10.49$ & $79.53\pm15.09$ & $91.46\pm\phantom{0}5.83$ & $44.71\pm20.65$ & $60.78\pm26.50$ & $53.92\pm27.66$ \\
        \multicolumn{1}{r|}{Linear + Colored} & $55.10\pm20.36$ & $87.02\pm13.31$ & \textbf{97.57$\pm\phantom{0}$3.94} & $80.96\pm\phantom{0}9.71$ & $94.36\pm\phantom{0}4.98$ & \textbf{96.90$\pm\phantom{0}$3.77} & $70.72\pm23.66$ & $86.40\pm13.95$ & $95.32\pm\phantom{0}6.29$ \\
        \multicolumn{1}{r|}{Gaussian + Colored} & $44.92\pm18.75$ & $90.17\pm\phantom{0}8.04$ & \textbf{97.01$\pm\phantom{0}$4.23} & $83.12\pm\phantom{0}8.54$ & $90.39\pm\phantom{0}7.73$ & \textbf{96.68$\pm\phantom{0}$3.44} & $55.21\pm21.56$ & $90.52\pm\phantom{0}6.41$ & \textbf{96.54$\pm\phantom{0}$4.24} \\
        \hline
        \multicolumn{1}{r|}{Random + Colored} & $45.37\pm19.42$ & $90.61\pm8.30$ & \textbf{96.49$\pm\phantom{0}$4.67} & $82.02\pm\phantom{0}8.98$ & $91.32\pm\phantom{0}6.93$ & \textbf{96.16$\pm\phantom{0}$3.98} & $57.32\pm22.09$ & $84.05\pm\phantom{0}9.83$ & \textbf{96.14$\pm\phantom{0}$4.95} \\
        \multicolumn{1}{r|}{X: Gaussian + Colored + Linear} & $49.79\pm20.07$ & $85.88\pm12.74$ & \textbf{97.94$\pm\phantom{0}$3.32} & $75.49\pm13.58$ & $93.80\pm\phantom{0}5.29$ & \textbf{97.13$\pm\phantom{0}$3.38} & $67.27\pm18.44$ & $87.73\pm10.91$ & \textbf{97.78$\pm\phantom{0}$3.32} \\
        \multicolumn{1}{r|}{Random + Colored + Linear} & $47.69\pm19.37$ & $86.70\pm10.71$ & \textbf{97.35$\pm\phantom{0}$4.02} & $77.28\pm10.98$ & $93.60\pm\phantom{0}5.27$ & \textbf{96.35$\pm\phantom{0}$3.63} & $65.36\pm22.41$ & $90.44\pm\phantom{0}7.01$ & $97.63\pm\phantom{0}4.31$ \\
        \multicolumn{1}{r|}{Random + Colored + Structured} & $51.04\pm20.38$ & $86.67\pm11.63$ & \textbf{97.79$\pm\phantom{0}$3.55} & $74.80\pm12.47$ & $91.32\pm\phantom{0}7.41$ & \textbf{97.09$\pm\phantom{0}$3.33} & $58.95\pm21.64$ & $86.94\pm\phantom{0}9.97$ & \textbf{97.91$\pm\phantom{0}$3.25} \\
        \hline
        \multicolumn{1}{r|}{X + Trace-wise} & $50.23\pm19.25$ & $93.26\pm\phantom{0}7.59$ & \textbf{98.04$\pm\phantom{0}$3.10} & $87.35\pm\phantom{0}7.18$ & \textbf{95.79$\pm\phantom{0}$4.14} & \textbf{97.03$\pm\phantom{0}$2.25} & $57.54\pm23.70$ & \textbf{92.31$\pm\phantom{0}$9.97} & \textbf{98.77$\pm\phantom{0}$1.34} \\
        \multicolumn{1}{r|}{X+ Lowpass} & $45.74\pm19.92$ & $88.59\pm10.11$ & \textbf{96.99$\pm\phantom{0}$1.71} & $78.93\pm13.62$ & $93.13\pm\phantom{0}4.89$ & $95.98\pm\phantom{0}1.79$ & $58.21\pm21.06$ & $89.32\pm\phantom{0}7.19$ & \textbf{96.47$\pm\phantom{0}$2.79} \\
        \multicolumn{1}{r|}{X + Trace-wise + Lowpass} & $50.54\pm21.85$ & \textbf{94.34$\pm\phantom{0}$4.55} & \textbf{97.39$\pm\phantom{0}$1.47} & \textbf{89.69$\pm\phantom{0}$6.53} & \textbf{95.63$\pm\phantom{0}$3.16} & $90.23\pm\phantom{0}0.29$ & \textbf{83.58$\pm\phantom{0}$9.23} & \textbf{95.23$\pm\phantom{0}$3.61} & \textbf{97.16$\pm\phantom{0}$1.41} \\
        \hline
        \multicolumn{10}{l}{\footnotesize $^{*}$ 50 Epochs, Random =[Gaussian, Gaussian (f-k)], Structured =[Linear, Hyperbolic]} 
    \end{tabular}
    }
    \vspace{1cm} 
    \resizebox{\textwidth}{!}{
    \begin{tabular}{r|ccc|ccc|ccc}
        \multicolumn{10}{c}{\textbf{(b) Denoising (RMSE)$^{*}$ }} \\
        \hline
        \multicolumn{1}{r}{Network} & \multicolumn{3}{c}{\textbf{U-Net}} & \multicolumn{3}{c}{\textbf{Swin-U}} & \multicolumn{3}{c}{\textbf{Restormer}} \\
        \multicolumn{1}{r}{Clean baseline} & \multicolumn{3}{c}{$0.195\pm0.031$} & \multicolumn{3}{c}{$0.252\pm0.076$} & \multicolumn{3}{c}{$0.254\pm0.079$} \\
        \hline
        \multicolumn{1}{r|}{Noise Scale ($\boldsymbol{\beta}$)} & 0.05 & 0.25 & 1.0 & 0.05 & 0.25 & 1.0 & 0.05 & 0.25 & 1.0 \\
        \hline
        \multicolumn{1}{r|}{Random} & $0.167\pm0.031$ & $0.147\pm0.023$ & $0.138\pm0.018$ & $0.176\pm0.060$ & $0.118\pm0.051$ & \textbf{0.096$\pm$0.046} & $0.140 \pm 0.051$ & $0.116\pm0.056$ & \textbf{0.086$\pm$0.055} \\
        \multicolumn{1}{r|}{Structured} & $0.173\pm0.034$ & $0.212\pm0.067$ & $0.227\pm0.075$ & $0.179\pm0.065$ & $0.192\pm0.096$ & $0.247\pm0.135$ & $0.173\pm0.069$ & $0.166\pm0.086$ & $0.188\pm0.113$ \\
        \multicolumn{1}{r|}{Linear + Colored} & $0.162\pm0.027$ & $0.139\pm0.020$ & $0.133\pm0.022$ & $0.165\pm0.062$ & $0.111\pm0.053$ & \textbf{0.089$\pm$0.054} & $0.134\pm0.056$ & \textbf{0.098$\pm$0.060} & \textbf{0.072$\pm$0.059} \\
        \multicolumn{1}{r|}{Gaussian + Colored} & $0.167\pm0.026$ &$0.140\pm0.019$ & \textbf{0.130$\pm$0.020} & $0.190\pm0.070$ & $0.115\pm0.052$ & \textbf{0.088$\pm$0.046} & $0.135\pm0.050$ & \textbf{0.102$\pm$0.059} & \textbf{0.072$\pm$0.058} \\
        \hline
        \multicolumn{1}{r|}{Random + Colored} & $0.173\pm0.030$ & $0.139\pm0.020$ & $0.131\pm0.020$  & $0.176\pm0.064$ & $0.119\pm0.049$ & \textbf{0.088$\pm$0.046} & $0.142\pm0.048$ & \textbf{0.108$\pm$0.058} & \textbf{0.075$\pm$0.056} \\
        \multicolumn{1}{r|}{X: Gaussian + Colored + Linear} & $0.162\pm0.026$ & $0.139\pm0.019$ & \textbf{0.130$\pm$0.021} & $0.161\pm0.059$ & \textbf{0.106$\pm$0.053} & \textbf{0.085$\pm$0.049} & $0.132\pm0.048$ & \textbf{0.100$\pm$0.060} & \textbf{0.063$\pm$0.058} \\
        \multicolumn{1}{r|}{Random + Colored + Linear} & $0.164\pm0.025$ & $0.139\pm0.020$ & \textbf{0.130$\pm$0.020} & $0.161\pm0.059$ & \textbf{0.107$\pm$0.053} & \textbf{0.085$\pm$0.047} & \textbf{0.129$\pm$0.049} & \textbf{0.102$\pm$0.060} & \textbf{0.067$\pm$0.055} \\
        \multicolumn{1}{r|}{Random + Colored + Structured} & $0.163\pm0.026$ & $0.139\pm0.020$ & $0.131\pm0.020$ & \textbf{0.150$\pm$0.053} & \textbf{0.107$\pm$0.053} & \textbf{0.085$\pm$0.050} & \textbf{0.128$\pm$0.048} & \textbf{0.101$\pm$0.060} & \textbf{0.072$\pm$0.057} \\
        \hline
        \multicolumn{1}{r|}{X + Trace-wise} & \textbf{0.157$\pm$0.023} & \textbf{0.138$\pm$0.019} & $0.133\pm0.022$ & $0.198\pm0.073$ & \textbf{0.104$\pm$0.048} & \textbf{0.091$\pm$0.049} & $0.138\pm0.052$ & \textbf{0.095$\pm$0.055} & \textbf{0.077$\pm$0.058} \\
        \multicolumn{1}{r|}{X + Lowpass} & $0.185\pm0.016$ & $0.173\pm0.018$ & $0.181\pm0.021$ & $0.212\pm0.030$ & $0.198\pm0.033$ & $0.214\pm0.052$ & $0.178\pm0.028$ & $0.184\pm0.038$ & $0.210\pm0.054$ \\
        \multicolumn{1}{r|}{X + Trace-wise + Lowpass} & $0.200\pm0.024$ & $0.171\pm0.018$ & $0.181\pm0.021$ & $0.209\pm0.029$ & $0.190\pm0.044$ & $0.211\pm0.044$ & $0.181\pm0.027$ & $0.193\pm0.045$ & $0.214\pm0.056$ \\
        \hline
        \multicolumn{10}{l}{\footnotesize $^{*}$ 100 Epochs, Random =[Gaussian, Gaussian (f-k)], Structured =[Linear, Hyperbolic]}  
    \end{tabular}
    }
\label{T5}
\end{sidewaystable}

\end{document}